%% file: main_paper.tex
\documentclass[10pt,twocolumn,letterpaper]{article}

\usepackage[dvipsnames]{xcolor}
\usepackage{cvpr}
\usepackage{times}
\usepackage{epsfig}
\usepackage{graphicx}
\usepackage{amsmath}
\usepackage{amssymb}
\usepackage{subcaption}
\usepackage{caption} % for teaser caption (since it has to be outside of floating figure environment)
\usepackage{eurosym}
\usepackage{wrapfig}
\usepackage{tikz}
\usetikzlibrary{positioning,calc}

\usepackage[breaklinks=true,bookmarks=false,colorlinks=true]{hyperref}

\input{macros.tex}

\cvprfinalcopy % *** Uncomment this line for the final submission

\def\cvprPaperID{xxxx} % *** Enter the CVPR Paper ID here

\begin{document}

%%%%%%%%% TITLE
\title{GANerated Hands for Real-Time 3D Hand Tracking from Monocular RGB}

\author{
	Franziska Mueller$^{1}$ \quad Florian Bernard$^{1}$ \quad
	Oleksandr Sotnychenko$^{1}$ \quad Dushyant Mehta$^{1}$ \\[0.1em]
	Srinath Sridhar$^{2}$ \quad Dan Casas$^{3}$\quad Christian Theobalt$^{1}$ \\[1.0em] 
	$^{1}$ Max-Planck-Institute for Informatics\quad
	$^{2}$ Stanford University \quad
	$^{3}$ Universidad Rey Juan Carlos\\[0.5em]
}

\twocolumn[{
 \renewcommand\twocolumn[1][]{#1}
 \maketitle
 {
  \centering
  \vspace{-0.8cm}
  \includegraphics[width=\linewidth]{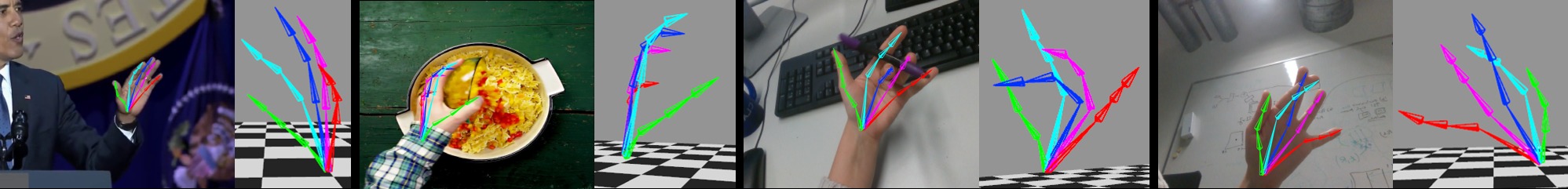} 
  \captionof{figure}{
   We present an approach for real-time 3D hand tracking from monocular RGB-only input. Our method is compatible with unconstrained video input such as community videos from YouTube (left), and robust to occlusions (center-left). We show real-time 3D hand tracking results using an off-the-shelf RGB webcam in unconstrained setups (center-right, right).
   \label{fig:teaser}
  }
 }
 \vspace{0.5cm}
}]

%%%%%%%%% ABSTRACT
\begin{abstract}
We address the highly challenging problem of real-time 3D hand tracking based on a monocular RGB-only sequence. Our tracking method combines a convolutional neural network with a kinematic 3D hand model, such that it generalizes well to unseen data, is robust to occlusions and varying camera viewpoints, and leads to anatomically plausible as well as temporally smooth hand motions. For training our CNN we propose a novel approach for the synthetic generation of training data that is based on a geometrically consistent image-to-image translation network. To be more specific, we use a neural network that translates synthetic images to ``real'' images, such that the so-generated images follow the same statistical distribution as real-world hand images.
For training this translation network we combine an adversarial loss and a cycle-consistency loss with a geometric consistency loss in order to preserve geometric properties (such as hand pose) during translation. We demonstrate that our hand tracking system outperforms the current state-of-the-art on challenging RGB-only footage.
\end{abstract}

%%%%%%%%% BODY TEXT
\input{intro}
\input{relatedwork}
\input{methods}
\input{experiments}
\input{conclusion}

\parahead{Acknowledgements}
This work was supported by the ERC Starting Grant CapReal (335545).
Dan Casas was supported by a Marie Curie Individual Fellow, grant 707326.

\clearpage

\input{supp.tex}

{\small
\bibliographystyle{ieee}
\bibliography{GANerativeHands}
}

\end{document}

%% file: macros.tex
\newcommand{\parahead}[1]{\textbf{#1}:\ }

\newenvironment{packed_itemize}
{\begin{itemize}
    \setlength{\itemsep}{1pt}
    \setlength{\parskip}{0pt}
    \setlength{\parsep}{0pt}
}{\end{itemize}}

\newcommand{\R}{\mathbb{R}}
\newcommand{\SO}{\operatorname{SO}}

\newcommand{\filluptopage}[1]{%
  \clearpage
  \loop\ifnum\value{page}<#1\relax
    \null\clearpage
  \repeat
  \loop\ifnum\value{page}=#1\relax
    \null\clearpage
  \repeat
}

\newcommand{\cref}[1]{Section \ref{#1}}

\newcommand{\gan}{\emph{GeoConGAN}}
\newcommand{\reg}{\emph{RegNet}}

%% file: intro.tex
\section{Introduction}
Estimating the 3D pose of the hand is a long-standing goal in computer vision with many applications such as in virtual/augmented reality (VR/AR)~\cite{lee2009multithreaded,piumsomboon2013user} and human--computer interaction~\cite{sridhar_chi2015,markussen2014vulture}.
While there is a large body of existing works that consider marker-free image-based hand tracking or pose estimation, many of them require depth cameras~\cite{sharp2015accurate, taylor_siggraph2016,sridhar_cvpr2015,tagliasacchi_sgp2015,qian_cvpr2014,ge_cvpr2016,wan2017crossing} or multi-view setups~\cite{sridhar_iccv2013,ballan_eccv2012,wang20116d}.
However, in many applications these requirements are unfavorable since such hardware is less ubiquitous, more expensive, and does not work in all scenes.

In contrast, we address these issues and propose a new algorithm for \emph{real-time skeletal 3D hand tracking} with a \emph{single color camera} that is robust under object \emph{occlusion and clutter}.
Recent developments that consider RGB-only markerless hand tracking problem~\cite{simon2017hand,Zimmermann:2017um,gomez_arxiv2017} come with clear limitations. 
For example, the approach by Simon \etal \cite{simon2017hand} achieves the estimation of 3D joint locations within a multi-view setup; however in the monocular setting only 2D joint locations are estimated. 
Similarly, the method by Gomez-Donoso \etal~\cite{gomez_arxiv2017} is also limited to 2D.
Recently, Zimmermann and Brox \cite{Zimmermann:2017um} presented a 3D hand pose estimation method from monocular RGB which, however, only obtains relative 3D positions and struggles with occlusions.

Inspired by recent work on hand and body tracking \cite{tompson_tog2014,mueller_iccv2017,mehta_SIGGRAPH2017}, we combine CNN-based 2D and 3D hand joint predictions with a kinematic fitting step to track hands in global 3D from monocular RGB.
The major issue of such (supervised) learning-based approaches is the requirement of suitable \emph{annotated} training data. 
While it has been shown to be feasible to manually annotate 2D joint locations in single-view RGB images \cite{johnson2010lsp}, it is impossible to accurately annotate in 3D due to the inherent depth ambiguities.
One way to overcome this issue is to leverage existing multi-camera methods for tracking hand motion in 3D \cite{sridhar_iccv2013,ballan_eccv2012,wang20116d,gomez_arxiv2017}.
However, the resulting annotations would lack precision due to inevitable tracking errors. 
Some works render synthetic hands for which the perfect ground truth is known~\cite{mueller_iccv2017,Zimmermann:2017um}.
However, CNNs trained on synthetic data may not always generalize well to real-world images.
Hence, we propose a method to \emph{generate suitable training data} by performing image-to-image translation between synthetic and real images. 
We impose two strong requirements on this method.
First, we want to be able to train on \emph{unpaired images} so that we can easily collect a large-scale real hands dataset.
Second, we need the algorithm to preserve the pose of the hand such that the annotations of the synthetic images are still valid for the translated images.
To this end, we leverage the seminal work on CycleGANs \cite{CycleGAN2017}, which successfully learns various image-to-image translation tasks with unpaired examples.
We extend it with a \textit{geometric consistency loss} which improves the results in scenarios where we only want to learn spatially localized (\eg only the hand part) image-to-image conversions, producing pose-preserving results with less texture bleeding and sharper contours. 
Once this network is trained, we can use it to translate any synthetically generated image into a ``real'' image while preserving the perfect (and inexpensive) ground truth annotation.
Throughout the rest of the paper we denote images as ``real'' (in quotes), or \emph{GANerated}, when we refer to synthetic images after they have been processed by our translation network such that they follow the same statistical distribution as real-world images.

Finally, using annotated RGB-only images produced by the proposed GAN, we train a CNN that jointly regresses image-space 2D and root-relative 3D hand joint positions. 
While the skeletal hand model in combination with the 2D predictions are sufficient to estimate the global translation of the hand, the relative 3D positions resolve the inherent ambiguities in global rotation and articulation which occur in the 2D joint positions.
In summary, our main contributions are:
\begin{packed_itemize}
\item The first real-time hand tracking system that tracks \emph{global 3D joint} positions from unconstrained monocular RGB-only images.
\item A novel geometrically consistent GAN that performs image-to-image translation while preserving poses during translation.
\item Based on this network, we are able to \emph{enhance synthetic hand image datasets} such that the statistical distribution resembles real-world hand images.
\item A \emph{new RGB dataset} with annotated 3D hand joint positions. We overcome existing datasets in terms of size (${>}260$k frames), image fidelity, and annotation precision. 
\end{packed_itemize}

\begin{figure*}[ht!] 
  \centerline{ 
  \includegraphics[width=\linewidth,trim={0cm 13.2cm 0cm 0cm},clip]{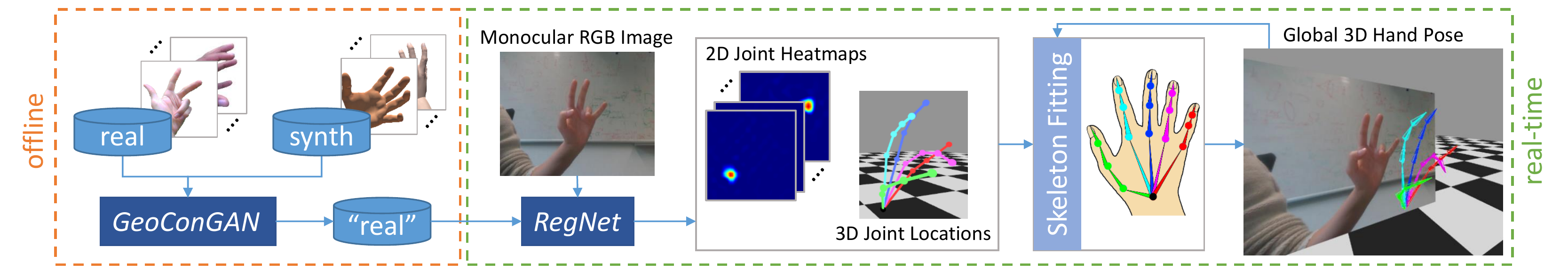} 
        }
        \vspace{-3mm}
    \caption{Pipeline of our real-time system for monocular RGB hand tracking in 3D.} 
    \label{fig:pipeline}
\end{figure*}

%% file: relatedwork.tex
\section{Related Work}
\label{sec:related_work}
Our goal is to track hand pose from \emph{unconstrained monocular RGB video streams} at real-time framerates.
This is a challenging problem due the large pose space, occlusions due to objects, depth ambiguity, appearance variation due to lighting and skin tone, and camera viewpoint variation.
While glove-based solutions would address some of these challenges~\cite{wang_TOG2009}, they are cumbersome to wear.
Thus, in the following we restrict our discussion to markerless camera-based methods that try to tackle these challenges.

\parahead{Multi-view methods}
The use of multiple RGB cameras considerably alleviates occlusions during hand motion and interaction.
Wang \etal~\cite{wang20116d} demonstrated hand tracking with two cameras using a discriminative approach to quickly find the closest pose in a database.
Oikonomidis \etal~\cite{oikonomidis2011full} showed tracking of both the hand and a manipulated object using 8 calibrated cameras in a studio setup.
Ballan \etal~\cite{ballan_eccv2012} also used 8 synchronized RGB cameras to estimate pose with added input from discriminatively detected points on the fingers.
Sridhar \etal~\cite{sridhar_iccv2013,sridhar2014real} used 5 RGB cameras and an additional depth sensor to demonstrate real-time hand pose estimation.
Panteleris and Argyros~\cite{panteleris2017back} propose using a short-baseline stereo camera for hand pose estimation without the need for a disparity map.
All of the above approaches utilize multiple calibrated cameras, making it hard to setup and operate on general hand motions in unconstrained scenes (\eg community videos).
More recently, Simon \etal~\cite{simon2017hand} proposed a method to generate large amounts of 2D and 3D hand pose data by using a panoptic camera setup which restricts natural motion and appearance variation.
They also leverage their data (which is not yet publicly available) for 2D hand pose estimation but cannot estimate 3D pose in monocular RGB videos.
Our contributions address both data variation for general scenes and the difficult 3D pose estimation problem.

\parahead{Monocular methods}
Monocular methods for 3D hand pose estimation are preferable because they can be used for many applications without a setup overhead.
The availability of inexpensive consumer depth sensors has lead to extensive research in using them for hand pose estimation.
Hamer \etal~\cite{hamer2009tracking} proposed one of the first generative methods to use monocular RGB-D data for hand tracking, even under partial occlusions.
As such methods often suffer from issues due to local optima, a learning-based discriminative method was proposed by Keskin \etal~\cite{kesin_iccvw2011}.
Numerous follow-up works have been proposed to improve the generative component~\cite{tagliasacchi_sgp2015,tang_iccv2015,tzionas_ijcv2016,tzionas20153d}, and the learning-based discriminator~\cite{xu2013efficient,krejov2013multi,tang2014latent,tompson_tog2014,wan_eccv2016,ge_cvpr2016,sinha_cvpr2016,oberweger_iccv2015,zhou_ijcai2016,choi2017learning,choi2017robust}.
Hybrid methods that combine the best of both generative and discriminative methods show the best performance on benchmark datasets~\cite{taylor_siggraph2016,sridhar_cvpr2015,sridhar_eccv2016,mueller_iccv2017,ye_eccv2016}.

Despite all the above-mentioned progress in monocular RGB-D or depth-based hand pose estimation, it is important to notice that these devices do not work in all scenes, \eg outdoors due to interference with sunlight, and have higher power consumption.
Furthermore, 3D hand pose estimation in unconstrained RGB videos would enable us to work community videos as shown in Figure~\ref{fig:teaser}.
Some of the first methods for this problem~\cite{heap1996towards,stenger2006model,romero_icra2010} did not produce metrically accurate 3D pose as they only fetched the nearest 3D neighbor for a given input or assume that the $z$-coordinate is fixed.
Zimmermann and Brox~\cite{Zimmermann:2017um} proposed a learning-based method to address this problem.
However, their 3D joint predictions are \emph{relative} to a canonical frame, \ie the absolute coordinates are unknown, and it is not robust to occlusions by objects.
Furthermore, their method is not able to distinguish 3D poses with the same 2D joint position projection since their 3D predictions are merely based on the abstract 2D heatmaps and do not directly take the image into account.
In contrast, our work addresses these limitations
by jointly learning 2D and 3D joint positions from image evidence, so that we are able to correctly estimate poses with ambiguous 2D joint positions.
In addition, our skeleton fitting framework combines a prior hand model with these predictions to obtain global 3D coordinates. 

\parahead{Training of learning-based methods}
One of the challenges in using learning-based models for hand pose estimation is the difficulty of obtaining annotated data with sufficient real-world variations.
For depth-based hand pose estimation, multiple training datasets have been proposed that leverage generative model fitting to obtain ground truth annotations~\cite{tompson_tog2014,sridhar_cvpr2015} or to sample pose space better~\cite{oberweger_cvpr2016}.
A multi-view bootstrapping approach was proposed by Simon \etal~\cite{simon2017hand}.
However, such outside-in capture setups could still suffer from occlusions due to objects being manipulated by the hand.
Synthetic data is promising for obtaining perfect ground truth but there exists a domain gap when models training on this data are applied to real input~\cite{mueller_iccv2017}.

Techniques like domain adaptation~\cite{ganin2014unsupervised,tzeng2017adversarial,peng2017synthetic} aim to bridge the gap between real and synthetic data by learning features that are invariant to the underlying differences.
Other techniques use real--synthetic image pairs~\cite{isola2017pix,sangkloy2016scribbler,chen_iccv2017} to train networks that can generate images that contain many features of real images.
Because it is hard to obtain real--synthetic image pairs for hands, we build upon the \emph{unpaired} image-to-image translation work of Zhu \etal~\cite{CycleGAN2017}.
Without the need for corresponding real--synthetic image pairs, we can generate images of hands that contain many of the features found in real datasets.

%% file: methods.tex
\section{Hand Tracking System}
The main goal of this paper is to present a real-time system for monocular RGB-only hand tracking in 3D. The overall system is outlined in Fig.~\ref{fig:pipeline}.
Given a live monocular RGB-only video stream we use a CNN hand joint regressor, the \reg{}, to predict 2D joint heatmaps and 3D joint positions (Sec.~\ref{sec:CNN_hand_pose_estimation}).
The \reg{} is trained with images that are generated by a novel image-to-image translation network, the \gan, (Sec.~\ref{sec:GAN_for_training}) that enriches synthetic hand images. The output images of the \gan--the \textit{GANerated} images--are better suited to train a CNN that will work on real imagery. 
After joint regression, we fit a kinematic skeleton to both the 2D and 3D predictions by minimizing our fitting energy (Sec.~\ref{sec:kinematic_fitting}), which has
several key advantages for achieving a robust 3D hand pose tracking: it enforces biomechanical plausibility; we can retrieve the \emph{absolute} 3D positions; and furthermore we are able to impose temporal stability across multiple frames.

\subsection{Generation of Training Data}
\begin{figure*}[ht!] 
  \centerline{ 
  \includegraphics[width=\linewidth]{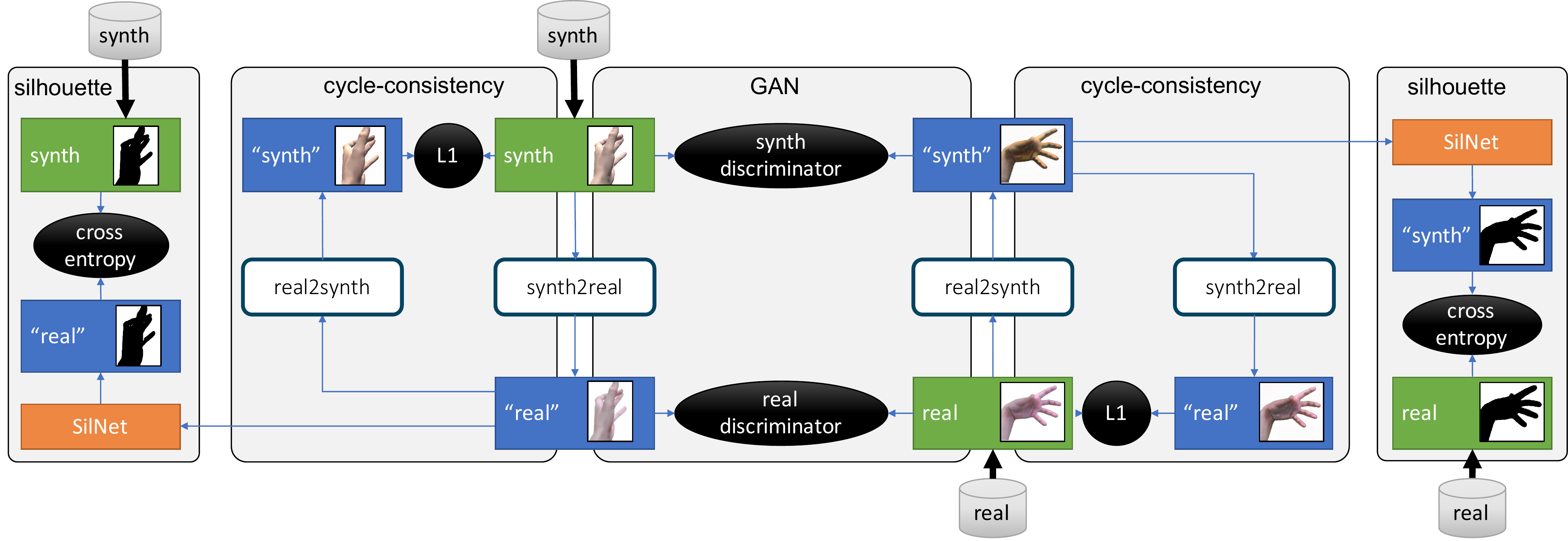} 
        }
        \vspace{-3mm}
    \caption{Network architecture of our \gan. The trainable part comprises the \emph{real2synth} and the \emph{synth2real} components, where we show both components twice for visualization purposes. The loss functions are shown in black, images from our database in green boxes, images generated by the networks in blue boxes, and the existing \emph{SilNet} in orange boxes.} 
    \label{fig:ganarchitecture}
\end{figure*}
\label{sec:GAN_for_training}
Since the annotation of 3D joint positions in hundreds of real hand images is infeasible, 
a common approach is to use synthetically generated images. 
While the main advantage of synthetic images is that the ground truth 3D joint positions are known, an important shortcoming is that they usually lack realism.
Such discrepancy between real and synthetic images limits the generalization ability of a CNN trained only on the latter.
In order to account for this disparity, we propose to use an image-to-image translation network, the \gan, with the objective to translate synthetic images to real images.
Most importantly, to train this network we use \emph{unpaired} real and synthetic images, as will be described in the following. Note that for both the real and the synthetic data we use only foreground-segmented images that contain a hand on white background, which facilitates background augmentation for learning the joint position regressor, as described in Sec.~\ref{sec:CNN_hand_pose_estimation}.

\parahead{Real hand image acquisition}
To acquire our dataset of real images we used a green-screen setup to capture hand images with varying poses and camera extrinsics from 7 different subjects with different skin tones and hand shapes. 
In total, we captured 28,903 real hand images using a desktop webcam with image resolution $640\times480$.
\newcommand{\figScaleRealHands}{.08}
\newcommand{\figScaleRealHandsBottom}{40px}

\parahead{Synthetic hand image generation}
Our synthetic hand image dataset is a combination of the SynthHands dataset \cite{mueller_iccv2017} that contains hand images from an egocentric viewpoint, with our own renderings of hand images from various third-person viewpoints. 
In order to generate the latter, we used the standard strategy in state-of-the-art datasets \cite{mueller_iccv2017,Zimmermann:2017um}, where the hand motion, obtained either via a hand tracker or a hand animation platform, is re-targeted to a kinematic 3D hand model.

\parahead{Geometrically consistent CycleGAN (\gan)}
While the above procedure allows to generate a large amount of synthetic training images with diverse hand pose configurations, training a hand joint regression network based on synthetic images alone has the strong disadvantage that the so-trained network has limited generalization to real images, as we will demonstrate in Sec.~\ref{sec:quantitative-evaluation}

To tackle this problem, we train a network that translates synthetic images to ``real'' (or \emph{GANerated}) images.
Our translation network is based on CycleGAN \cite{CycleGAN2017}, which uses adversarial discriminators \cite{goodfellow2014generative} to simultaneously learn cycle-consistent forward and backward mappings. Cycle-consistency means that the composition of both mappings (in either direction) is the identity mapping. In our case we learn mappings from synthetic to real images (\emph{synth2real}), and from real to synthetic images (\emph{real2synth}). In contrast to many existing image-to-image or style transfer networks \cite{isola2017pix,sangkloy2016scribbler}, CycleGAN has the advantage that it does not require paired images, \ie there must not exist a real image counterpart for a given synthetic image, which is crucial for our purpose due to the unavailability of such pairs.

The architecture of this \gan{} is illustrated in Fig.~\ref{fig:ganarchitecture}. 
The input to this network are (cropped)
synthetic and real images of the hand on a white background in conjunction with their respective silhouettes, \ie foreground segmentation masks.
In its core, The \gan{} resembles CycleGAN \cite{CycleGAN2017} with its discriminator and cycle-consistency loss, as well as the two trainable translators \emph{synth2real} and \emph{real2synth}. 
However, unlike CycleGAN, we incorporate an additional geometric consistency loss (based on cross-entropy) that ensures that the \emph{real2synth} and \emph{synth2real} components produce images that \emph{maintain the hand pose} during image translation.
Enforcing consistent hand poses is of utmost importance in order to ensure that the ground truth joint locations of the synthetic images are also valid for the ``real'' images produced by \emph{synth2real}.
Fig. \ref{fig:silhouette_loss_comparison} shows the benefits of adding this new loss term.

{
\newcommand{\figScale}{0.158}
\begin{figure}[t]
	\centering
	
		\begin{subfigure}[t]{\columnwidth}
			\begin{tikzpicture}
			\node (image) at (0,0) {\includegraphics[width=\figScale\columnwidth]{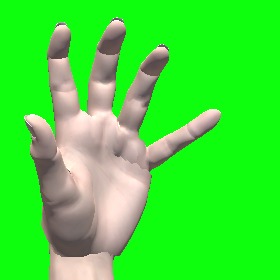}};
			\node[above left = -1mm and -14mm of image] (title1)  {\footnotesize{Synthetic}};
			
			\node[right = -0.2cm of image] (image2)  {\includegraphics[width=\figScale\columnwidth]{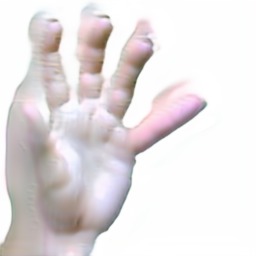}};
			\node[above left = -1mm and -15.5mm of image2] (title2)  {\footnotesize{CycleGAN}};
			
			\node[right = -0.2cm of image2] (image3)  {\includegraphics[width=\figScale\columnwidth]{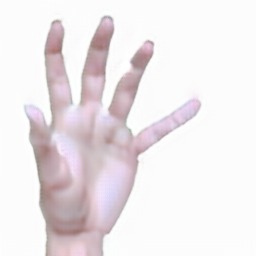}};
			\node[above left = -0.5mm and -16mm of image3] (title3)  {\footnotesize{GANerated}};
			
			\node[right = 0cm of image3] (image4)  {\includegraphics[width=\figScale\columnwidth]{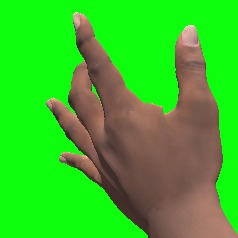}};
			\node[above left = -1mm and -13.5mm of image4] (title1)  {\footnotesize{Synthetic}};
			
			\node[right = -0.3cm of image4] (image5)  {\includegraphics[width=\figScale\columnwidth]{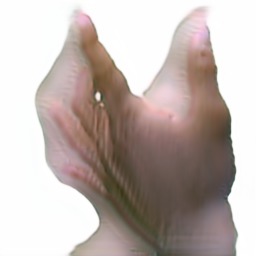}};
			\node[above left = -1mm and -15.5mm of image5] (title1)  {\footnotesize{CycleGAN}};
			
			\node[right = -0.2cm of image5] (image6)  {\includegraphics[width=\figScale\columnwidth]{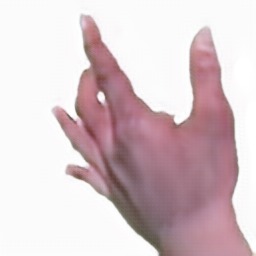}};
			\node[above left = -0.5mm and -16mm of image6] (title5)  {\footnotesize{GANerated}};
			
			\end{tikzpicture}
	\end{subfigure}  
\vspace{-7mm} 
    \caption{Our \gan{} translates from synthetic to real images by using an additional geometric consistency loss.
    	} 
    \label{fig:silhouette_loss_comparison}
\end{figure} 
}

In order to extract the silhouettes of the images that are produced by both \emph{real2synth} and \emph{synth2real} (blue boxes in Fig.~\ref{fig:ganarchitecture}), 
we train a binary classification network based on a simple UNet~\cite{unet2015} that has three 2-strided convolutions and three deconvolutions, which we call \emph{SilNet}. 
Note that this is a relatively easy task as the images show hands on a white background. Our \emph{SilNet} is trained beforehand and it is kept fixed while training \emph{synth2real} and \emph{real2synth}.
Training details can be found in the supplementary document. 

\parahead{Data augmentation}
Once the \gan{} is trained, we feed all synthetically generated images into the \emph{synth2real} component and obtain the set of ``real'' images that have associated ground truth 3D joint locations.
By using the background masks from the original synthetic images, we perform background augmentation by compositing GANerated images (foreground) with random images (background) \cite{varol_cvpr2017,mehta_SIGGRAPH2017,rhodin2016egocap}. 
Similarly, we also perform an object augmentation by leveraging the object masks produced when rendering the synthetic sequences \cite{mueller_iccv2017}. Fig.~\ref{fig:GANeratedImages} shows some GANerated images.

{
\newcommand{\figScale}{0.158}
\begin{figure}[t]
	\begin{subfigure}[t]{\columnwidth}
    \begin{tikzpicture}
    \node (image) at (0,0) {\includegraphics[width=\figScale\columnwidth]{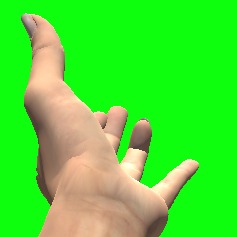}};
    \node[above left = -1mm and -14mm of image] (title1)  {\footnotesize{Synthetic}};
    		    
	\node[right = -0.2cm of image] (image2)  {\includegraphics[width=\figScale\columnwidth]{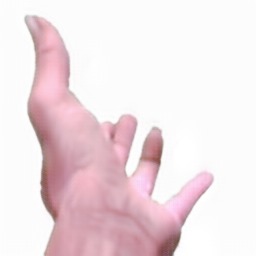}};
	\node[above left = -0.3mm and -15.5mm of image2] (title2)  {\footnotesize{GANerated}};
	
	\node[right = -0.2cm of image2] (image3)  {\includegraphics[width=\figScale\columnwidth]{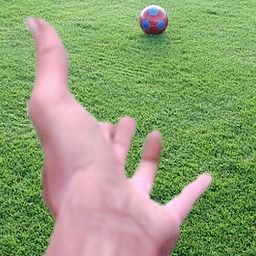}};
	\node[above left = 0.5mm and -16mm of image3] (title3)  {\footnotesize{GANerated}};
	\node[below = -0.2cm of title3] {\footnotesize{+ BG}};
	
	\node[right = 0cm of image3] (image4)  {\includegraphics[width=\figScale\columnwidth]{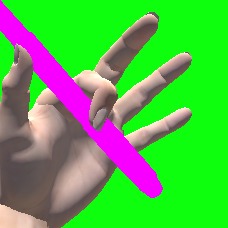}};
	\node[above left = -1mm and -13.5mm of image4] (title1)  {\footnotesize{Synthetic}};

	\node[right = -0.3cm of image4] (image5)  {\includegraphics[width=\figScale\columnwidth]{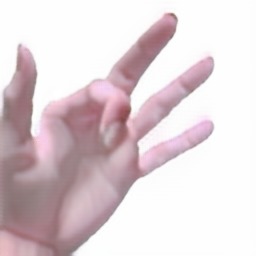}};
	\node[above left = -0.3mm and -15.5mm of image5] (title1)  {\footnotesize{GANerated}};
	
	\node[right = -0.2cm of image5] (image6)  {\includegraphics[width=\figScale\columnwidth]{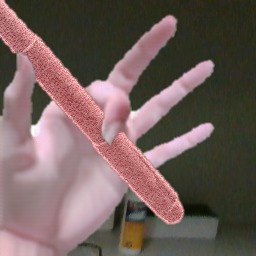}};
	\node[above left = 0.5mm and -16mm of image6] (title5)  {\footnotesize{GANerated}};
	\node[below = -0.2cm of title5] {\footnotesize{+ BG + FG}};
		\end{tikzpicture}
	\end{subfigure} 
	     
\vspace{-3mm} 
    \caption{Two examples of synthetic images with background/object masks in green/pink. 
    }
    \label{fig:GANeratedImages}
\end{figure} 
}

\subsection{Hand Joints Regression}
\label{sec:CNN_hand_pose_estimation}
\begin{figure}[t] 
  \centerline{ 
  \includegraphics[width=\linewidth]{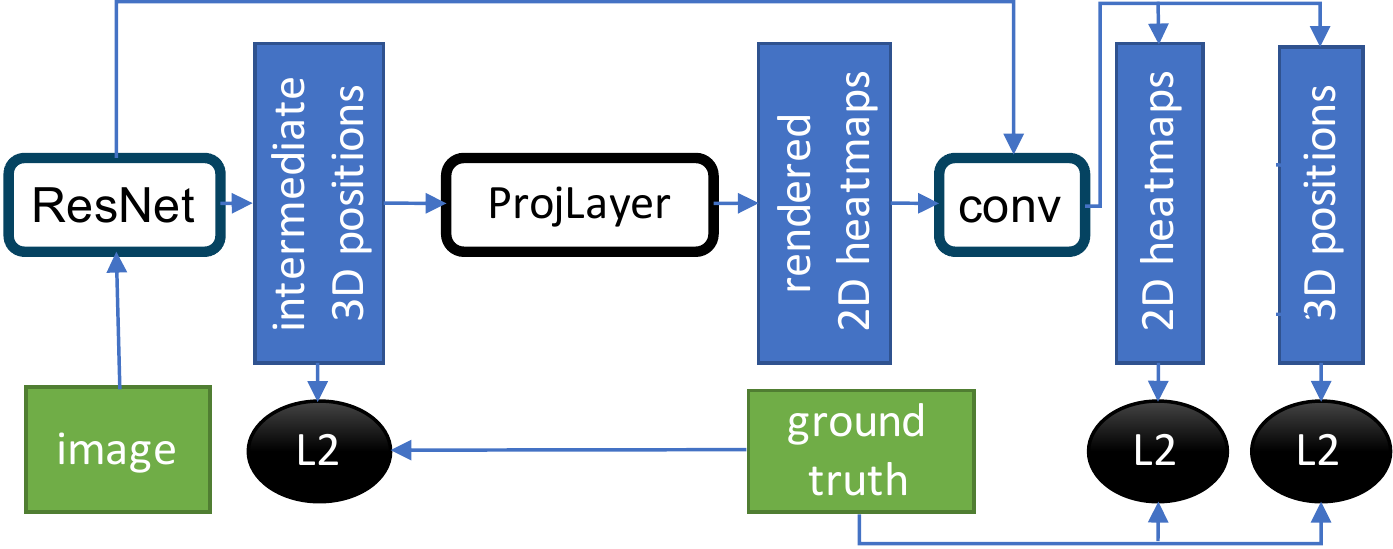} 
        }
        \vspace{-3mm}
    \caption{Architecture of \reg{}. While only \emph{ResNet} and \emph{conv} are trainable, errors are still back-propagated through our \emph{ProjLayer}. Input data is shown in green, data generated by the network in blue, and the loss is shown in black.} 
    \label{fig:regnetarchitecture}
\end{figure}
In order to regress the hand pose from a (cropped) RGB image of the hand, we train a CNN, the \reg{}, that predicts 2D and 3D positions of $21$ hand joints. The 2D joint positions are represented as heatmaps in image space, and the 3D positions are represented as 3D coordinates relative to the root joint. We have found that regressing both 2D and 3D joints are complementary to each other, as the 2D heatmaps are able to represent uncertainties, 
whereas the 3D positions resolve the depth ambiguities. 

The \reg{}, shown in Fig.~\ref{fig:regnetarchitecture}, is based on a residual network consisting of 10 residual blocks that is derived from the \emph{ResNet50} architecture~\cite{he_cvpr2016}, as done in \cite{mueller_iccv2017}.  
Additionally, we incorporate a (differentiable) refinement module based on a projection layer (\emph{ProjLayer}) to better coalesce the 2D and 3D predictions.
The idea of the \emph{ProjLayer} is to perform an orthographic projection of (preliminary) intermediate 3D predictions, from which 2D Gaussian heatmaps are created (within the layer). These heatmaps are then leveraged in the remaining part of the network (\emph{conv}) to obtain the final 2D and 3D predictions.
In Fig.~\ref{fig:jointregressionwithsyntheticimages} we show that this leads to improved results.

The training is based on a mixture of GANerated (Sec.~\ref{sec:GAN_for_training}) and synthetic images, 
in conjunction with corresponding (normalized) 3D ground truth joint positions.
The training set contains $\approx 440,000$ examples in total from which 60\% are GANerated.
We train the \reg{} with \emph{relative} 3D joint positions, which we compute by normalizing the absolute 3D ground truth joint positions such that the middle finger metacarpophalangeal (MCP) joint is at the origin and the distance between the wrist joint and the middle MCP joint is 1.  Further details can be found in the supplementary document. 

During test time, \ie for hand tracking, the input to the \reg{} is a cropped RGB image, where the (square) bounding box is derived from the 2D detections of the previous frame. In the first frame, the square bounding box is located at the center of the image, with size equal to the input image height. Also, we filter the output of \reg{} with the 1\euro{} filter \cite{casiez20121} to obtain temporally smoother predictions.

\subsection{Kinematic Skeleton Fitting}
\label{sec:kinematic_fitting}
After obtaining the 2D joint position predictions in the form of heatmaps in image space, and the 3D joint coordinates relative to the root joint, we fit a kinematic skeleton model to this data. This ensures that the resulting hand pose is anatomically plausible, while at the same time it allows to retrieve the \emph{absolute} hand pose, as we will describe below. Moreover, when processing a sequence of images, \ie performing hand \emph{tracking}, we can additionally impose temporal smoothness.

\parahead{Kinematic Hand Model}
Our kinematic hand model is illustrated as \emph{Skeleton Fitting} block in Fig.~\ref{fig:pipeline}. The model comprises one root joint (the wrist) and $20$ finger joints, resulting in a total number of $21$ joints.
Note, that this number includes the finger tips as joints without any degree-of-freedom.
Let $\mathbf{t} \in \R^{3}$ and $\mathbf{R} \in \SO(3)$ (represented in Euler angles) be the global position and rotation of the root joint, and $\theta \in 
\R^{20}$ be the hand articulation angles of the $15$ finger joints with one or two degrees-of-freedom. We stack all parameters into $\Theta = (\mathbf{t},\mathbf{R},\theta)$.
By $\mathcal{M}(\Theta) \in \R^{J \times 3}$ we denote the \emph{absolute} 3D positions of all $J{=}21$ hand joints (including the root joint and finger tips), where we use $\mathcal{M}_j(\Theta) \in \R^3$ to denote the position of the $j$-th joint. In order to compute the position for non-root joints, a traversal of the kinematic tree is conducted.
To account for bone length variability across different users, we perform a \emph{per-user skeleton adaptation}. The user-specific bone lengths are obtained by averaging  relative bone lengths of the 2D prediction over 30 frames while the users are asked to hold their hand parallel to the camera image plane. 

For fitting the kinematic model to data, we minimize the energy
\begin{align}\label{eq:energy}
E(\Theta) = E_{\text{2D}}(\Theta) {+} E_{\text{3D}}(\Theta) {+} E_{\text{limits}}(\Theta) {+} E_{\text{temp}}(\Theta) \,,
\end{align}
where the individual energy terms are described below.

\parahead{2D Fitting Term} The purpose of the term $E_{\text{2D}}$ is to minimize the distance between the hand joint position projected onto the image plane and the heatmap maxima. It is given by
\begin{align}
E_{\text{2D}}(\Theta) = \sum_{j} \omega_j \| \Pi(\mathcal{M}_j(\Theta))) - u_j\|_2^2 \,,
\end{align}
where $u_j \in \R^2$ denotes the heatmap maxima of the $j$-th joint, $\omega_j > 0$ is a scalar confidence weight derived from the heatmap, and $\Pi: \R^3 \mapsto \R^2$ is the projection from 3D space to the 2D image plane, which is based on the camera intrinsics. Note that this 2D fitting term is essential in order to retrieve the \emph{absolute} 3D joint positions since the 3D fitting term $E_{\text{3D}}$ takes only root-relative articulation into account, as described next.

\parahead{3D Fitting Term}
The term $E_{\text{3D}}$ has the purpose to obtain a good hand articulation by using the predicted \emph{relative} 3D joint positions. Moreover, this term resolves depth ambiguities that are present when using 2D joint positions only. We define $E_{\text{3D}}$ as
\begin{align}
E_{\text{3D}}(\Theta) =\sum_{j}  \| (\mathcal{M}_j(\Theta) - \mathcal{M}_{\text{root}}(\Theta)) - z_j\|_2^2 \,.
\end{align}
The variable $z_j \in \R^3$ is the user-specific position of the $j$-th joint relative to the root joint, which is computed from the output of the \reg{}, $x_j$, as 
\begin{align}
\label{eq:skeletonNorm}
z_j = z_{p(j)} + \frac{\|\mathcal{M}_j(\Theta) - \mathcal{M}_{p(j)}(\Theta) \|_2}{\|x_j - x_{p(j)} \|_2} (x_j - x_{p(j)}) \,,
\end{align}
where $p(j)$ is the parent of the joint $j$ and we set $z_{\text{root}} = \mathbf{0} \in \R^3$.
The idea of using user-specific positions is to avoid mispredictions caused by bone length inconsistencies in the hand model by normalizing the predictions to comply with the model.

\parahead{Joint Angle Constraints}
The term $E_{\text{limits}}$ penalizes anatomically implausible hand articulations by enforcing that joints do not bend too far. Mathematically, we define
\begin{equation}
E_{\text{limits}}(\theta) = \| \max(\begin{bmatrix}
\mathbf{0}, \theta{-}\theta^{\max}, \theta^{\min}{-}\theta
\end{bmatrix}) \|^2_2 \,,
\end{equation}
where $\theta^{\max} \in \R^{20}$ and $\theta^{\min} \in \R^{20}$ are the upper and lower joint angle limits for the degrees-of-freedom of the non-root joints, and the function $\max: \R^{20 \times 3} \mapsto \R^{20}$ computes the row-wise maximum.

\parahead{Temporal Smoothness} The term $E_{\text{temp}}$ enforces temporal smoothness for hand pose tracking by making sure that the current velocity of the change in $\Theta$ is close to the velocity in the previous time step. We formulate
\begin{align}
E_{\text{temp}}(\Theta) = \| (\nabla \Theta^{\text{prev}} - \nabla \Theta)\|^2_2\,,
\end{align}
where the gradients of the pose parameters $\Theta$ are determined using finite (backward) differences.  

\parahead{Optimization}
In order to minimize the energy in \eqref{eq:energy} we use a gradient-descent strategy. For the first frame, $\theta$ and $\mathbf{t}$ are initialized to represent a neutral hand pose (an open hand) that is centered in the image and 45cm away from the camera plane. 
For the remaining frames we use the translation and articulation parameters $\mathbf{t}$ and $\theta$ from the previous frame as initialization.
In our experiments we have found that global hand rotations are problematic for the kinematic model fitting as they may lead to unwanted local minima due to the non-convexity of energy~\eqref{eq:energy}. In order to deal with this problem, for the global rotation $\mathbf{R}$ we do not rely on the previous value $\mathbf{R}^{\text{prev}}$, but instead initialize it based on the relative 3D joint predictions. Specifically, we make use of the observation that in the human hand the root joint and its four direct children joints of the non-thumb fingers (the respective MCP joints) are (approximately) rigid (cf.~Fig.~\ref{fig:pipeline}, \emph{Skeleton Fitting} block). Thus, to find the global rotation $\mathbf{R}$ we solve the problem 
\begin{align}\label{eq:rotInit}
\min_{\mathbf{R} \in \SO(3)} \|\mathbf{R} \bar{Z} - \tilde{Z} \|_F^2\,,
\end{align}
where $\bar{Z}$ contains (fixed) direction vectors derived from the \emph{hand model}, and $\tilde{Z}$ contains the corresponding direction vectors that are derived from the current \reg{} \emph{predictions}.
Both have the form $Z = \begin{bmatrix}
y_{j_1}, y_{j_2}, y_{j_3}, y_{j_4}, n
\end{bmatrix} \in \R^{3 \times 5}$, where
the $y_{j_k} =  \frac{1}{\|x_{j_k}{-}x_{\text{root}}\|}(x_{j_k}{-}x_{\text{root}}) \in \R^3$ are (normalized) vectors  that point from the root joint to the respective non-thumb MCP joints $j_1,\ldots,j_4$, and $n = y_{j_1} \times y_{j_4}$ is the (approximate) normal vector 
of the ``palm-plane''. To obtain $\bar{Z}$ we compute the $y_j$ based on the $x_j$ of the 3D \emph{model points} in world space, which is done only once for a skeleton at the beginning of the tracking when the global rotation of the model is identity.
To obtain $\tilde{Z}$ in each frame, the $x_j$ are set to the \reg{} \emph{predictions} for computing the $y_j$. While problem~\eqref{eq:rotInit} is non-convex, it still admits the efficient computation of a global minimum as it is an instance of the \emph{Orthogonal Procrustes Problem} \cite{Schonemann:1966ch,TenBerge:2006ih}: for $U\Sigma V^T$ being the singular value decomposition of $\tilde{Z} \bar{Z}^T \in \R^{3 \times 3}$, the global optimum of \eqref{eq:rotInit} is given by $\mathbf{R} = U \operatorname{diag}(1,1,\operatorname{det}(UV^T)) V^T$.

%% file: experiments.tex
\section{Experiments}\label{sec:experiments}
 
\begin{figure*}[t] 
   \begin{subfigure}[t]{0.32\textwidth}
  \includegraphics[width=\columnwidth,trim={3.8cm 8.5cm 4.5cm 9.3cm},clip]{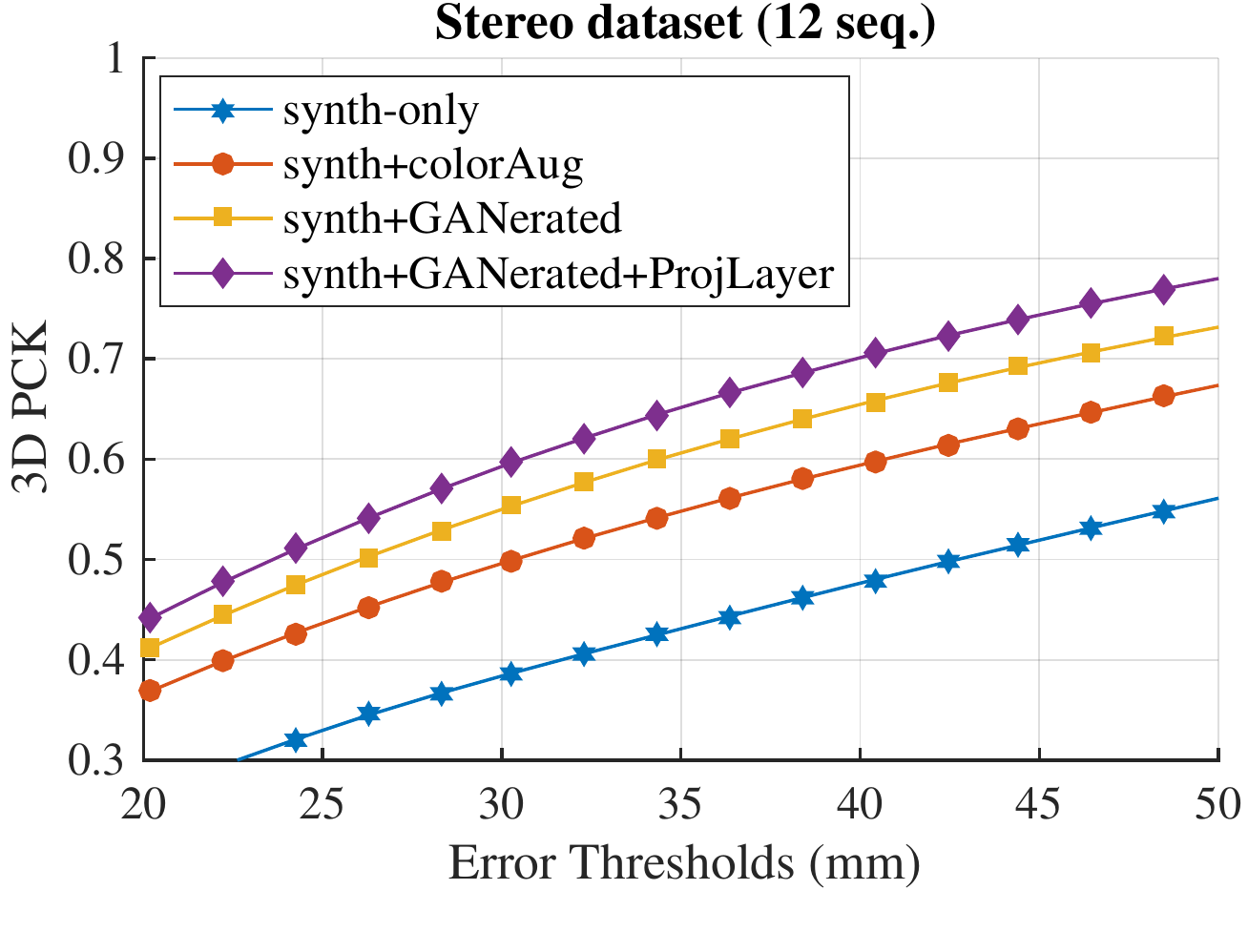}
  \vspace{-7mm} 
  \caption{3D PCK on Stereo dataset~\cite{zhang2016stereo}. Using GANerated images (yellow) outperforms using only synthetic images (blue). Our projection layer (purple) to enforce 2D/3D consistency further improves accuracy.} 
  \label{fig:jointregressionwithsyntheticimages}
	\end{subfigure}
    \hspace{0.25cm}
    \begin{subfigure}[t]{0.32\textwidth}
    \includegraphics[width=\columnwidth,trim={3.8cm 8.5cm 4.5cm 9.3cm},clip]{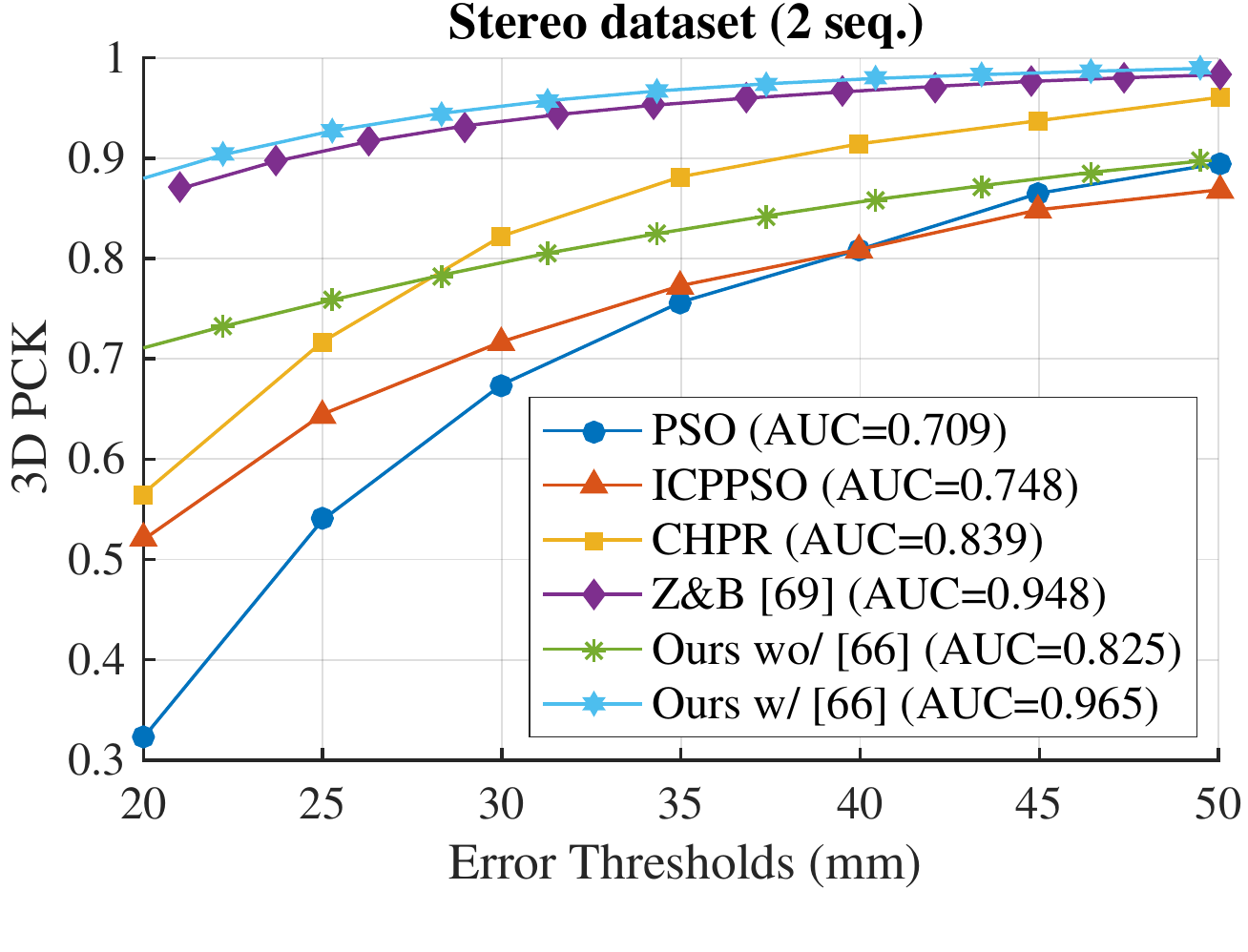}
    \vspace{-7 mm} 
     \caption{3D PCK on the Stereo dataset~\cite{zhang2016stereo}. We use 10 sequences for training and 2 for testing, as in \cite{zhang2016stereo,Zimmermann:2017um}. Our method (light blue) achieves the best results. } 
  \label{fig:stereo_vsZimmermann}
	\end{subfigure}
   \hspace{0.25cm} 
       \begin{subfigure}[t]{0.32\textwidth}
\includegraphics[width=\columnwidth,trim={3.8cm 8.5cm 4.5cm 9.3cm},clip]{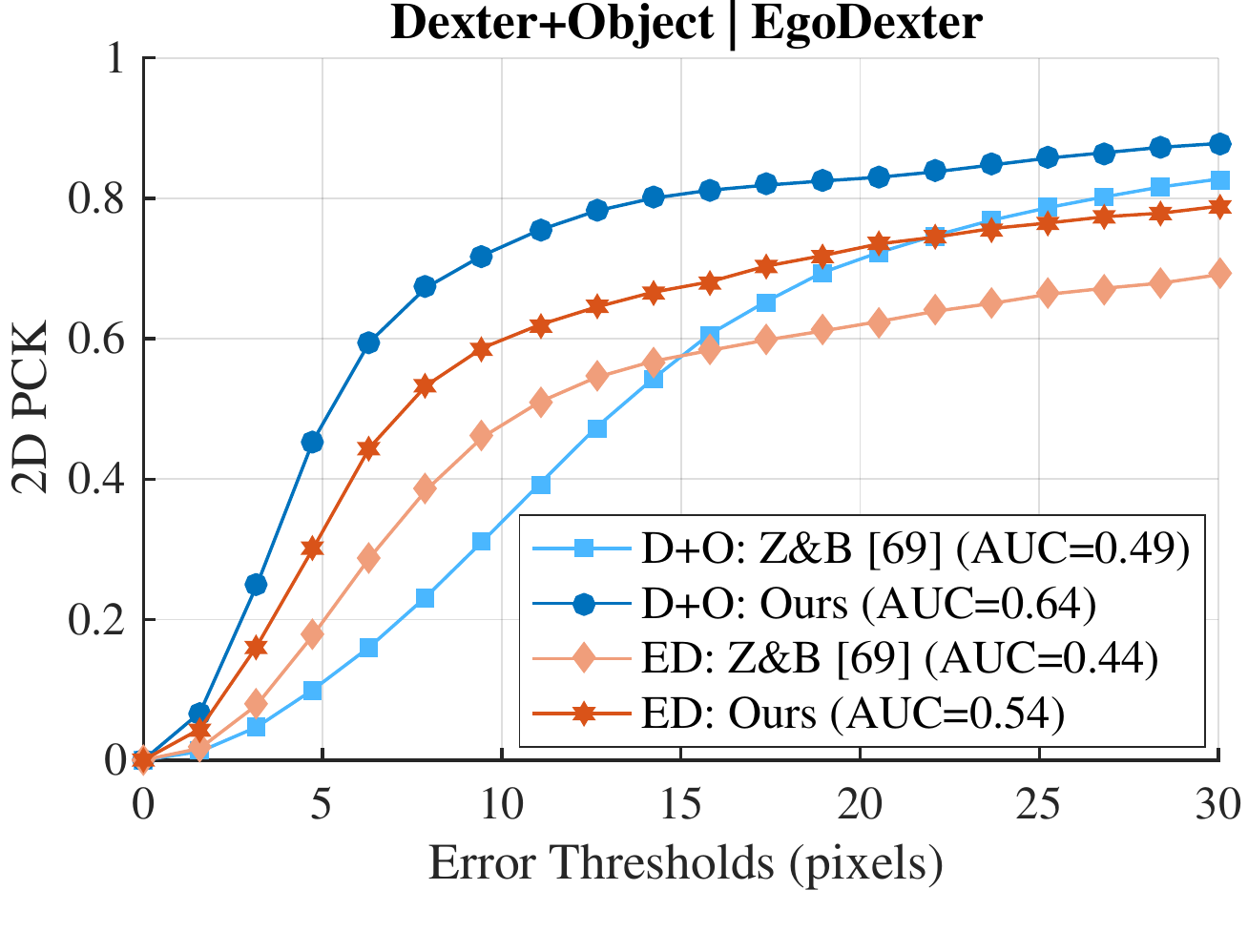} 
\vspace{-7mm} 
    \caption{2D PCK on the Dexter+Object (D+O) dataset~\cite{sridhar_eccv2016} and EgoDexter (ED)~\cite{mueller_iccv2017}. Our method (saturated blue/red) outperforms Z\&B \cite{Zimmermann:2017um} on both datasets. 
   } 
  \label{fig:dexter+object_vsZimmermann}
	\end{subfigure}
  \caption{Quantitative evaluation. (a) Ablative study on different training options. (b), (c) 3D and 2D PCK comparison with state-of-the-art methods on publicly available datasets.}
\end{figure*} 

We quantitatively and qualitatively  evaluate our method and compare our results with other state-of-the-art methods on a variety of publicly available datasets.
For that, we use the Percentage of Correct Keypoints (PCK) score, a popular criterion to evaluate pose estimation accuracy. 
PCK defines a candidate keypoint to be correct if it falls within a circle (2D) or sphere (3D) of given radius around the ground truth.

\subsection{Quantitative Evaluation} \label{sec:quantitative-evaluation}
\parahead{Ablative study}
In Fig.~\ref{fig:jointregressionwithsyntheticimages} we compare the detection accuracy when training our hand joint regression network \reg{} with different configurations of training data. Specifically, we compare the cases of using synthetic images only, synthetic images plus color augmentation, and synthetic images in combination with GANerated images, where for the latter we also considered the case of additionally using the \emph{ProjLayer} in \reg.
While we evaluated the \reg{} on the entire Stereo dataset \cite{zhang2016stereo} comprising 12 sequences, we did \textit{not train on any} frame of the Stereo dataset for this test. 
We show that training on purely synthetic data leads to poor accuracy (3D PCK@50mm $\approx 0.55$).
While color augmentation on synthetic images improves the results, our GANerated images significantly outperform standard augmentation techniques, achieving a 3D PCK@50mm $\approx 0.80$.
This test validates the argument for using GANerated images for training.

\parahead{Comparison to state-of-the-art}
Fig.~\ref{fig:stereo_vsZimmermann} evaluates our detection accuracy on the Stereo dataset, and compares it to existing methods \cite{zhang2016stereo,Zimmermann:2017um}. 
We followed the same evaluation protocol used in \cite{Zimmermann:2017um}, \ie we train on 10 sequences and test it on the other 2.
Our method outperforms all existing methods. Additionally, we test our approach \textit{without} training on any sequence of the Stereo dataset, and demonstrate that we still outperform some of the existing works (green line in Fig.~\ref{fig:stereo_vsZimmermann}). This demonstrates the generalization of our approach. 

Figure \ref{fig:dexter+object_vsZimmermann} shows the 2D PCK, in pixels, on the Dexter+Object \cite{sridhar_eccv2016} and EgoDexter \cite{mueller_iccv2017} datasets. We significantly outperform Zimmerman and Brox (Z\&B) \cite{Zimmermann:2017um}, which fails under difficult occlusions. 
Note that here we cannot report 3D PCK because \cite{Zimmermann:2017um} only outputs root-relative 3D, and these datasets do not have root joint annotations.

\subsection{Qualitative Evaluation} \label{sec:qualitative-evaluation}

\begin{figure*}[t] 
	\centerline{ 
		\includegraphics[width=\linewidth,trim={0cm 0.5cm 0cm 0cm},clip,page=1]{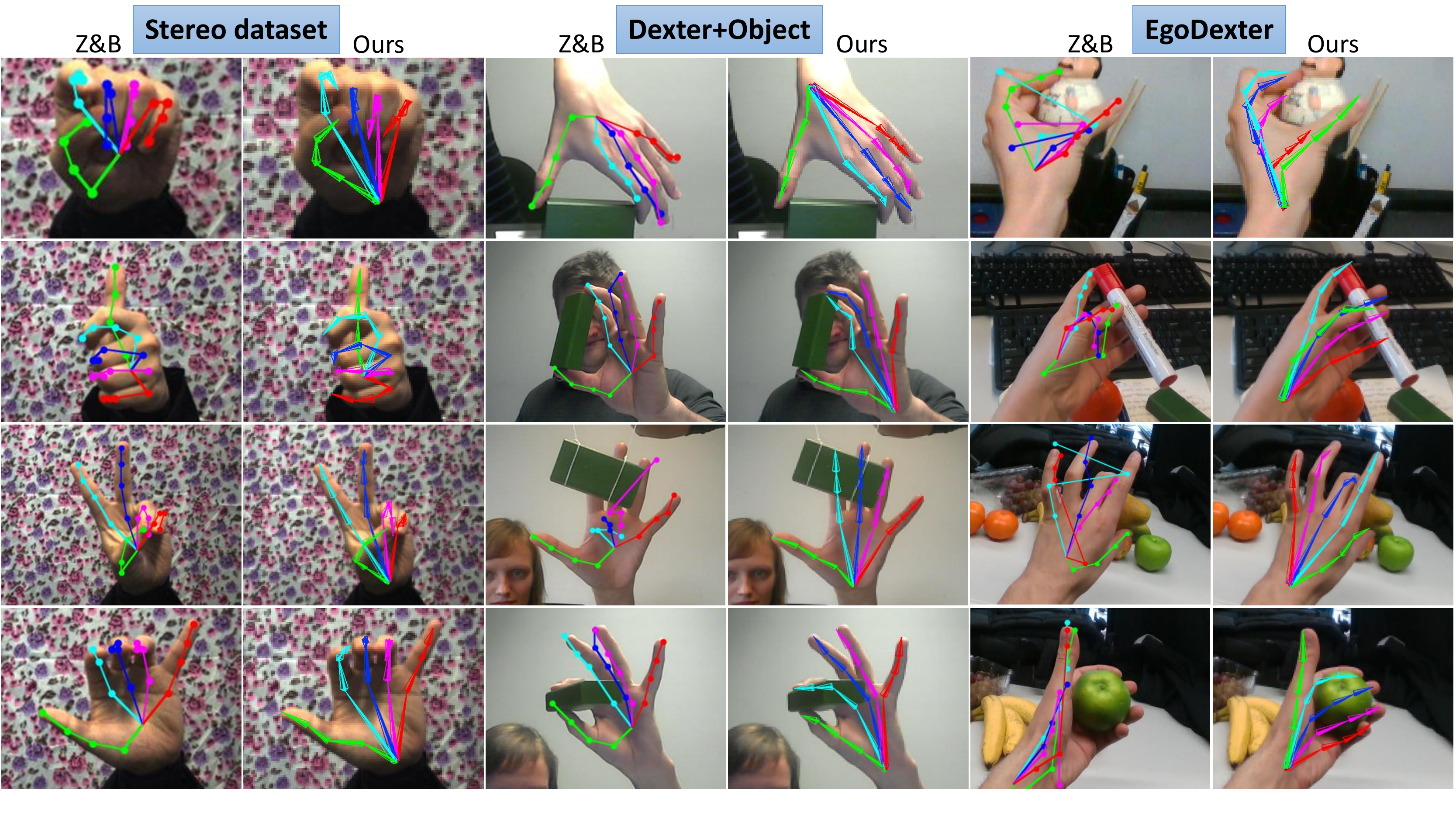} 
	}
	\caption{We compare our results with Zimmermann and Brox \cite{Zimmermann:2017um} on three different datasets. Our method is more robust in cluttered scenes and it even correctly retrieves the hand articulation when fingers are hidden behind objects.} 
	\label{fig:qual_comp}
\end{figure*}

\begin{figure*}[t] 
	\centerline{ 
		\includegraphics[width=\linewidth]{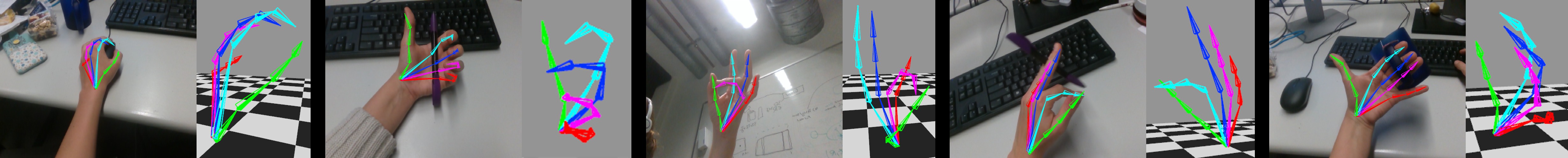} 
	}
	\caption{Representative frames of a live sequence captured with an off-the-shelf RGB webcam and tracked in real-time. Our method successfully recovers the global 3D positions of the hand joints. Here, for each input frame, we show the 3D tracked result projected to the camera plane, and the recovered 3D articulated hand skeleton visualized from a different viewpoint.} 
	\label{fig:qual_comp_real_time}
	\vspace{-0.3cm}
\end{figure*}

We qualitatively evaluate our method on three different video sources: publicly available datasets, real-time capture, and community (or vintage) video (\ie YouTube). 

Fig.~\ref{fig:qual_comp} presents qualitative results on three datasets, Stereo \cite{zhang2016stereo}, Dexter+Object \cite{sridhar_eccv2016} and EgoDexter \cite{mueller_iccv2017}, for both Z\&B~\cite{Zimmermann:2017um} and our method. We show that our method provides robust tracking of the hand even in severely occluded situations, and significantly outperforms \cite{Zimmermann:2017um} in these cases.
While we already outperformed Z\&B~\cite{Zimmermann:2017um} in our quantitative evaluation (Fig. \ref{fig:dexter+object_vsZimmermann}), we emphasize that this is not the full picture, since the datasets from~\cite{sridhar_eccv2016,mueller_iccv2017} only provide annotations for \emph{visible} finger tips due to the manual annotation process. Thus, the error due to occluded joints is not at all reflected in the quantitative analysis. Since our method is explicitly trained to deal with occlusion--in contrast to \cite{Zimmermann:2017um}--our qualitative analysis in the supplementary video and in columns 3--6 of Fig. \ref{fig:qual_comp} highlights the superiority of our method in such scenarios. 

We show real-time tracking results in Fig.~\ref{fig:qual_comp_real_time} as well as in the supplementary video.
This sequence was tracked live 
with a regular desktop webcam in a standard office environment. 
Note how our method accurately recovers the full 3D articulated pose of the hand.
In Fig.~\ref{fig:teaser} we demonstrate that our method is also compatible with community or vintage RGB-only video. In particular, we show 3D hand tracking in YouTube videos, which demonstrates the generalization of our method.

%% file: conclusion.tex
\section{Limitations \& Discussion}
One difficult scenario for our method is when the background has similar appearance as the hands, as our \reg{} struggles to obtain good predictions and thus tracking becomes unstable. 
This can potentially be addressed by using an explicit segmentation stage, similar to Zimmermann and Brox~\cite{Zimmermann:2017um}.
Moreover, when two or more hands are close in the input image, detections may also become unreliable.
While our approach can handle sufficiently separate hands ---due to our bounding box tracker--- tracking of interacting hands, or hands of multiple persons, is an interesting direction for follow-up work.

The 3D tracking of hands in purely 2D images is an extremely challenging problem. 
While our real-time method for 3D hand tracking outperforms state-of-the-art RGB-only methods, there is still an accuracy gap between our results and existing RGB-D methods (mean error of ${\approx}5\text{cm}$ for our proposed RGB approach vs. ${\approx}2\text{cm}$ for the RGB-D method of \cite{sridhar_eccv2016} on their dataset Dexter+Object).
Nevertheless, we believe that our method is an important step towards democratizing RGB-only 3D hand tracking.

\section{Conclusion}
Most existing works consider either the problem of 2D hand tracking from monocular RGB images, or they use additional inputs, such as depth images or multi-view RGB, to track the hand motions in 3D. 
While the recent method by Zimmermann and Brox~\cite{Zimmermann:2017um} considers monocular 3D hand tracking from RGB images, our proposed approach tackles the same problem but goes one step ahead with regards to several dimensions: we obtain the \emph{absolute} 3D hand pose due to our model fitting procedure, our approach is \emph{more robust} to occlusions, and our method \emph{generalizes better} due to enrichment of our synthetic dataset such that it resembles the distribution of real hand images. 
Our experimental evaluation demonstrates these benefits as our method significantly outperforms the method by Zimmermann and Brox, particularly in difficult occlusion scenarios. 
In order to further encourage future work on monocular 3D RGB hand tracking we make our dataset available to the research community.

%% file: supp.tex
\appendix
{
	\noindent
	\Large
	\textbf{Appendix}
}
\vspace{0.3cm}

%%%%%%%%% BODY TEXT
In this appendix we provide details of the \reg{} and \gan{} networks (Sec.~\ref{sec:network_details}), additional quantitative evaluations (Sec.~\ref{sec:quantitative}), as well as detailed visualizations of our CNN \reg{} output and final results (Sec.~\ref{sec:qualitative})
\section{CNN and GAN Details}
\label{sec:network_details}
\subsection{\textbf{\gan} network}
\parahead{Network Design}
The architecture of \gan~is based on the CycleGAN~\cite{CycleGAN2017}, \ie we train two conditional generator and two discriminator networks for synthetic and real images, respectively.
Recently, also methods using only one generator and discriminator for enrichment of synthetic images from unpaired data have been proposed.
Shrivastava \etal~\cite{shrivastava2017learning} and Liu \etal~\cite{liu2017learning} both employ an L1 loss between the conditional synthetic input and the generated output (in addition to the common discriminator loss) due to the lack of image pairs.
This loss forces the generated image to be similar to the synthetic image in all aspects, \ie it might hinder the generator in producing realistic outputs if the synthetic data is not already close.
Instead, we decided to use the combination of cycle-consistency and geometric consistency loss to enable the generator networks to move farther from the synthetic data thus approaching the distribution of real world data more closely while preserving the pose of the hand.
Our \gan~contains \emph{ResNet} generator and Least Squares \emph{PatchGAN} discriminator networks.

\parahead{Training Details}
We train \gan~in Tensorflow~\cite{abadi2016tensorflow} for 20,000 iterations with a batch size of 8.
We initialize the Adam optimizer~\cite{kingma2014adam} with a learning rate of $0.0002$, $\beta_1 = 0.5$, and $\beta_2 = 0.999$.

\subsection{\textbf{\reg} network}

\parahead{Projection Layer \textit{ProjLayer}}
Recent work in 3D body pose estimation has integrated projection layers to leverage 2D-only annotated data for training 3D pose prediction~\cite{brau20163d}.
Since our training dataset provides perfect 3D ground truth, we employ our projection layer merely as refinement module to link the 2D and 3D predictions.
We project the intermediate relative 3D joint position prediction using orthographic projection where the origin of the 3D predictions (the middle MCP joint) projects onto the center of the rendered heatmap.
Hence, our rendered heatmaps are also relative and not necessarily in pixel-correspondence with the ground truth 2D heatmaps.
Therefore, we apply further processing to the rendered heatmaps before feeding them back into the main network branch. 
Note that the rendered heatmaps are differentiable with respect to the 3D predictions which makes backpropagation of gradients through our \emph{ProjLayer} possible.

\begin{figure}[t] 
	\centering
	\includegraphics[width=\columnwidth,trim={5.5cm 10.5cm 6.2cm 10.5cm},clip]{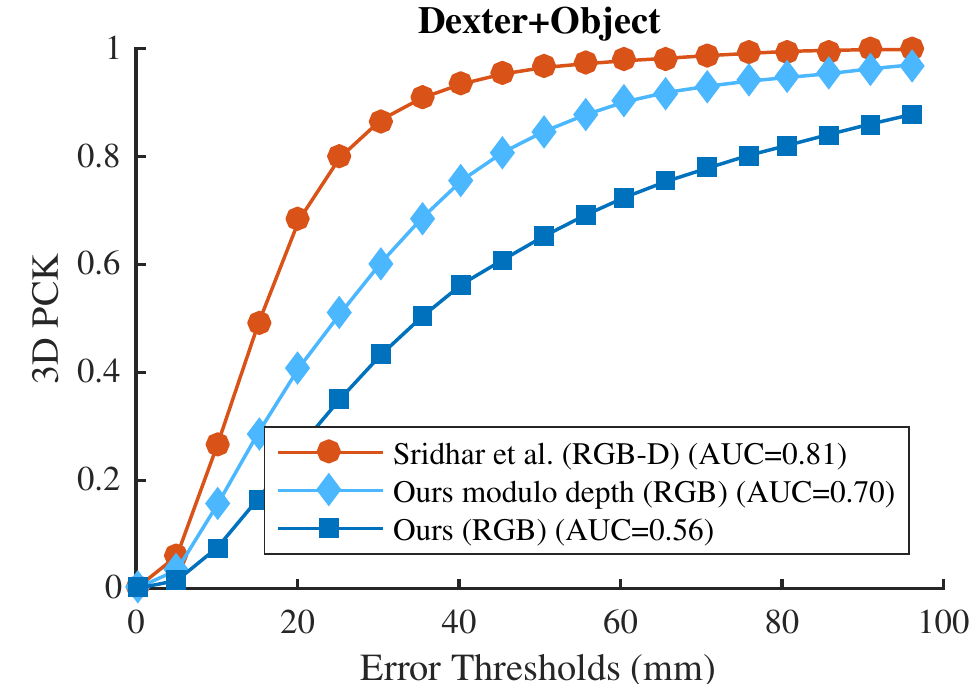} 
	\caption{3D PCK on Dexter+Object. Note that Sridhar \etal~\cite{sridhar_eccv2016} requires RGB-D input, while we use RGB-only.} 
	\label{fig:quantitative_3d}
\end{figure}

{
	\newcommand{\figScale}{0.18\textwidth}
	\begin{figure*}[t]
		\centering
		\resizebox{.82\textwidth}{!}{
		\begin{tikzpicture}
		\node[rotate=90] (textA) at (0,0) {Prediction 2D};
		\node[right of=textA, xshift=1.1cm] (2d_1) 
		{\includegraphics[width=\figScale]{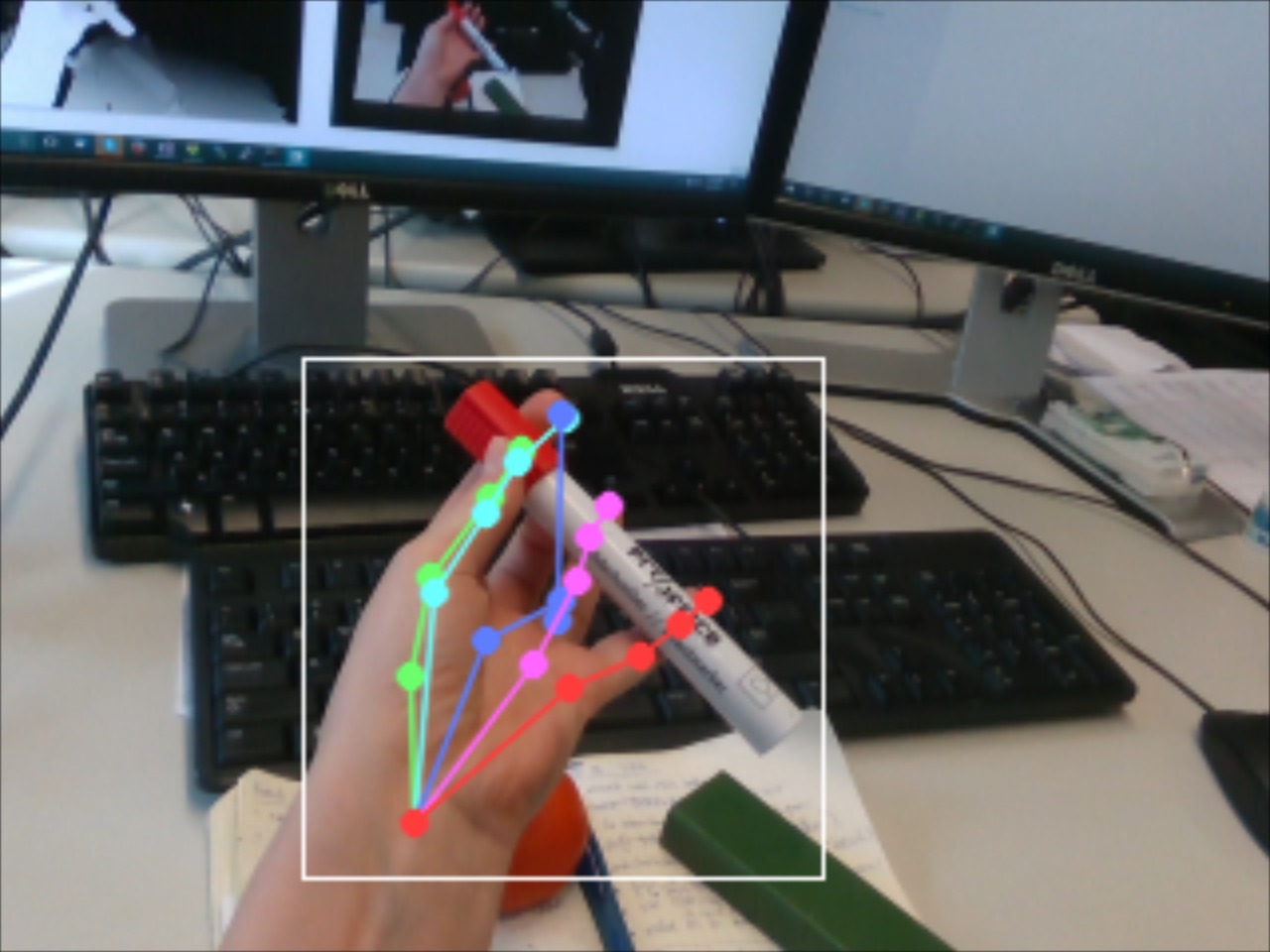}};
		\node[right of=2d_1, xshift=2.3cm] (2d_2) 
		{\includegraphics[width=\figScale]{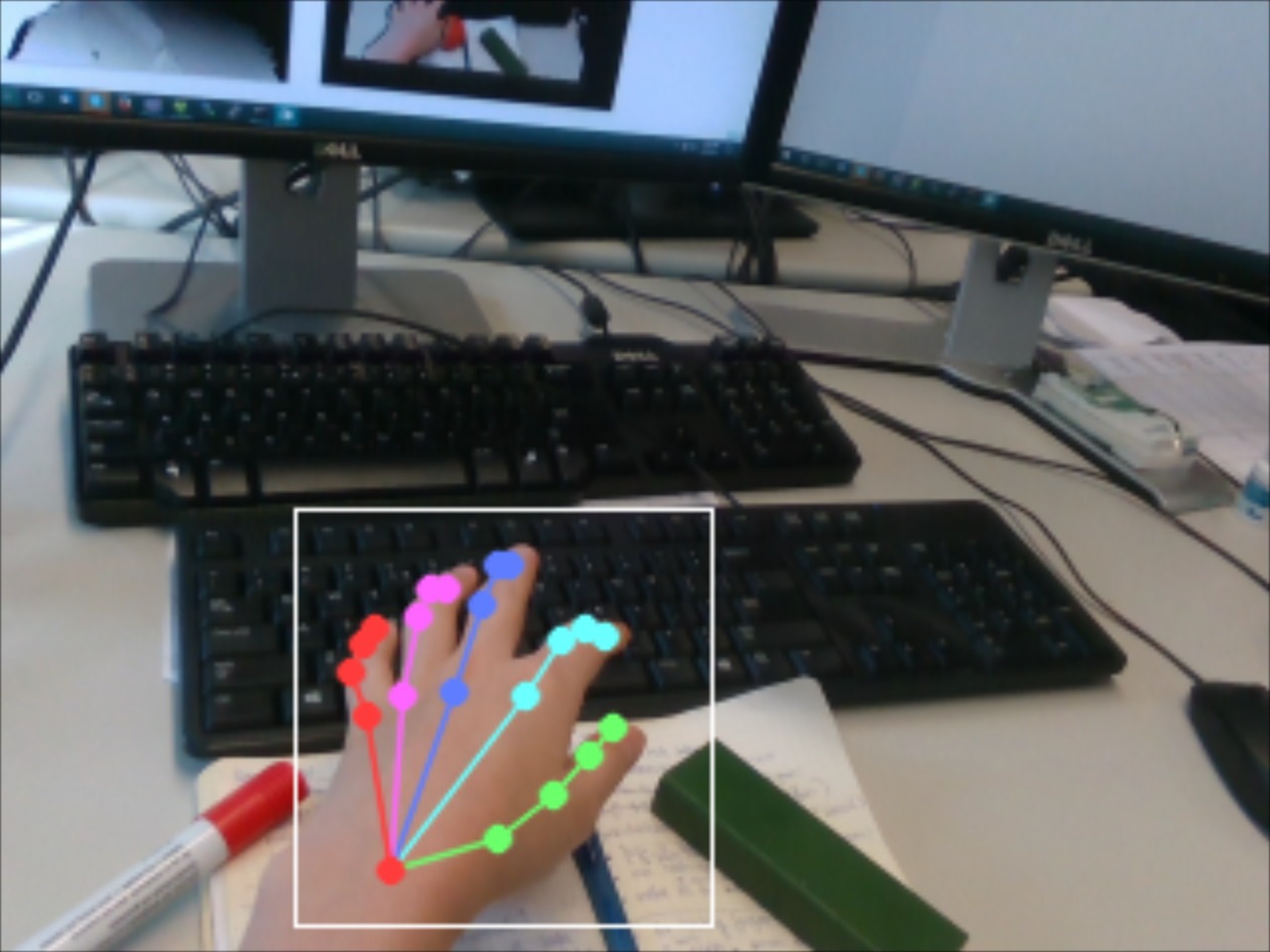}};
		\node[right of=2d_2, xshift=2.3cm] (2d_3) 
		{\includegraphics[width=\figScale]{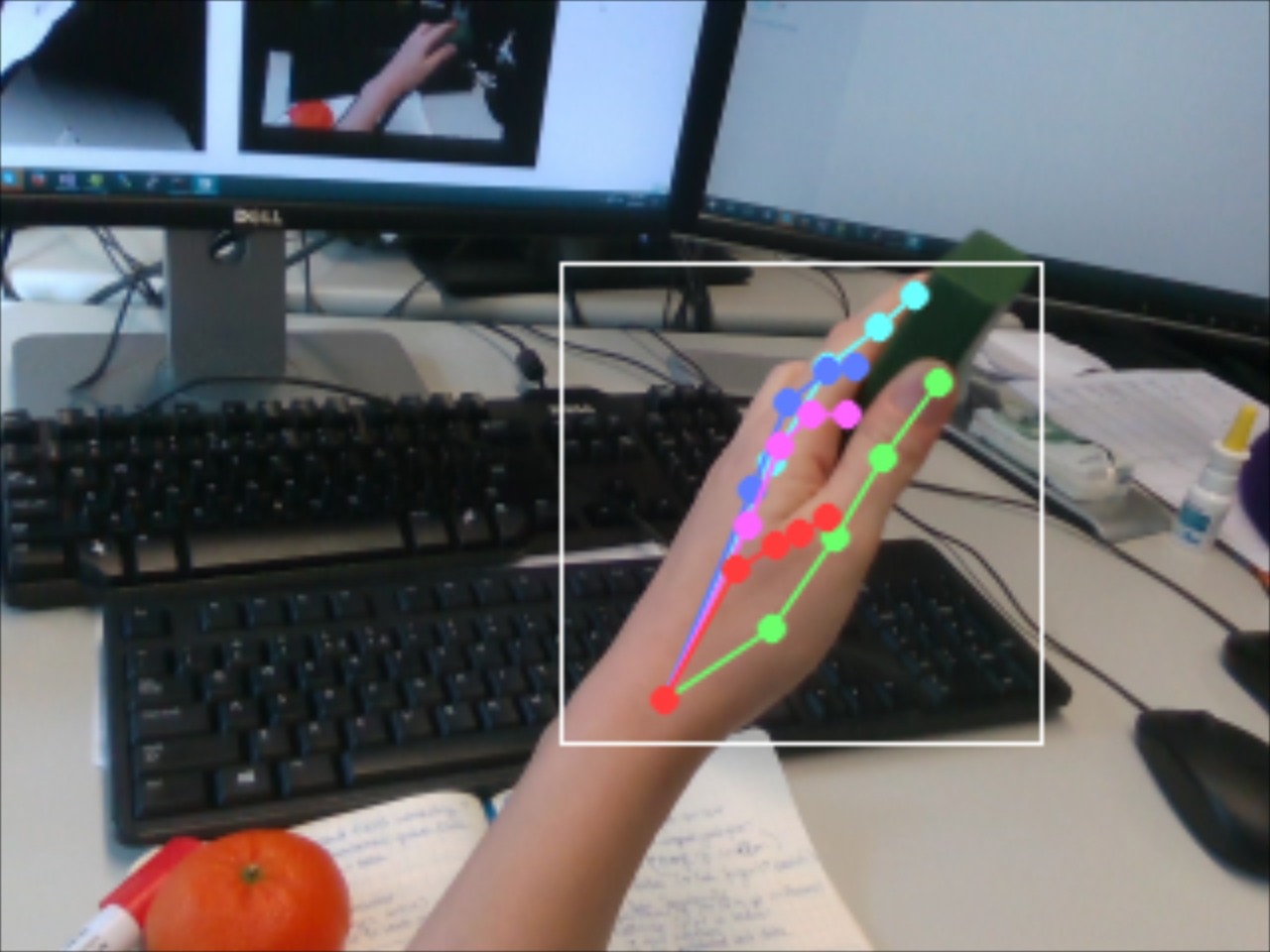}};
		\node[right of=2d_3, xshift=2.3cm] (2d_4) 
		{\includegraphics[width=\figScale]{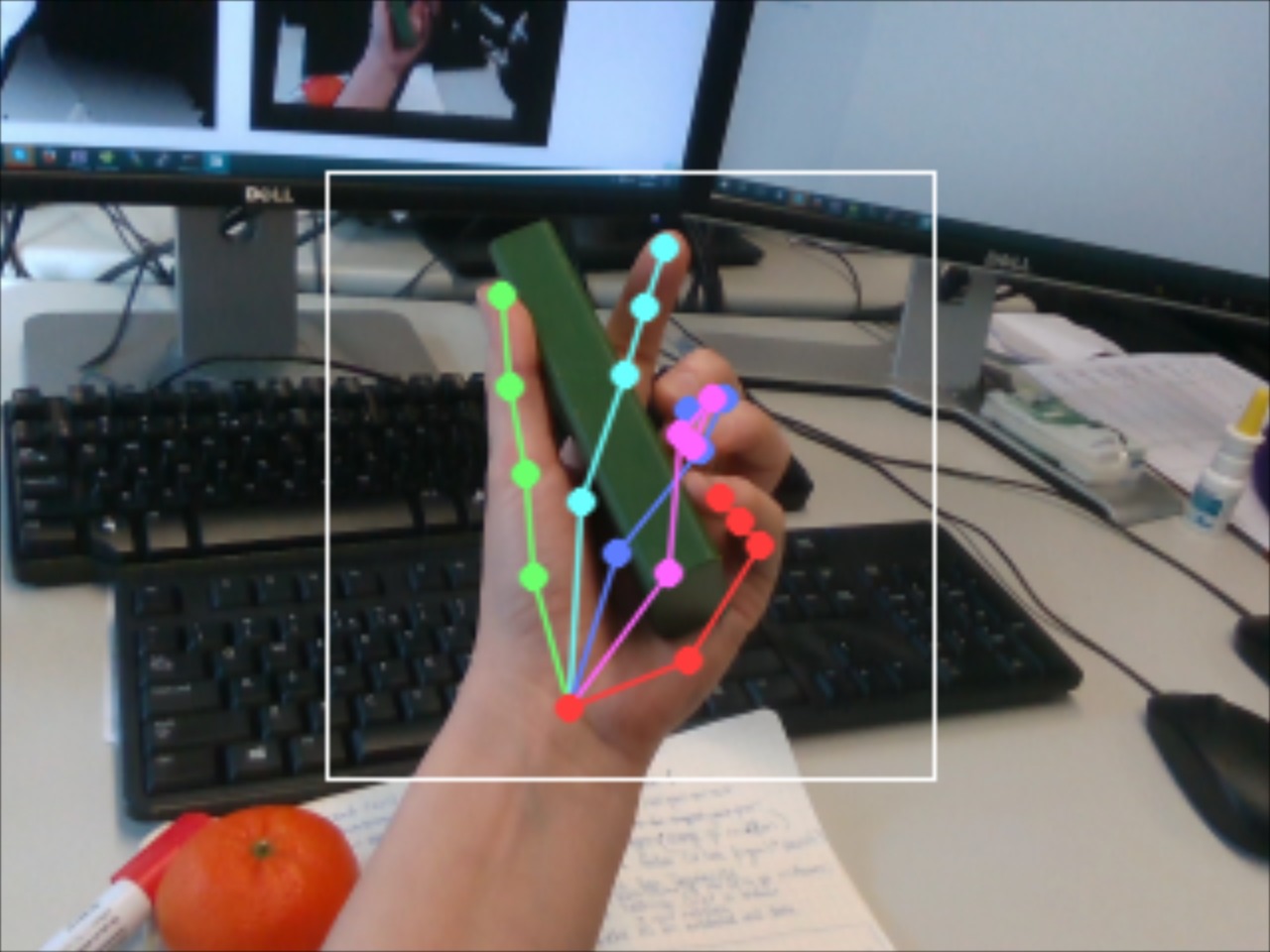}};
		\node[right of=2d_4, xshift=2.3cm] (2d_5) 
		{\includegraphics[width=\figScale]{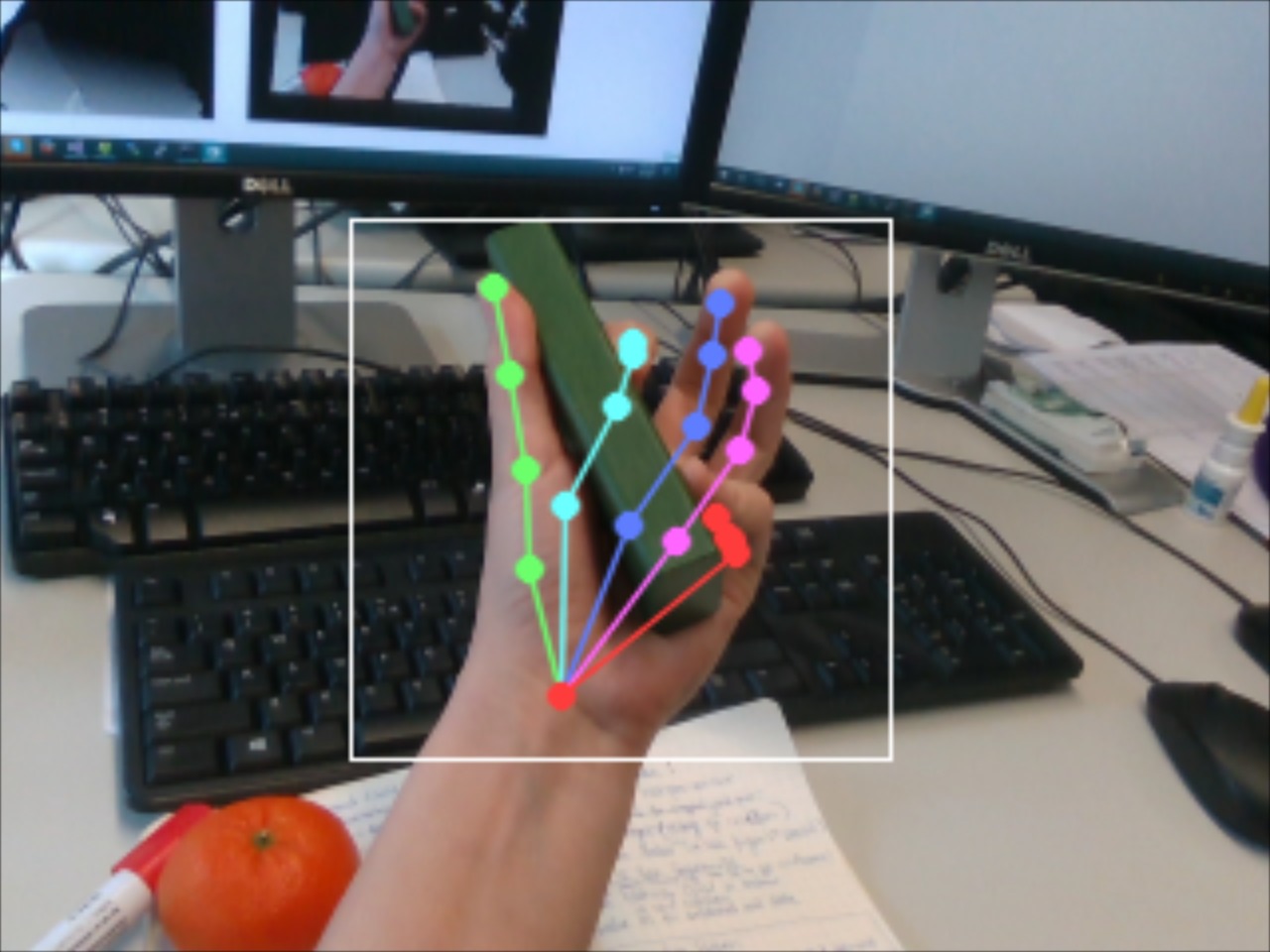}};
		\node[rotate=90, left of=textA, xshift=-1.5cm] (textB)  {Prediction 3D};
		\node[right of=textB,,xshift=1.1cm] (3d_1) 
		{\includegraphics[width=\figScale]{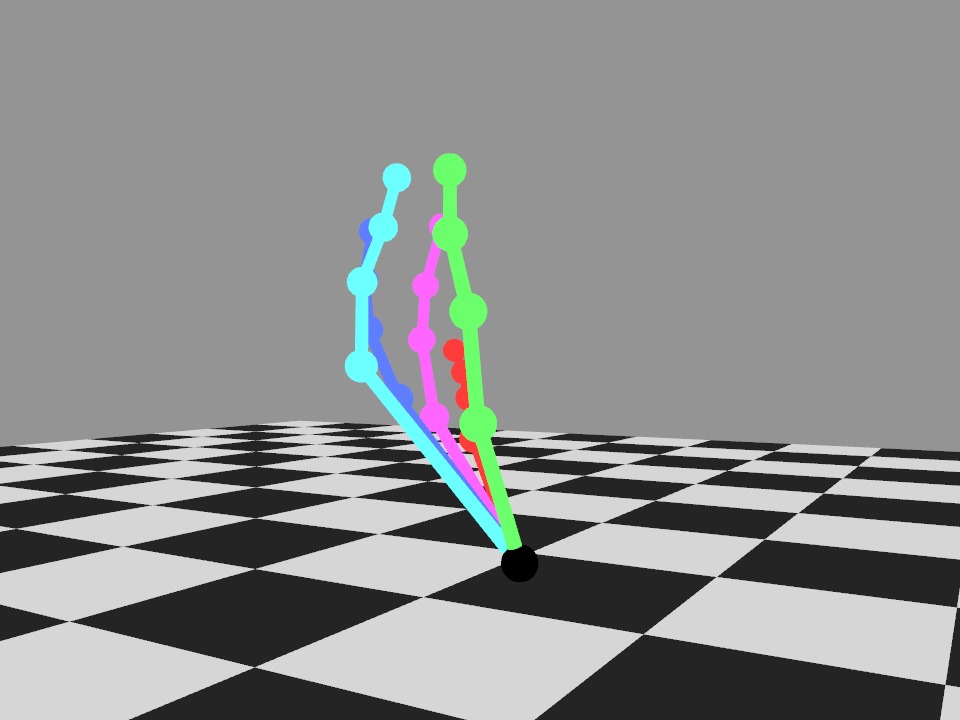}};
		\node[right of=3d_1, xshift=2.3cm] (3d_2) 
		{\includegraphics[width=\figScale]{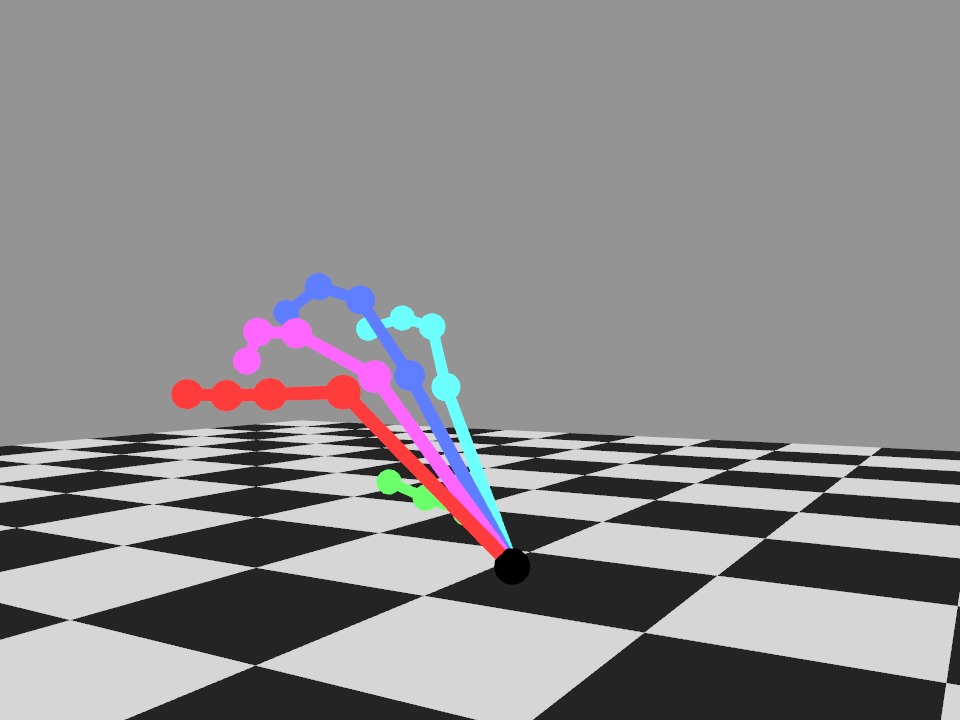}};
		\node[right of=3d_2, xshift=2.3cm] (3d_3) 
		{\includegraphics[width=\figScale]{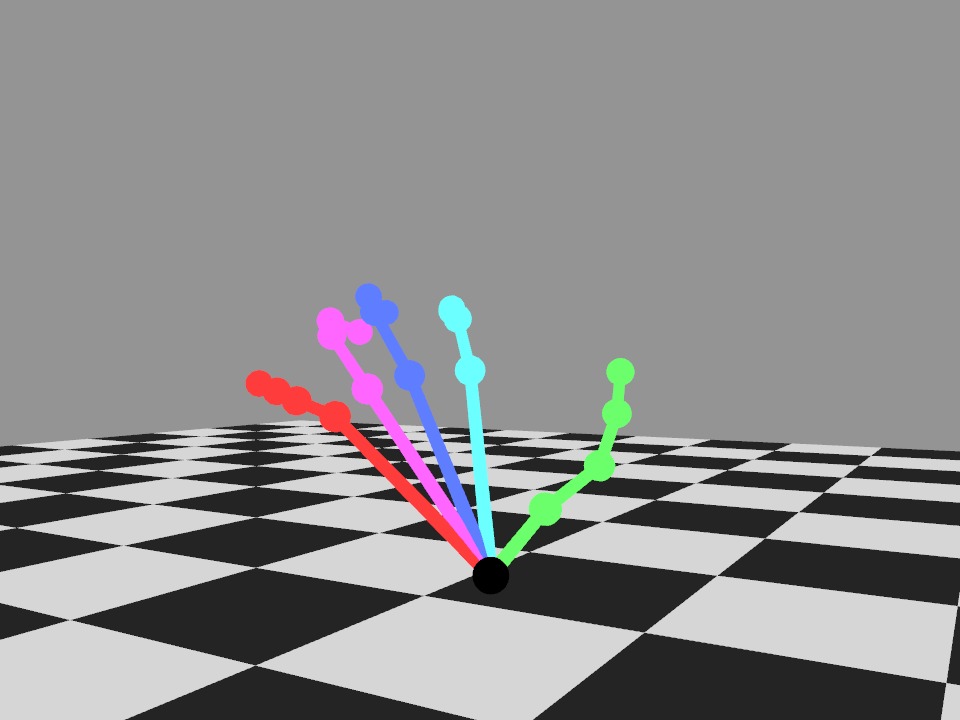}};
		\node[right of=3d_3, xshift=2.3cm] (3d_4) 
		{\includegraphics[width=\figScale]{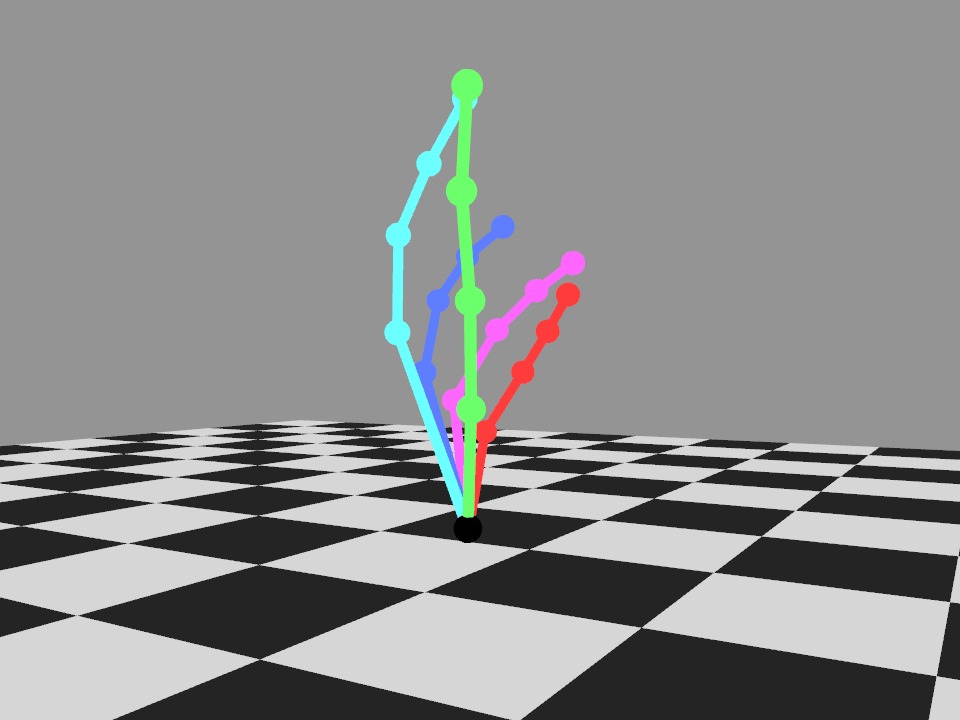}};
		\node[right of=3d_4, xshift=2.3cm] (3d_5) 
		{\includegraphics[width=\figScale]{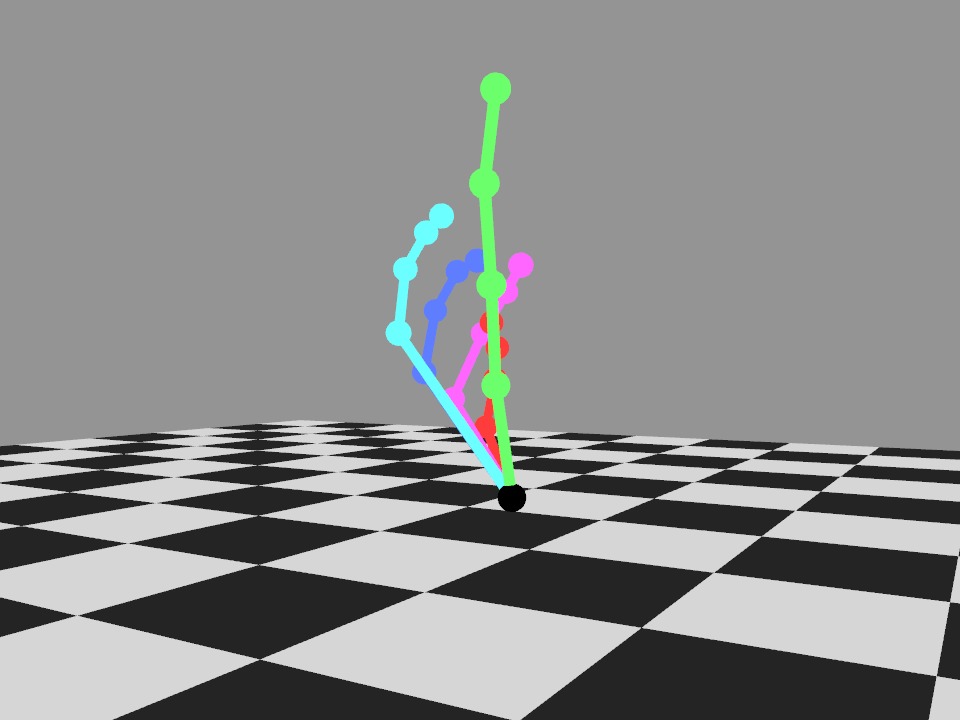}};
		\node[rotate=90, left of=textB, xshift=-1.5cm,yshift=0.3cm] (textC_top)  {Tracked 3D};
		\node[rotate=90, above of=textC_top, yshift=-1.5cm] (textC)  {(projected)};
		\node[right of=textC,xshift=0.9cm] (skel_2d_1) 
		{\includegraphics[width=\figScale]{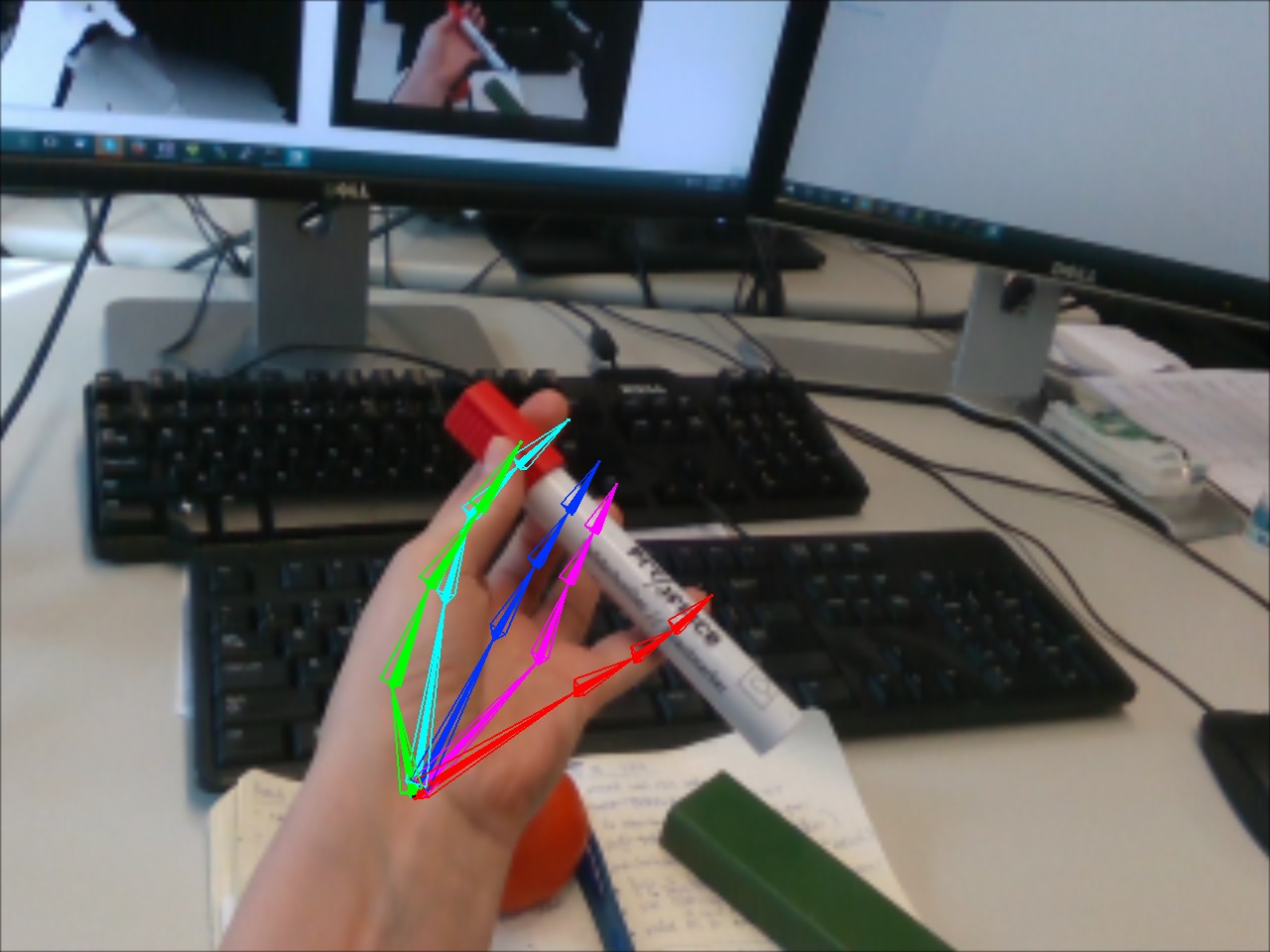}};
		\node[right of=skel_2d_1, xshift=2.3cm] (skel_2d_2) 
		{\includegraphics[width=\figScale]{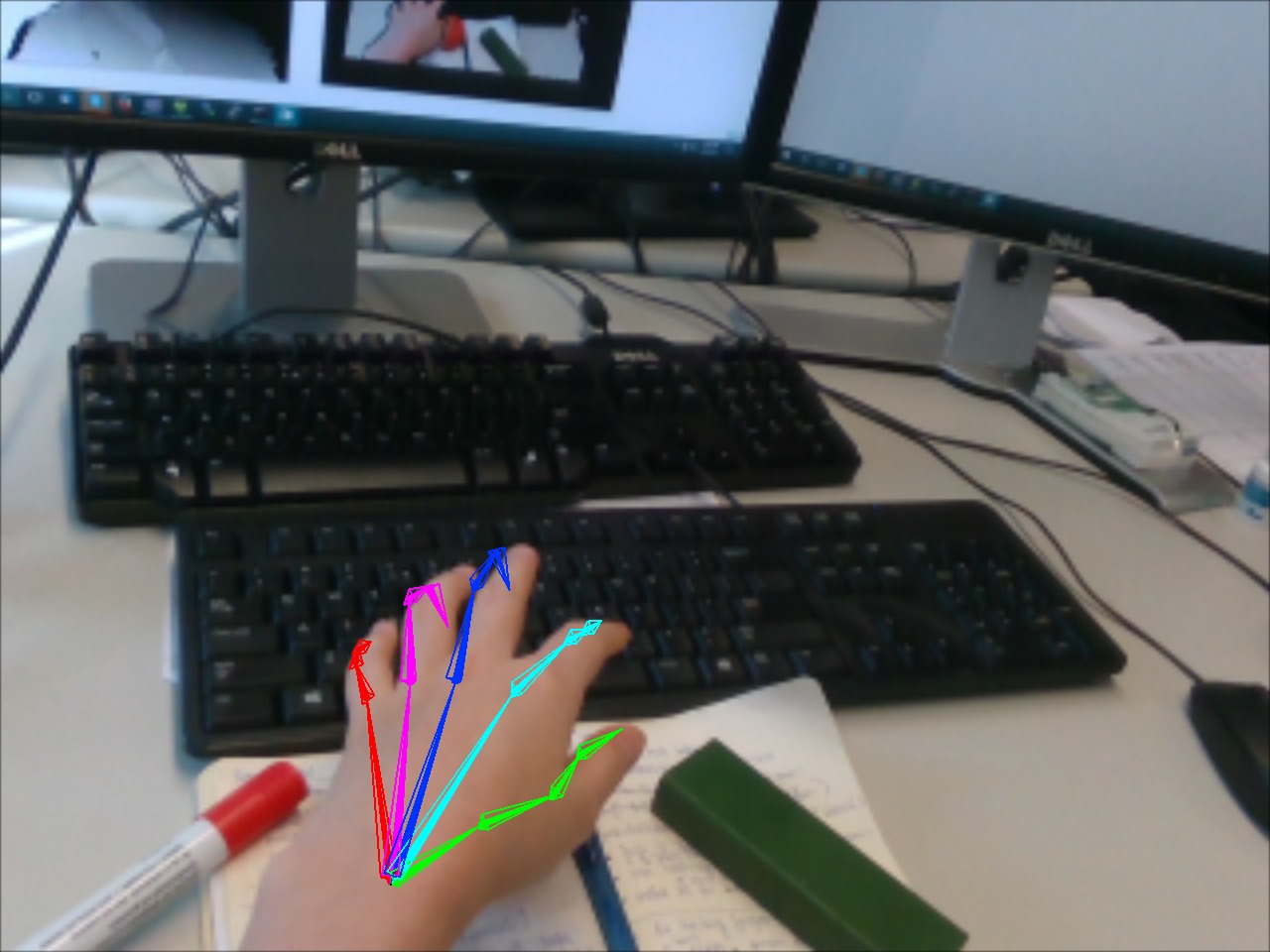}};
		\node[right of=skel_2d_2, xshift=2.3cm] (skel_2d_3) 
		{\includegraphics[width=\figScale]{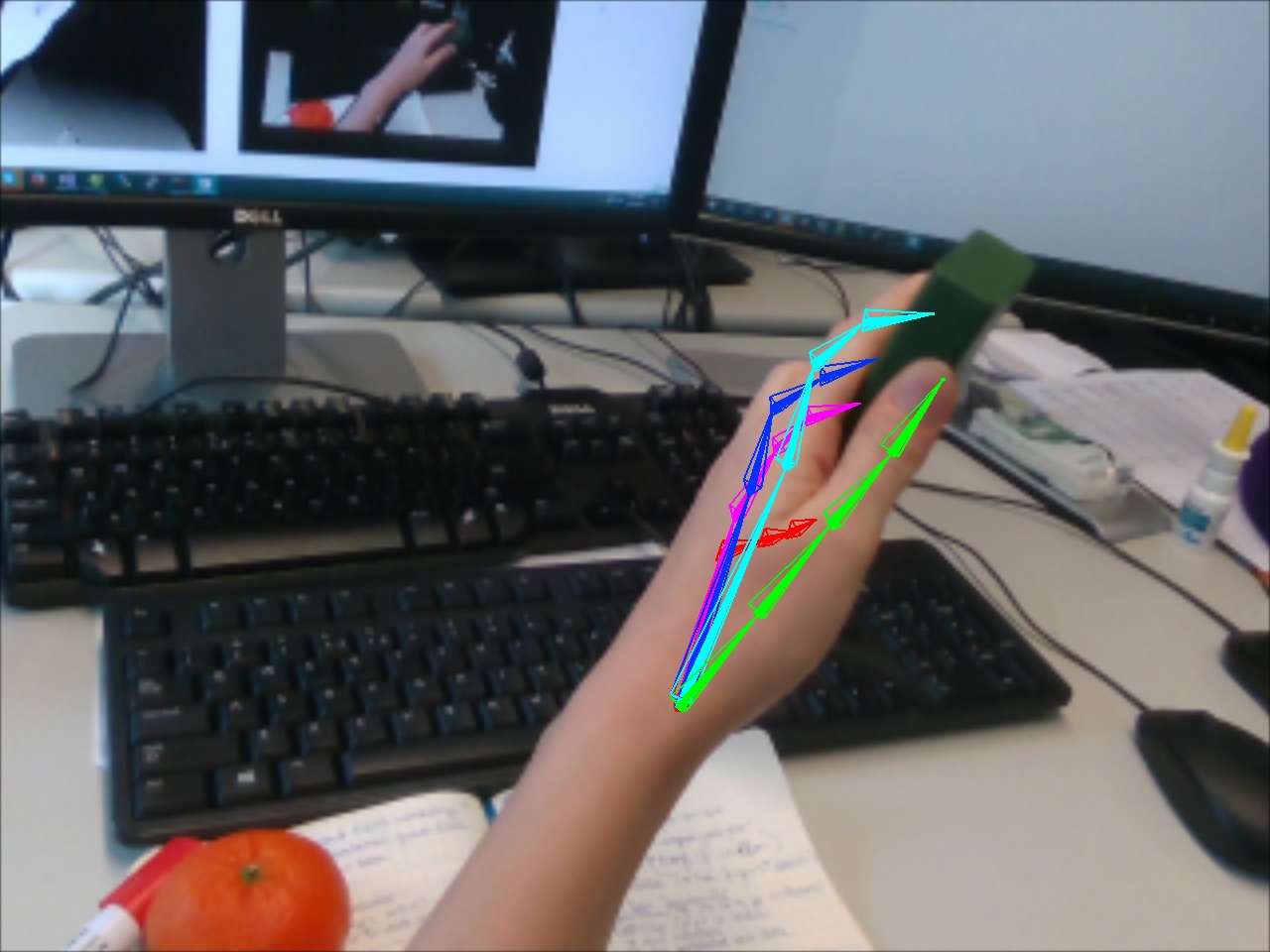}};
		\node[right of=skel_2d_3, xshift=2.3cm] (skel_2d_4) 
		{\includegraphics[width=\figScale]{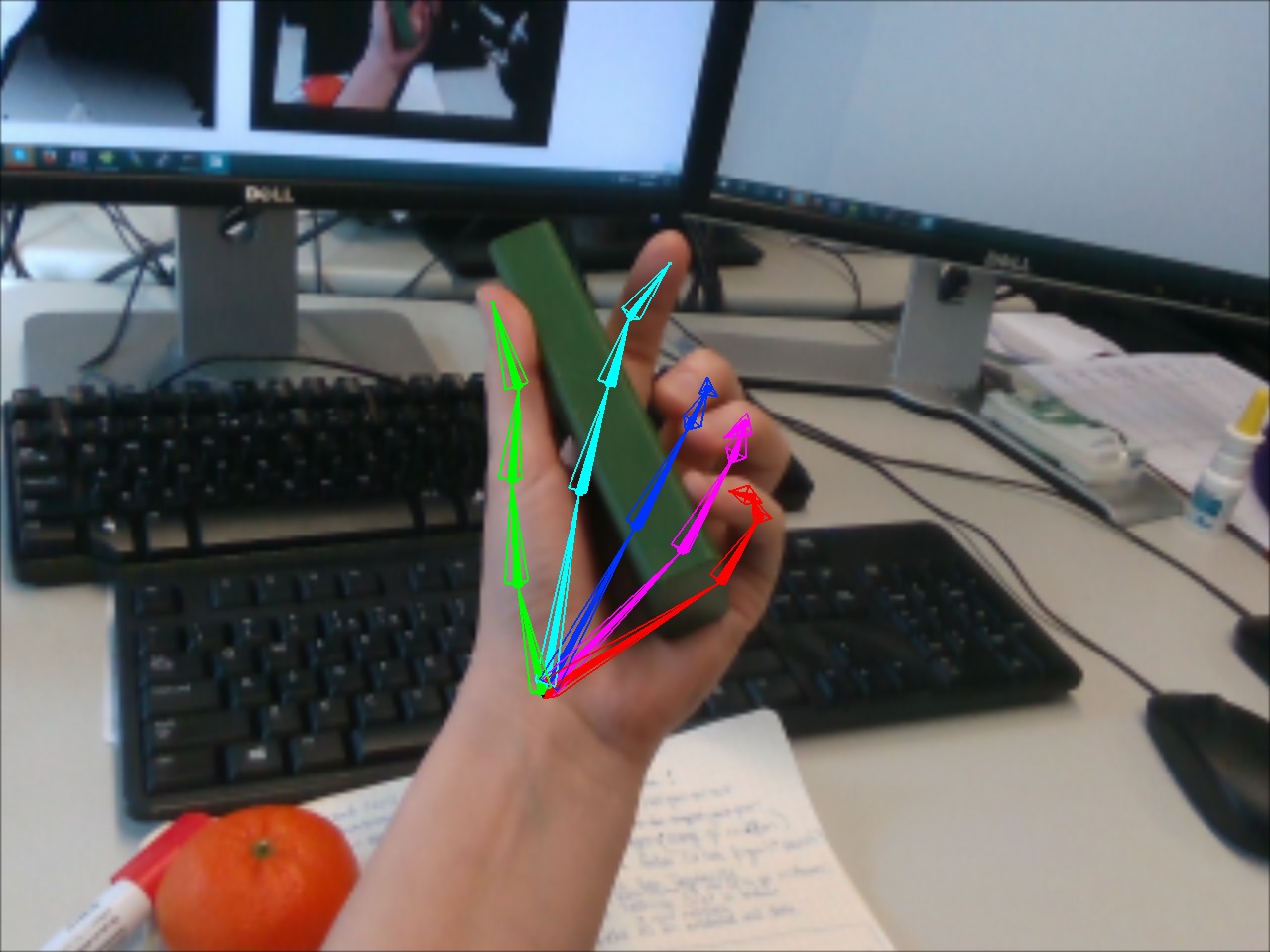}};
		\node[right of=skel_2d_4, xshift=2.3cm] (skel_2d_5) 
		{\includegraphics[width=\figScale]{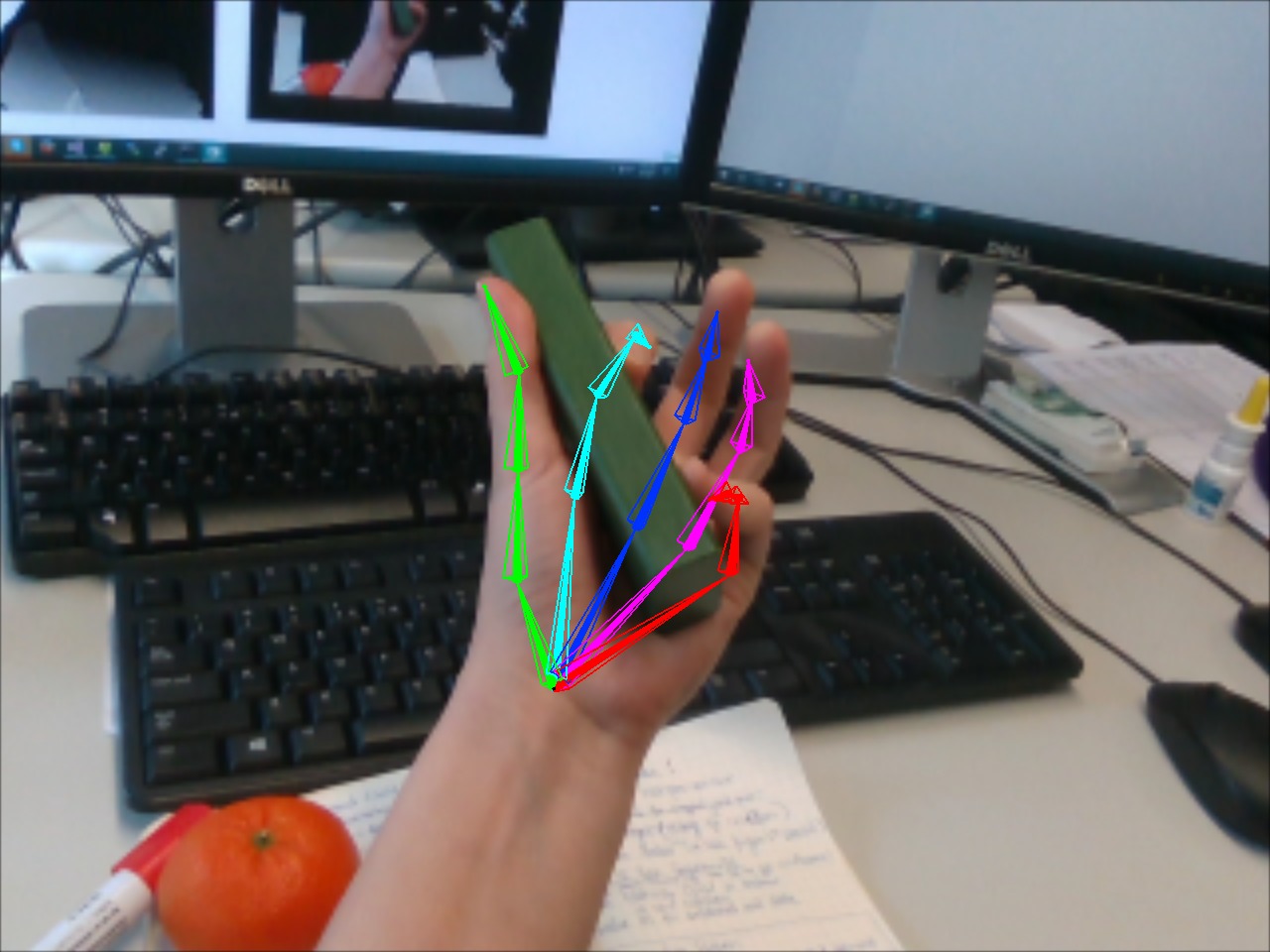}};
		\node[rotate=90, left of=textC, xshift=-1.5cm,yshift=0.2cm] (textD)  {Tracked 3D};
		\node[right of=textD,,xshift=1.1cm] (skel_3d_1) 
		{\includegraphics[width=\figScale]{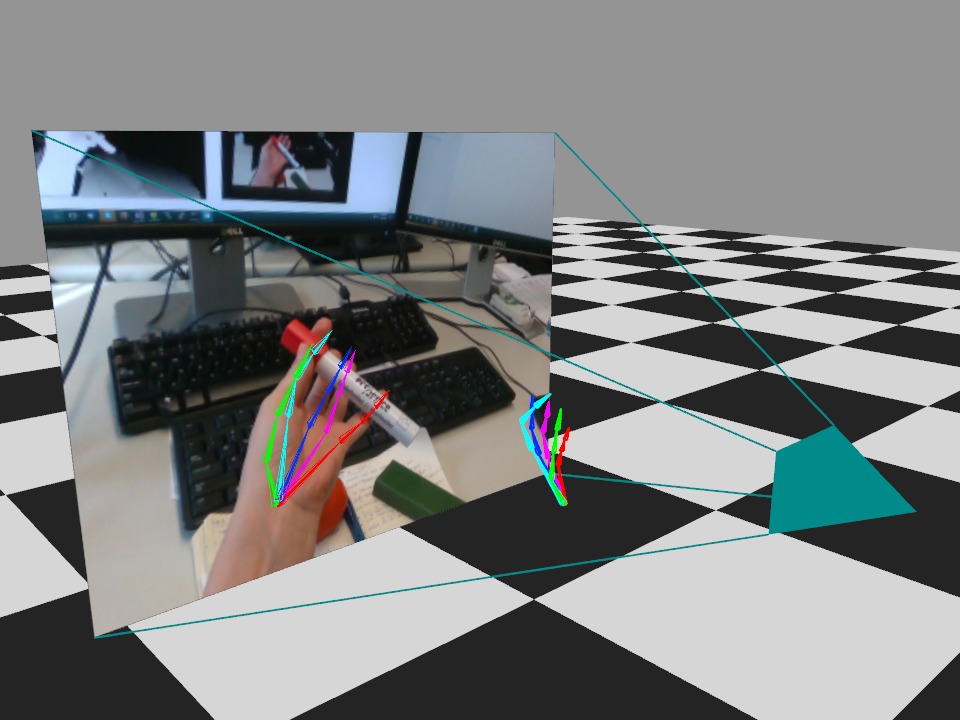}};
		\node[right of=skel_3d_1, xshift=2.3cm] (skel_3d_2) 
		{\includegraphics[width=\figScale]{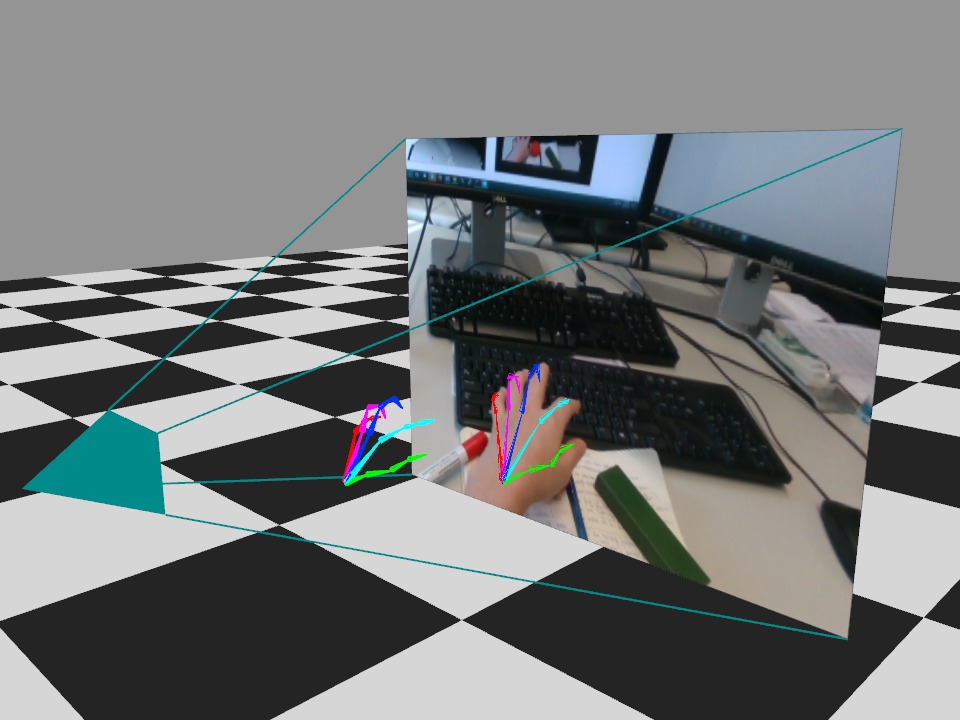}};
		\node[right of=skel_3d_2, xshift=2.3cm] (skel_3d_3) 
		{\includegraphics[width=\figScale]{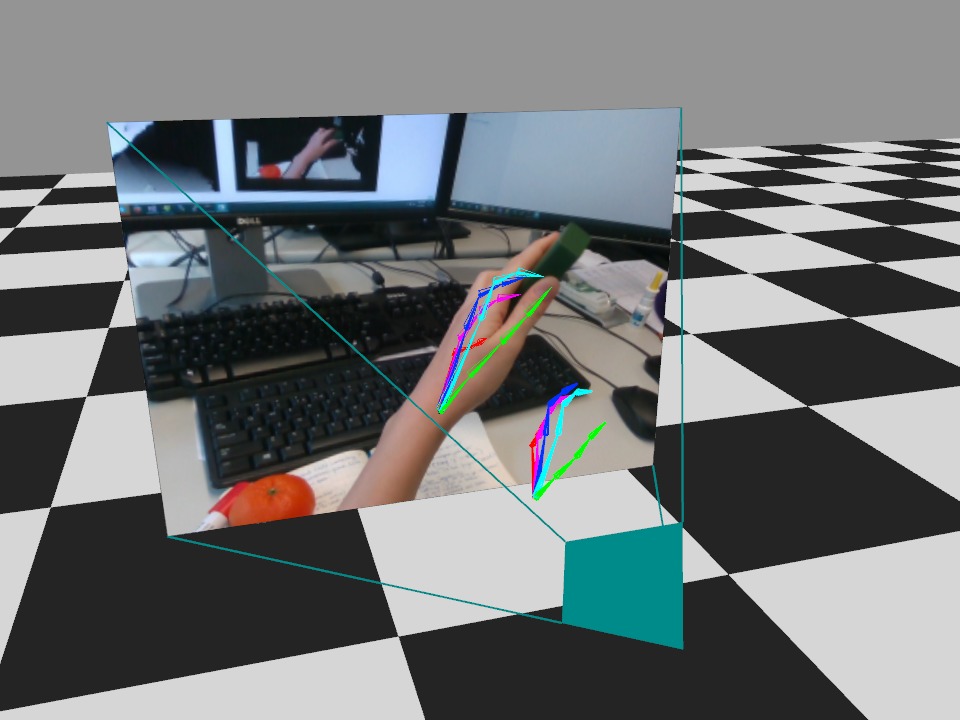}};
		\node[right of=skel_3d_3, xshift=2.3cm] (skel_3d_4) 
		{\includegraphics[width=\figScale]{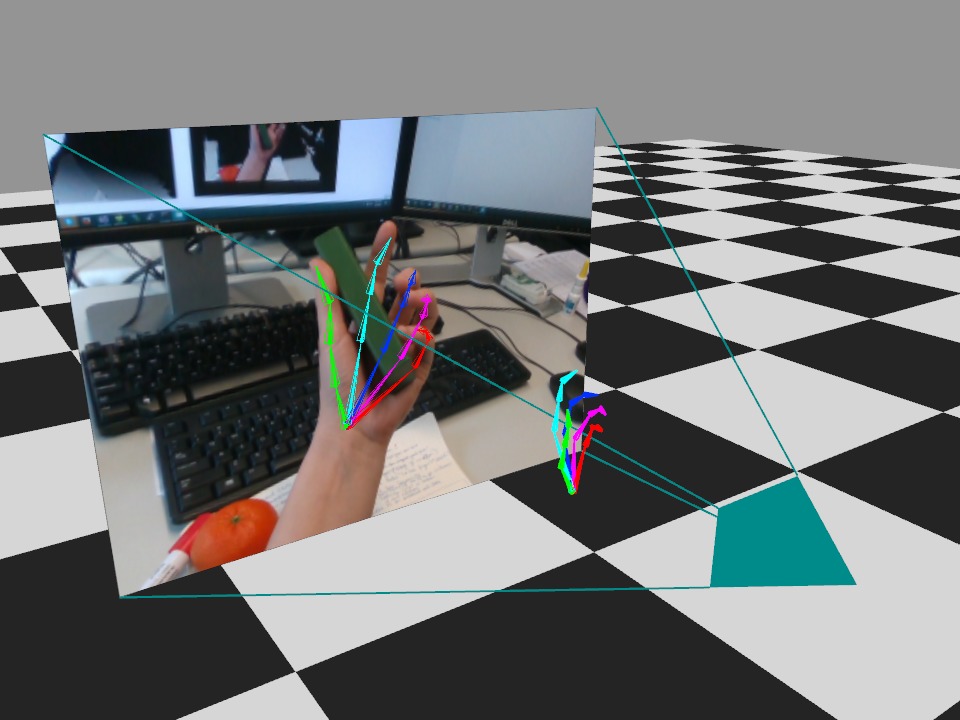}};
		\node[right of=skel_3d_4, xshift=2.3cm] (skel_3d_5) 
		{\includegraphics[width=\figScale]{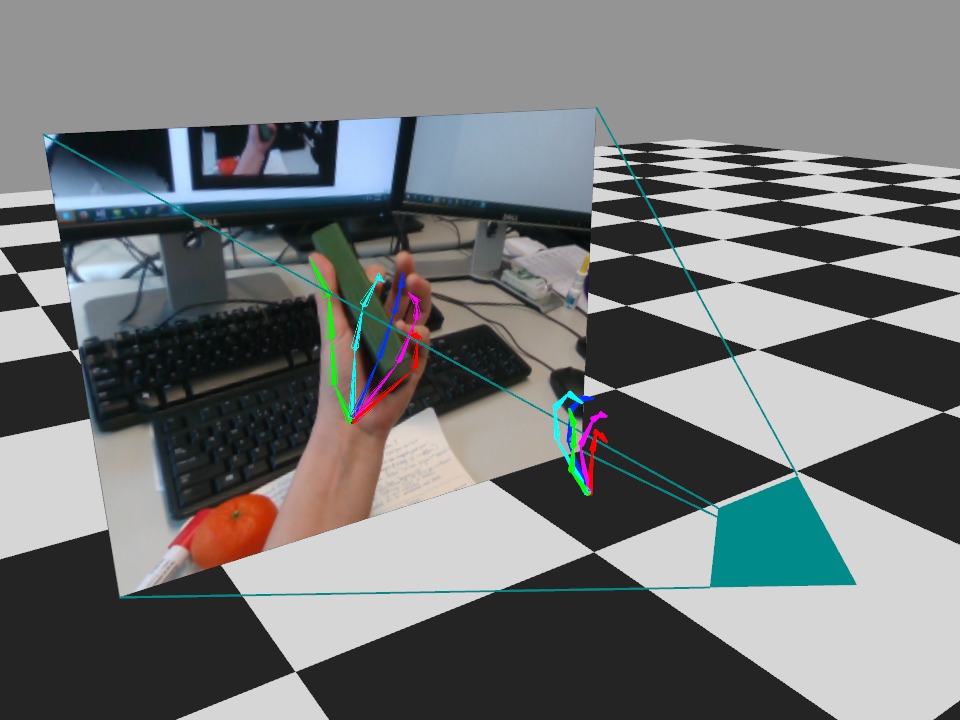}};
		\node[rotate=90, left of=textD, xshift=-1.5cm,yshift=0.3cm] (textE_top)  {Tracked 3D};
		\node[rotate=90, above of=textE_top, yshift=-1.5cm] (textE)  {(zoomed)};
		\node[right of=textE,xshift=0.9cm] (zoom_skel_3d_1) 
		{\includegraphics[width=\figScale,trim=8cm 5.2cm 11.5cm 10cm,
			clip]{images_supp/skel3D/Desk/000089.jpg}};
		\node[right of=zoom_skel_3d_1, xshift=2.3cm] (zoom_skel_3d_2) 
		{\includegraphics[width=\figScale,trim=10.5cm 6.1cm 12.5cm 11.6cm,
			clip]{images_supp/skel3D/Desk/000202.jpg}};
		\node[right of=zoom_skel_3d_2, xshift=2.3cm] (zoom_skel_3d_3) 
		{\includegraphics[width=\figScale,trim=12.5cm 7.1cm 8.5cm 9.1cm,
			clip]{images_supp/skel3D/Desk/000352.jpg}};
		\node[right of=zoom_skel_3d_3, xshift=2.3cm] (zoom_skel_3d_4) 
		{\includegraphics[width=\figScale,trim=10.5cm 7.85cm 10cm 8.1cm,
			clip]{images_supp/skel3D/Desk/000416.jpg}};
		\node[right of=zoom_skel_3d_4, xshift=2.3cm] (zoom_skel_3d_5) 
		{\includegraphics[width=\figScale,trim=10.5cm 7.85cm 10cm 8.1cm,
			clip]{images_supp/skel3D/Desk/000439.jpg}};
		\draw[orange,ultra thick,dashed] (1.2,-7.2) rectangle (2.6,-8.2);
		\draw[orange,ultra thick,dashed] (4.7,-7.3) rectangle (5.9,-8.2);
		\draw[orange,ultra thick,dashed] (8.2,-7.0) rectangle (9.5,-8.0);
		\draw[orange,ultra thick,dashed] (11.3,-7.0) rectangle (12.6,-8.0);
		\draw[orange,ultra thick,dashed] (14.7,-7.0) rectangle (16,-8.0);
		\draw[orange,ultra thick] (1.2,-8.2) -- (0.55,-8.9);
		\draw[orange,ultra thick] (2.6,-8.2) -- (3.65,-8.9);
		\draw[orange,ultra thick] (4.7,-8.2) -- (3.85,-8.9);
		\draw[orange,ultra thick] (5.9,-8.2) -- (6.97,-8.9);
		\draw[orange,ultra thick] (8.2,-8.0) -- (7.10,-8.9);
		\draw[orange,ultra thick] (9.5,-8.0) -- (10.25,-8.9);
		\draw[orange,ultra thick] (11.3,-8.0) -- (10.45,-8.9);
		\draw[orange,ultra thick] (12.65,-8.0) -- (13.55,-8.9);
		\draw[orange,ultra thick] (14.7,-8.0) -- (13.75,-8.9);
		\draw[orange,ultra thick] (16,-8.0) -- (16.85,-8.9);
		\draw[orange,ultra thick] (0.55,-8.9) rectangle (3.65,-11.1);
		\draw[orange,ultra thick] (3.85,-8.9) rectangle (6.97,-11.1);
		\draw[orange,ultra thick] (7.10,-8.9) rectangle (10.25,-11.1);	
		\draw[orange,ultra thick] (10.45,-8.9) rectangle (13.55,-11.1);	
		\draw[orange,ultra thick] (13.75,-8.9) rectangle (16.85,-11.1);	
		
		\end{tikzpicture}
		}
		\caption{Qualitative results on the \emph{Desk} sequence from EgoDexter~\cite{mueller_iccv2017}. \reg{} output (rows 1,2) and final tracking.}
		\label{fig:qualitative-desk}
	\end{figure*}
}
{
	\newcommand{\figScale}{0.18\textwidth}
	\begin{figure*}[h!]
		\centering
		\resizebox{.82\textwidth}{!}{
		\begin{tikzpicture}
		\node[rotate=90] (textA) at (0,0) {Prediction 2D};
		\node[right of=textA, xshift=1.1cm] (2d_1) 
		{\includegraphics[width=\figScale]{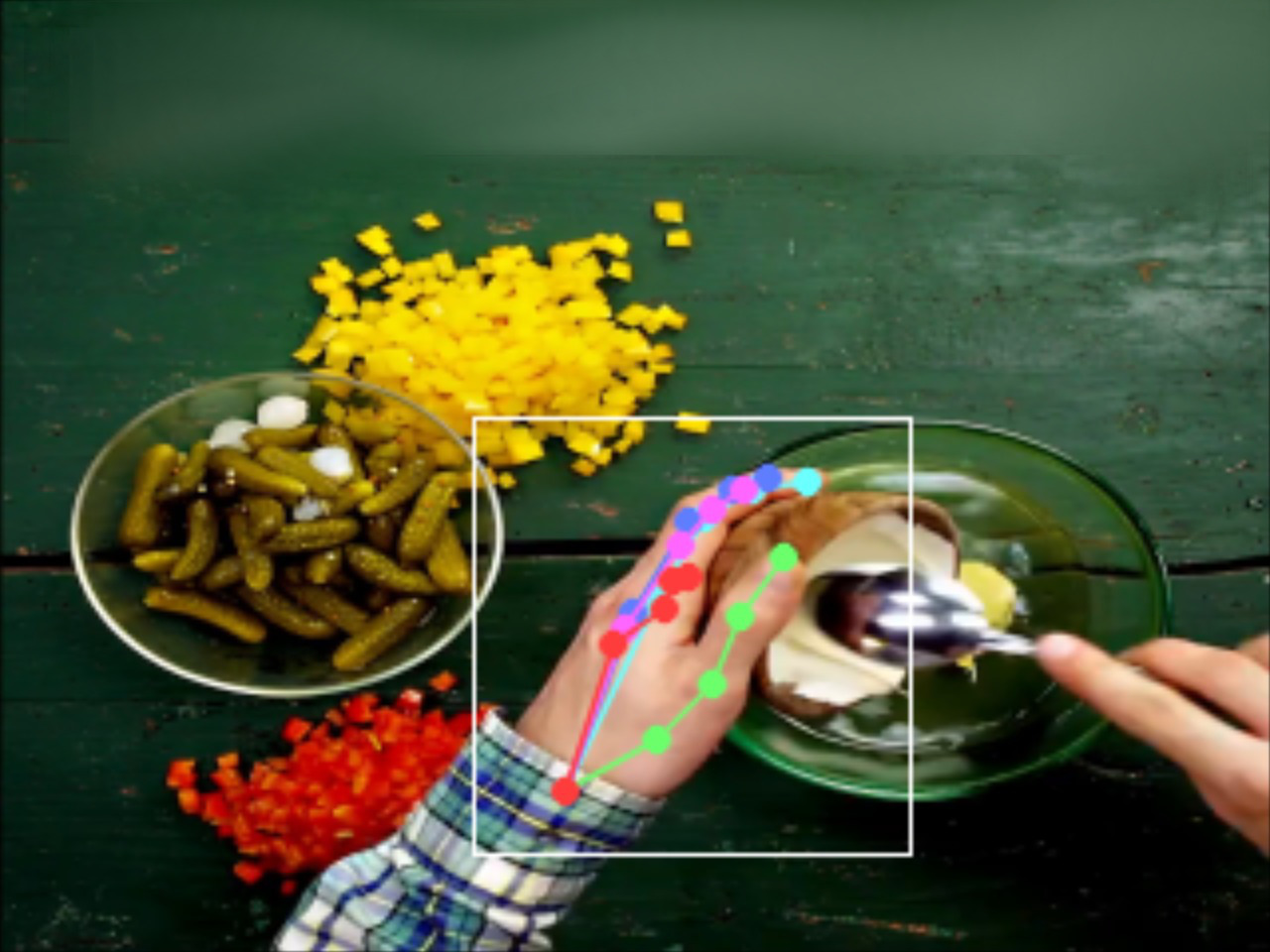}};
		\node[right of=2d_1, xshift=2.3cm] (2d_2) 
		{\includegraphics[width=\figScale]{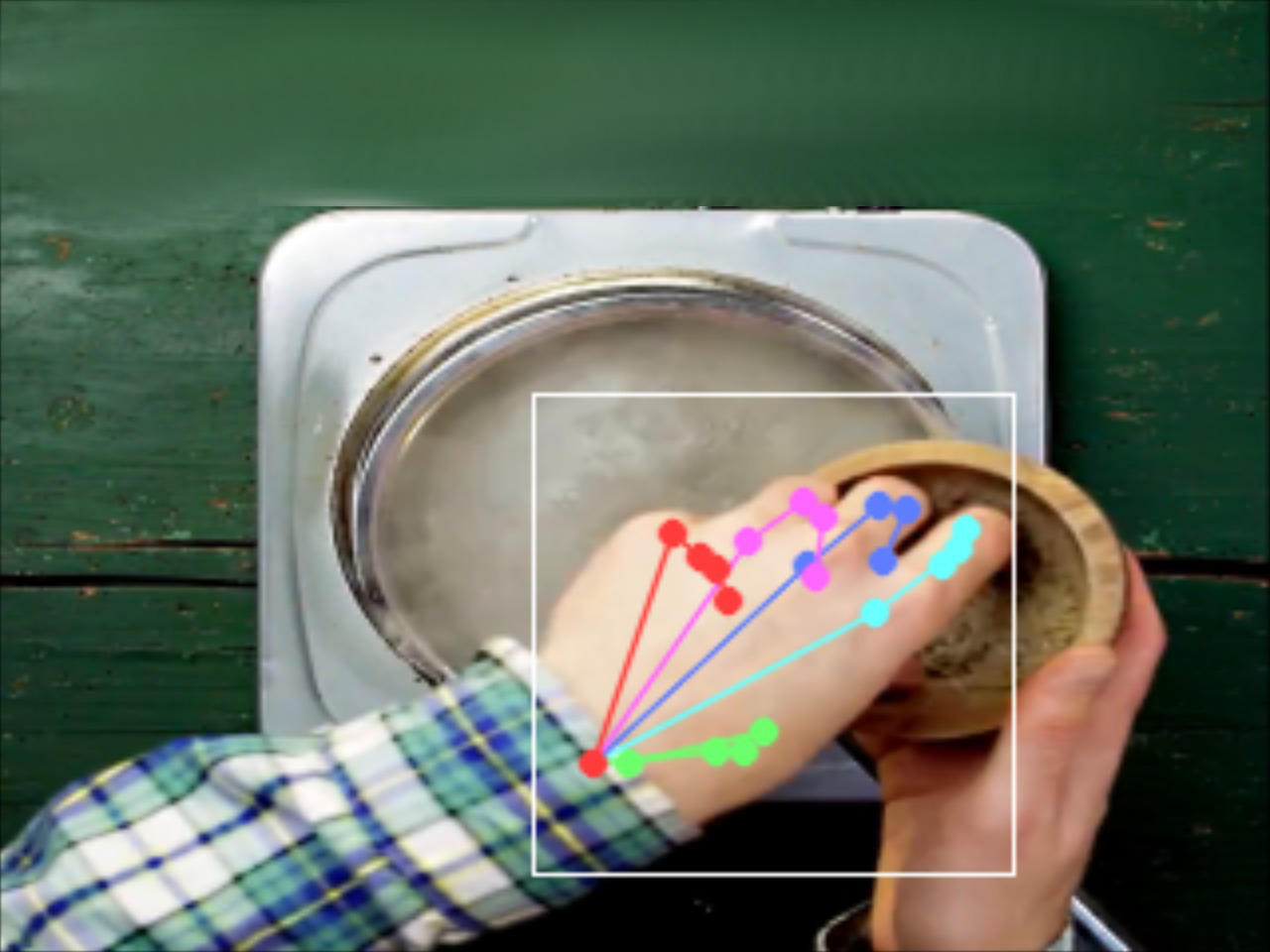}};
		\node[right of=2d_2, xshift=2.3cm] (2d_3) 
		{\includegraphics[width=\figScale]{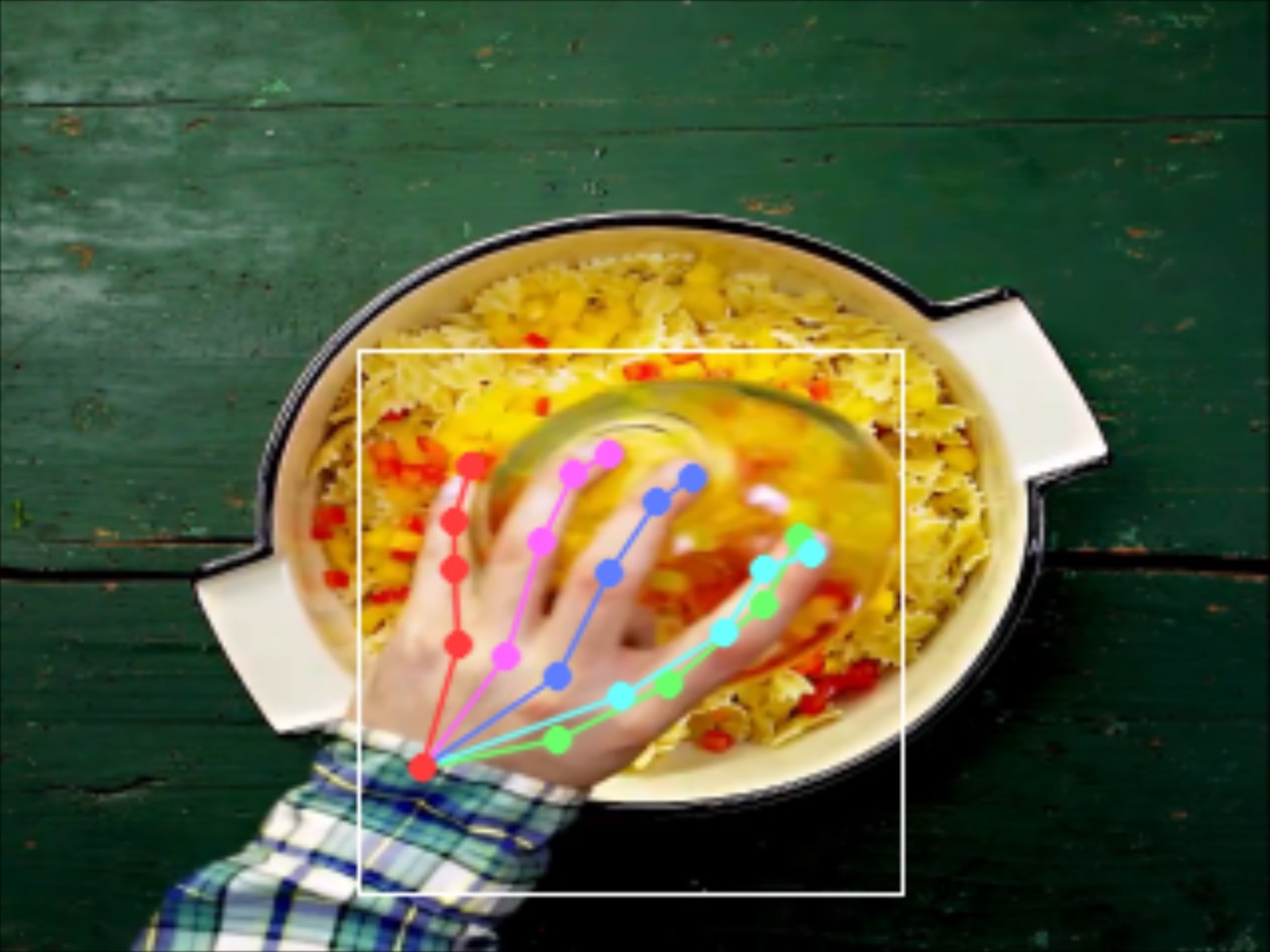}};
		\node[right of=2d_3, xshift=2.3cm] (2d_4) 
		{\includegraphics[width=\figScale]{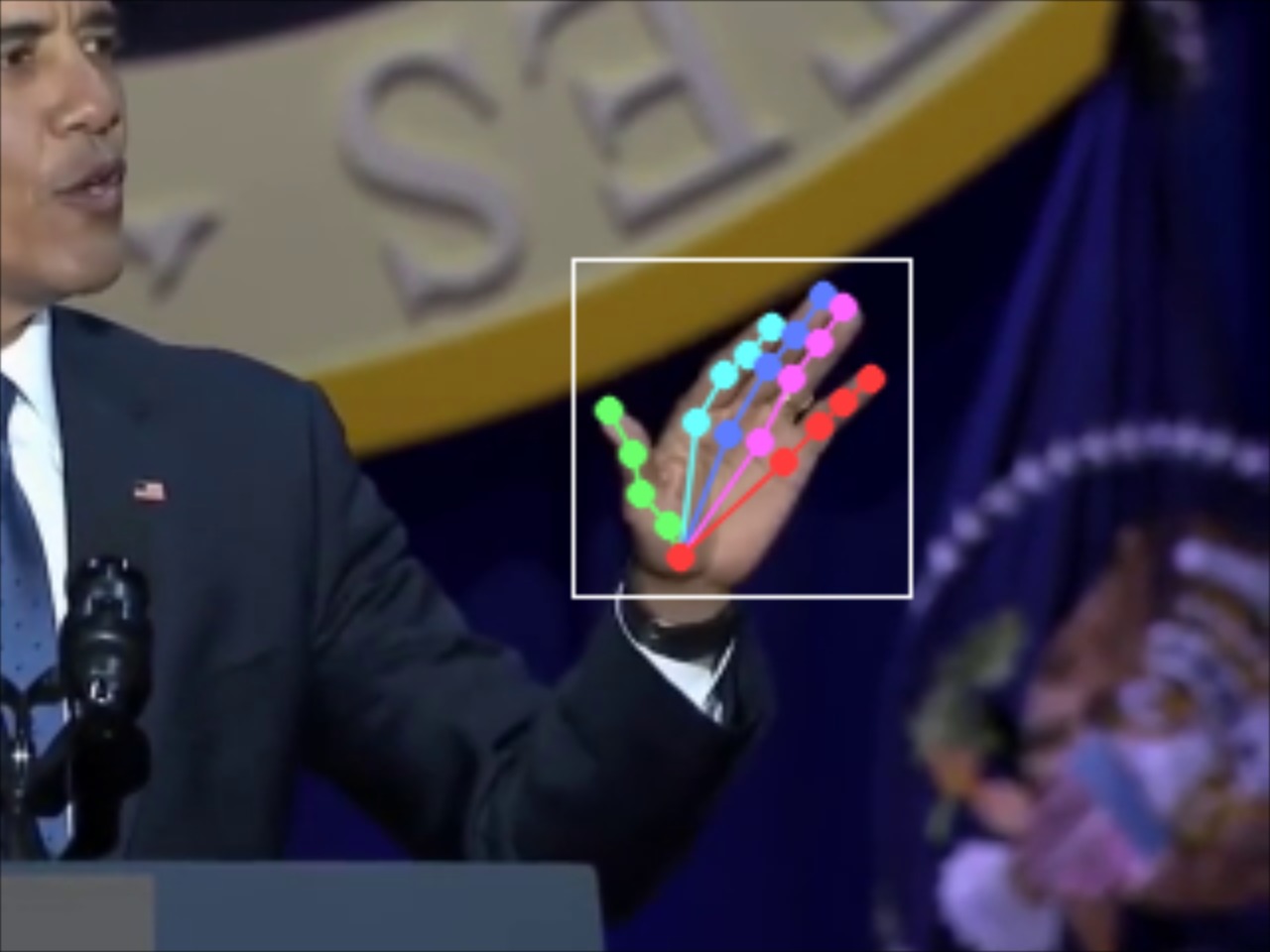}};
		\node[right of=2d_4, xshift=2.3cm] (2d_5) 
		{\includegraphics[width=\figScale]{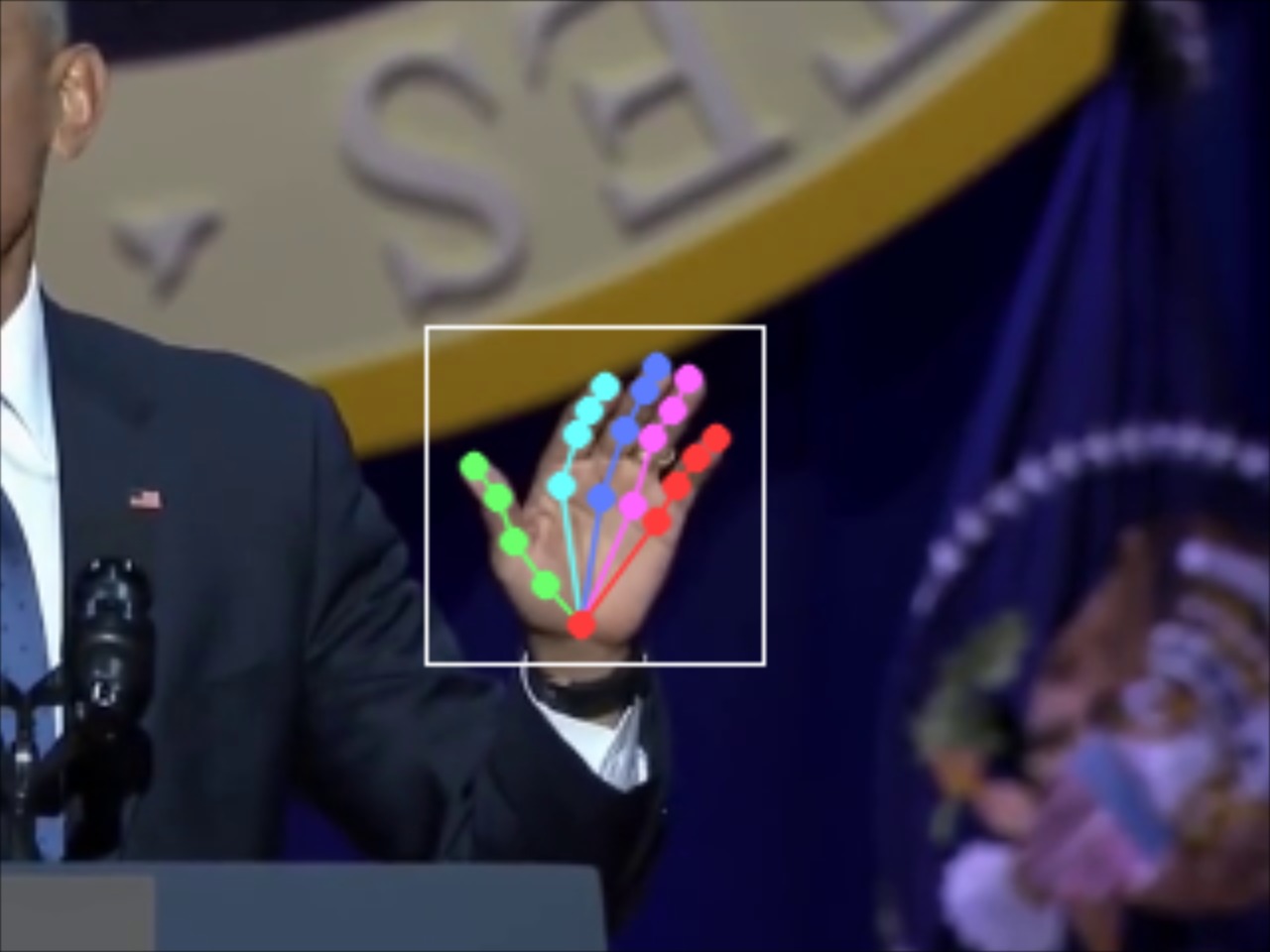}};
		\node[rotate=90, left of=textA, xshift=-1.5cm] (textB)  {Prediction 3D};
		\node[right of=textB,,xshift=1.1cm] (3d_1) 
		{\includegraphics[width=\figScale]{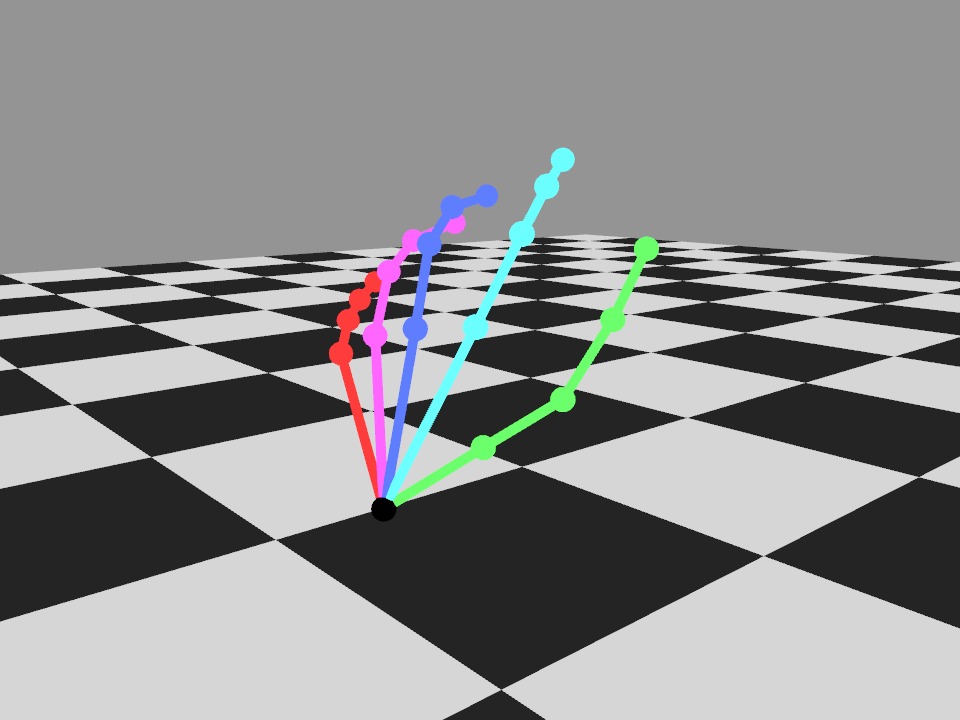}};
		\node[right of=3d_1, xshift=2.3cm] (3d_2) 
		{\includegraphics[width=\figScale]{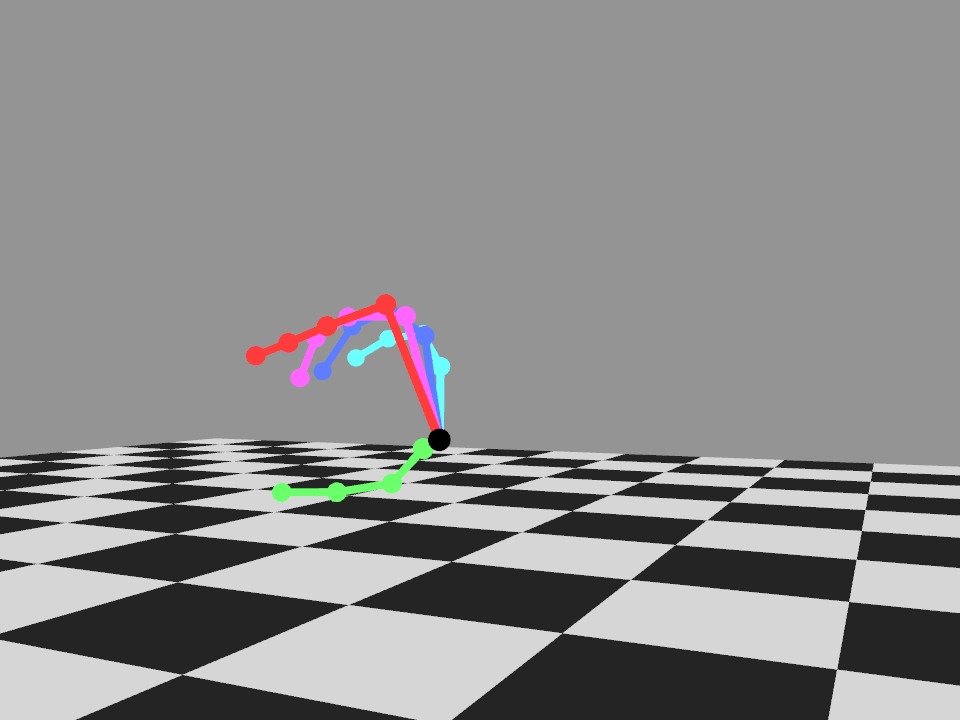}};
		\node[right of=3d_2, xshift=2.3cm] (3d_3) 
		{\includegraphics[width=\figScale]{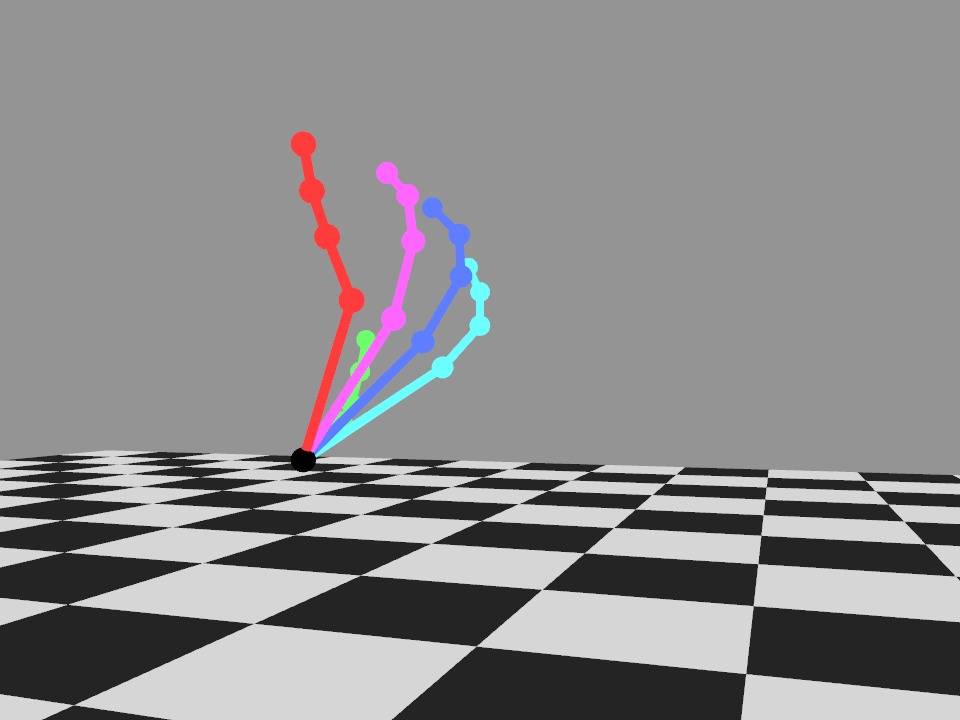}};
		\node[right of=3d_3, xshift=2.3cm] (3d_4) 
		{\includegraphics[width=\figScale]{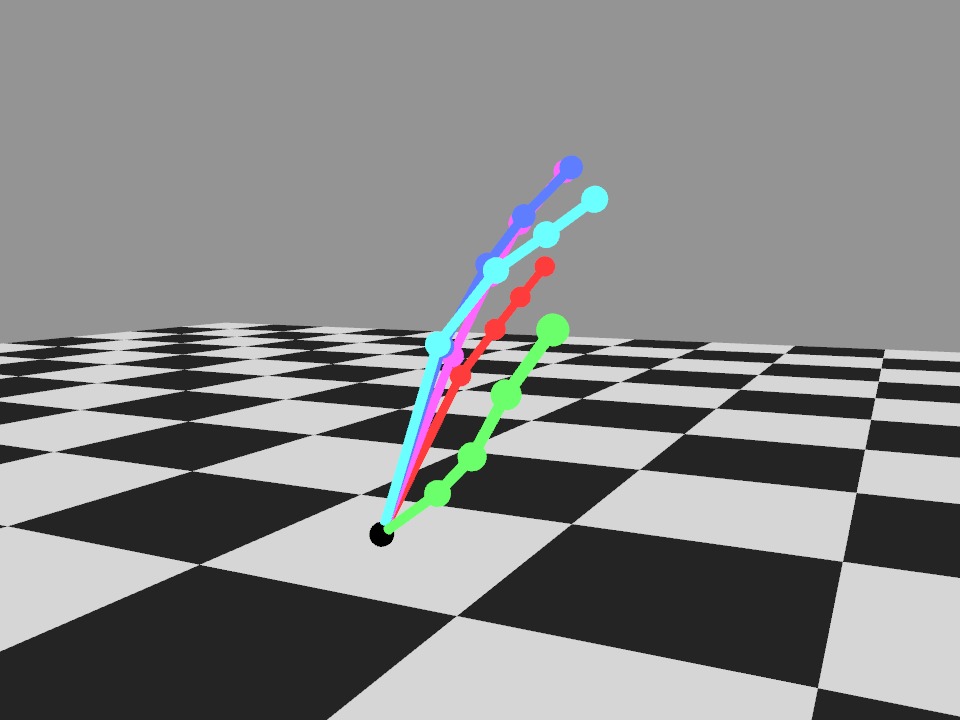}};
		\node[right of=3d_4, xshift=2.3cm] (3d_5) 
		{\includegraphics[width=\figScale]{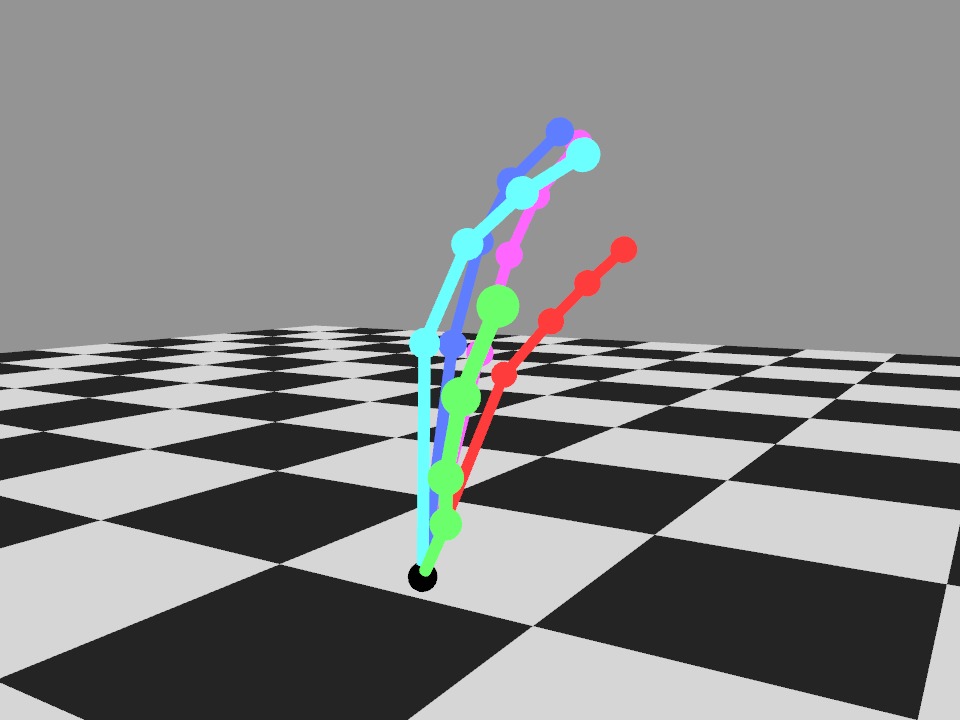}};
		\node[rotate=90, left of=textB, xshift=-1.5cm,yshift=0.3cm] (textC_top)  {Tracked 3D};
		\node[rotate=90, above of=textC_top, yshift=-1.5cm] (textC)  {(projected)};
		\node[right of=textC,xshift=0.9cm] (skel_2d_1) 
		{\includegraphics[width=\figScale]{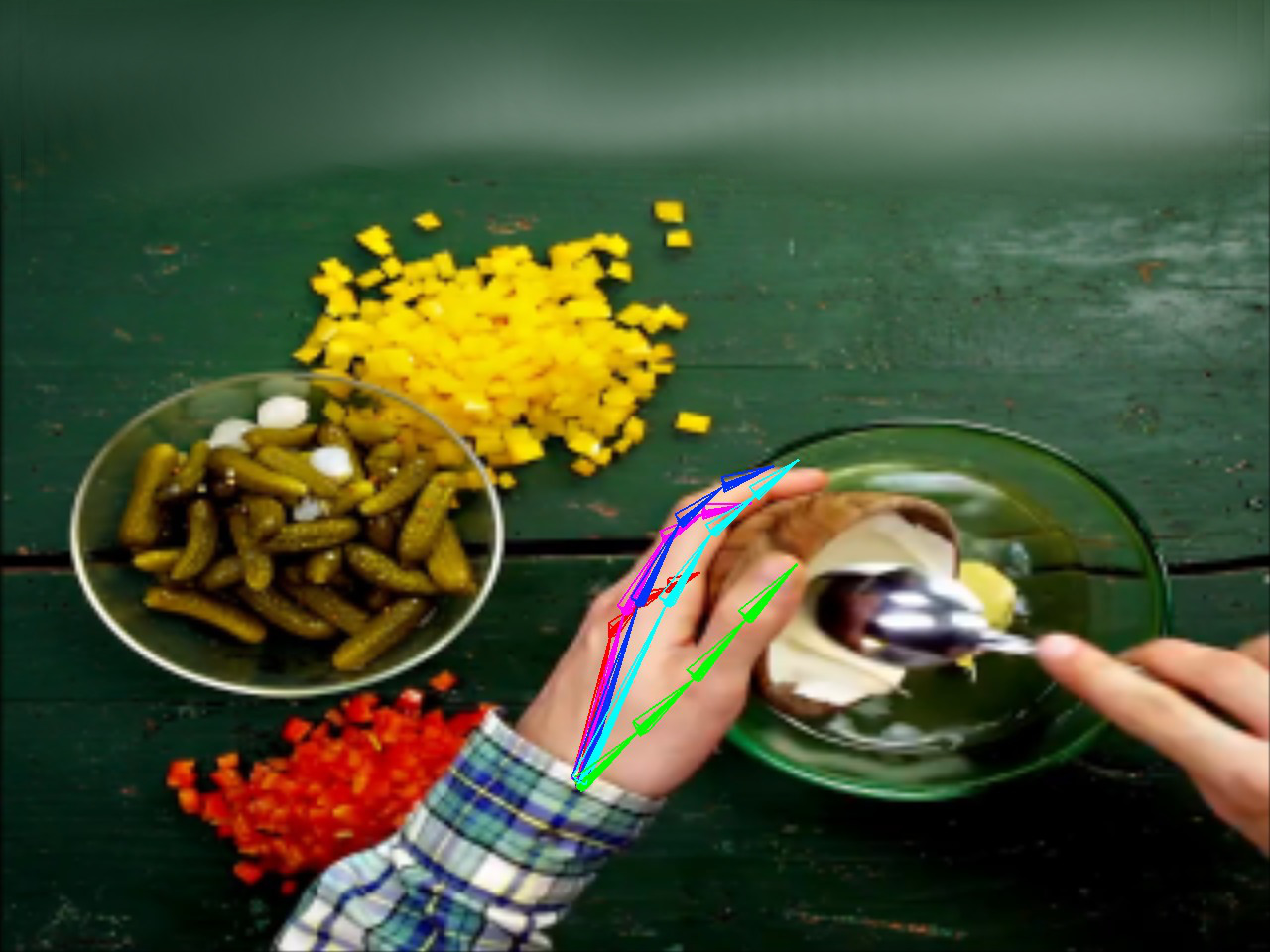}};
		\node[right of=skel_2d_1, xshift=2.3cm] (skel_2d_2) 
		{\includegraphics[width=\figScale]{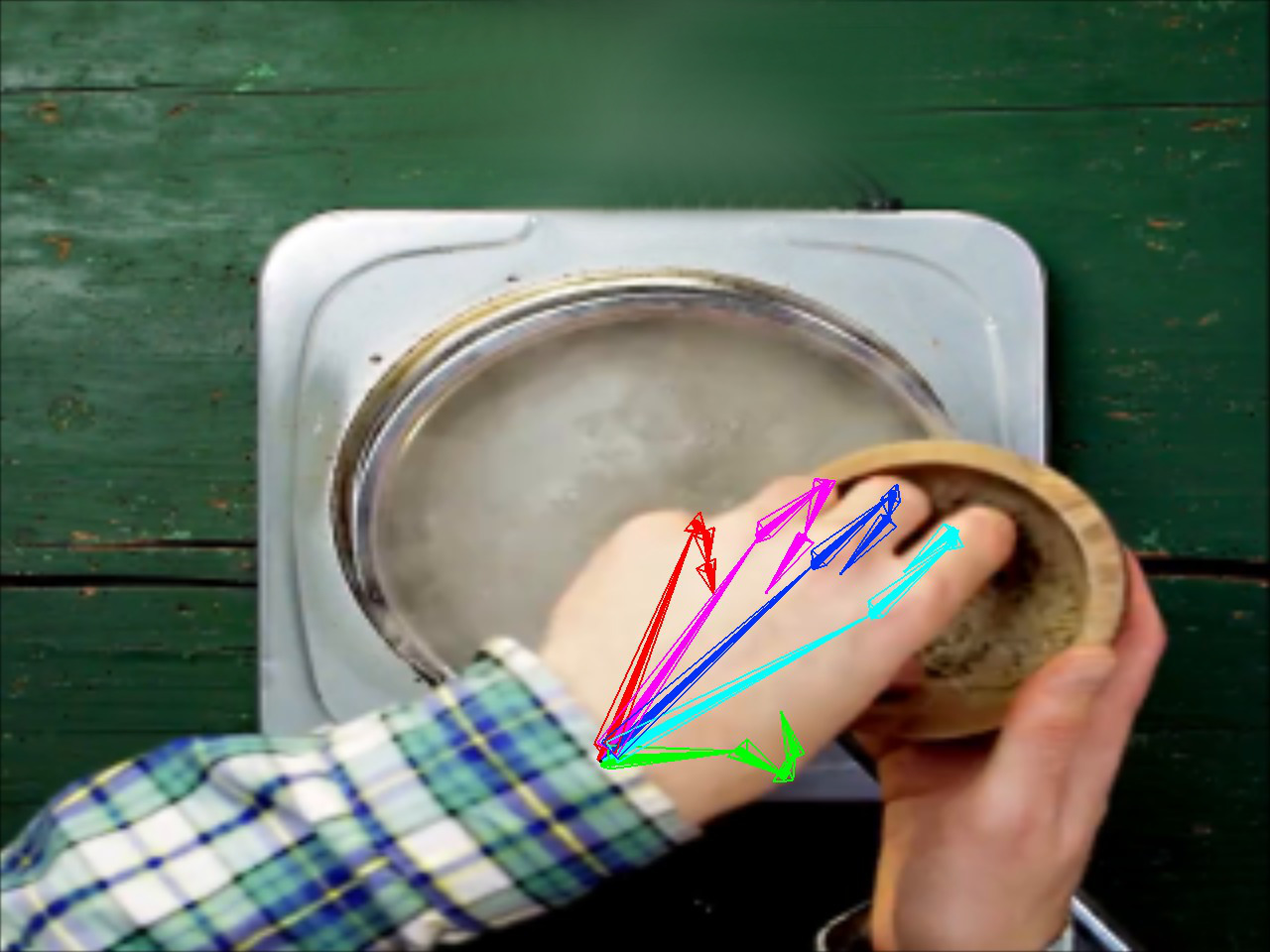}};
		\node[right of=skel_2d_2, xshift=2.3cm] (skel_2d_3) 
		{\includegraphics[width=\figScale]{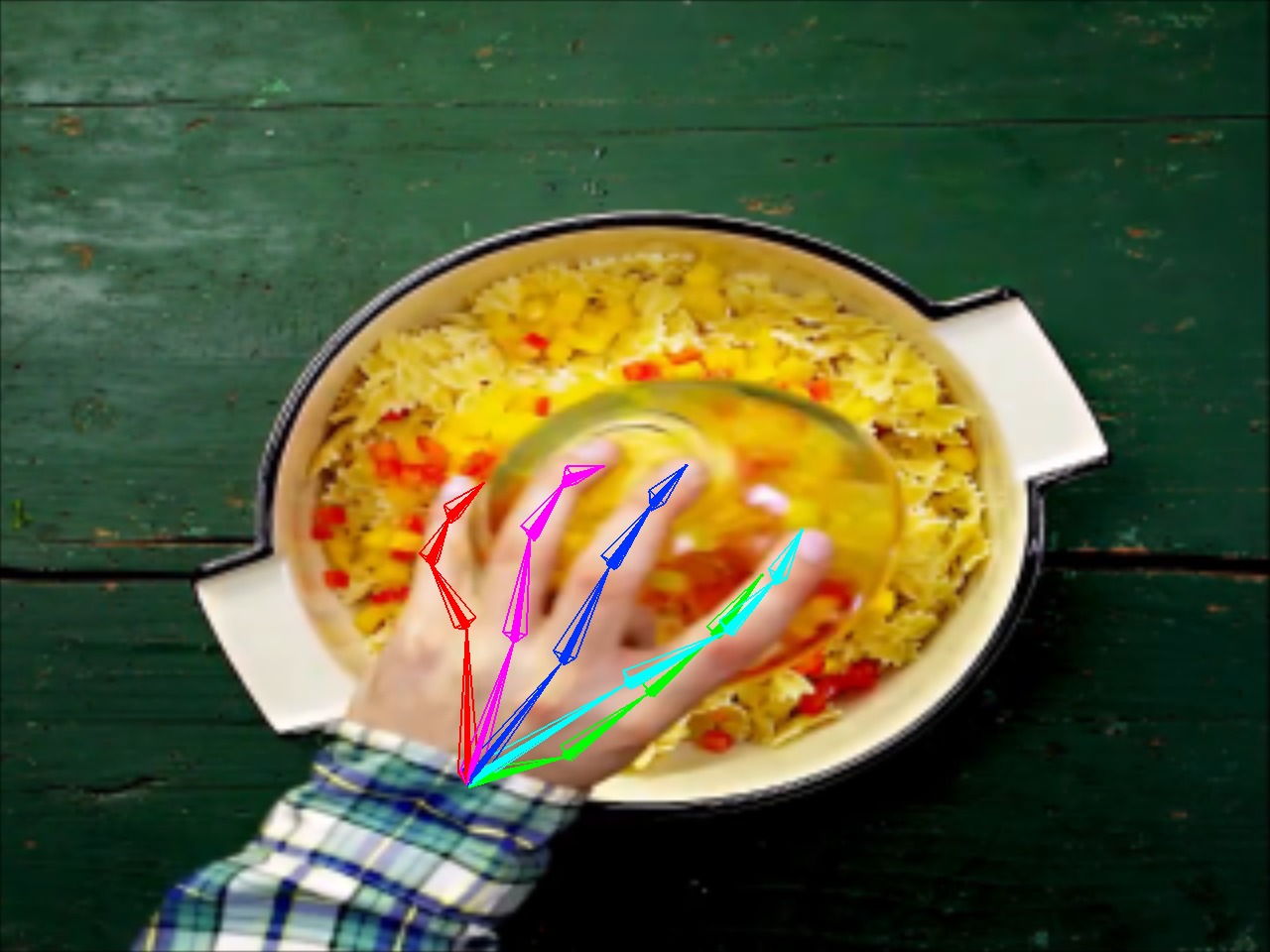}};
		\node[right of=skel_2d_3, xshift=2.3cm] (skel_2d_4) 
		{\includegraphics[width=\figScale]{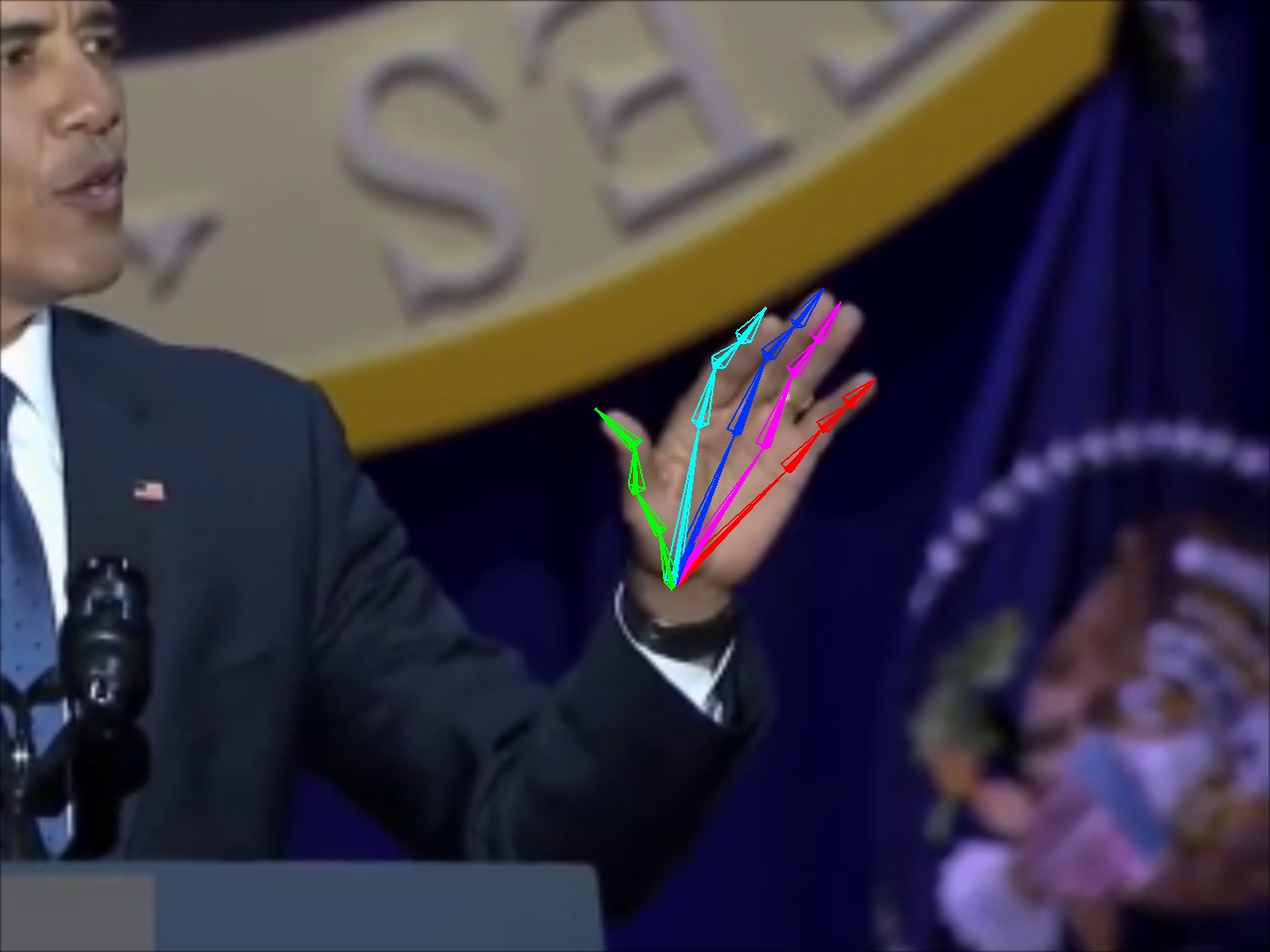}};
		\node[right of=skel_2d_4, xshift=2.3cm] (skel_2d_5) 
		{\includegraphics[width=\figScale]{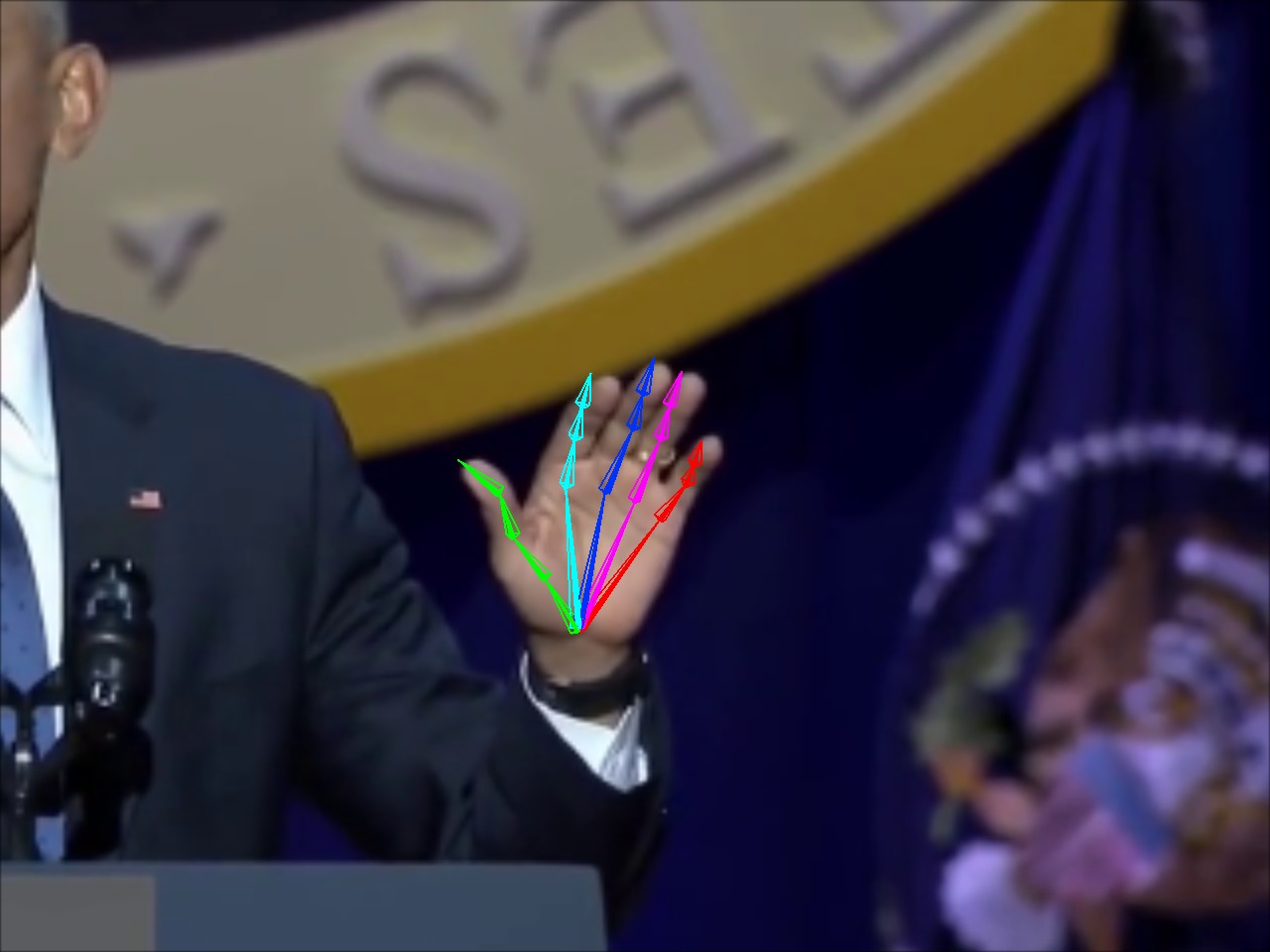}};
		\node[rotate=90, left of=textC, xshift=-1.5cm,yshift=0.2cm] (textD)  {Tracked 3D};
		\node[right of=textD,,xshift=1.1cm] (skel_3d_1) 
		{\includegraphics[width=\figScale]{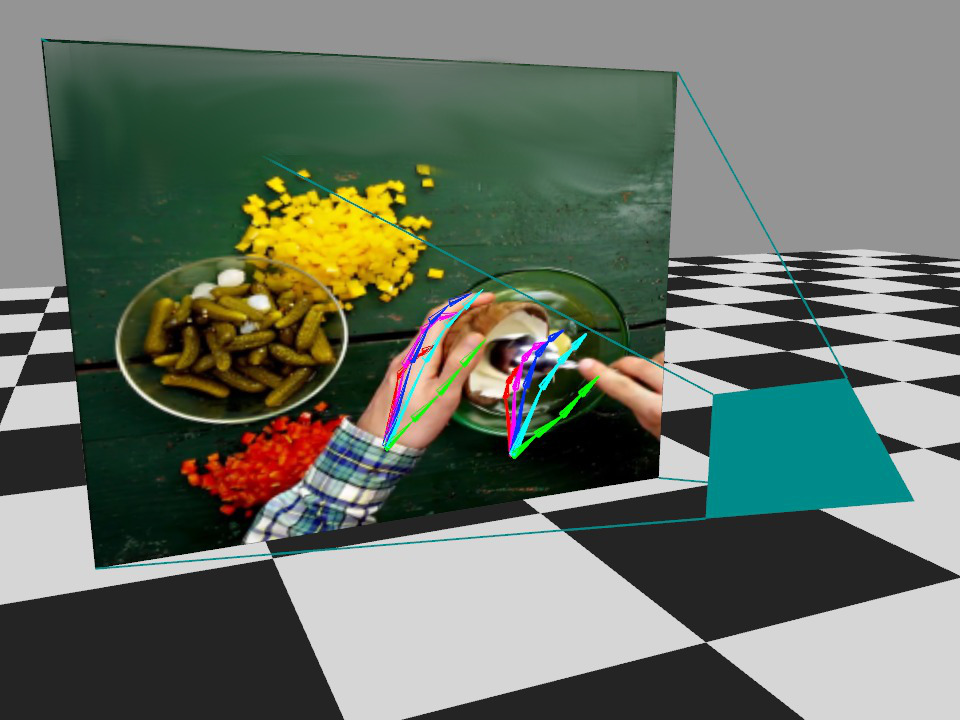}};
		\node[right of=skel_3d_1, xshift=2.3cm] (skel_3d_2) 
		{\includegraphics[width=\figScale]{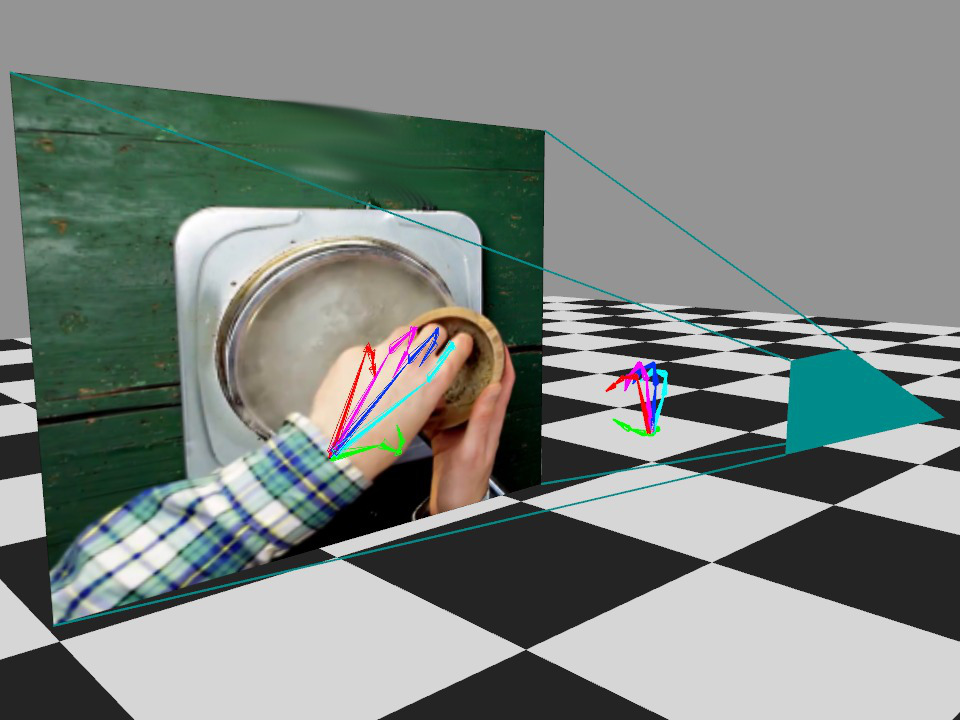}};
		\node[right of=skel_3d_2, xshift=2.3cm] (skel_3d_3) 
		{\includegraphics[width=\figScale]{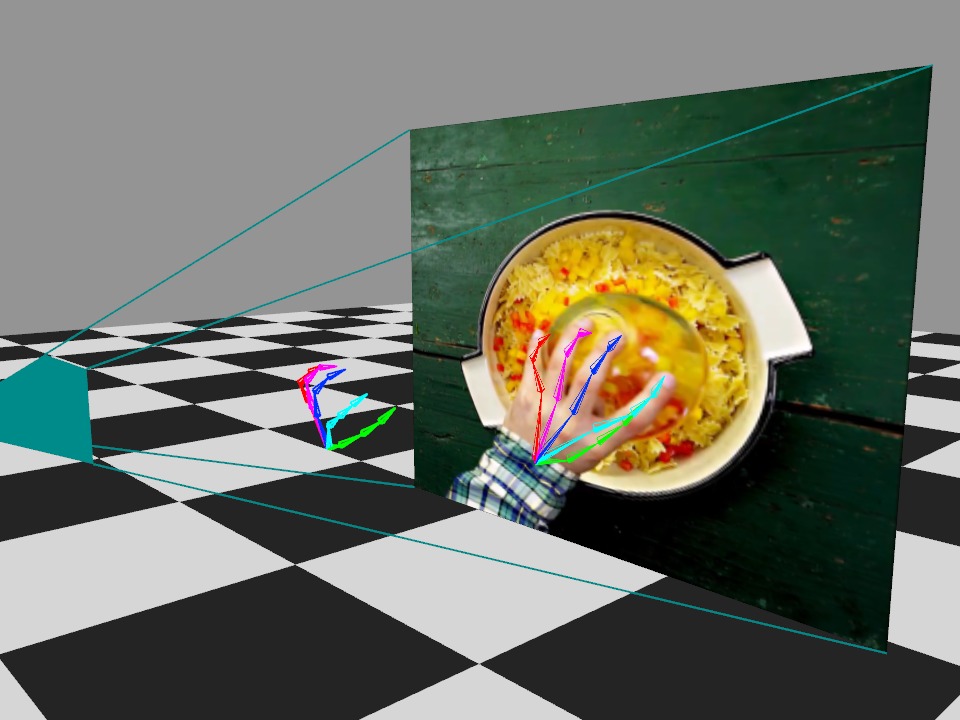}};
		\node[right of=skel_3d_3, xshift=2.3cm] (skel_3d_4) 
		{\includegraphics[width=\figScale]{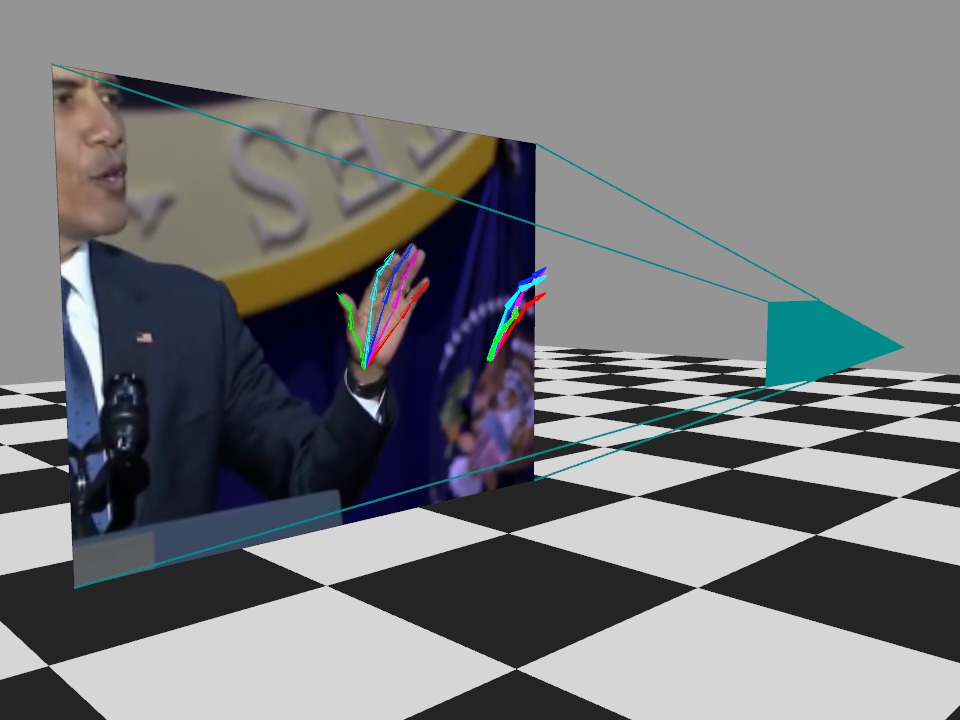}};
		\node[right of=skel_3d_4, xshift=2.3cm] (skel_3d_5) 
		{\includegraphics[width=\figScale]{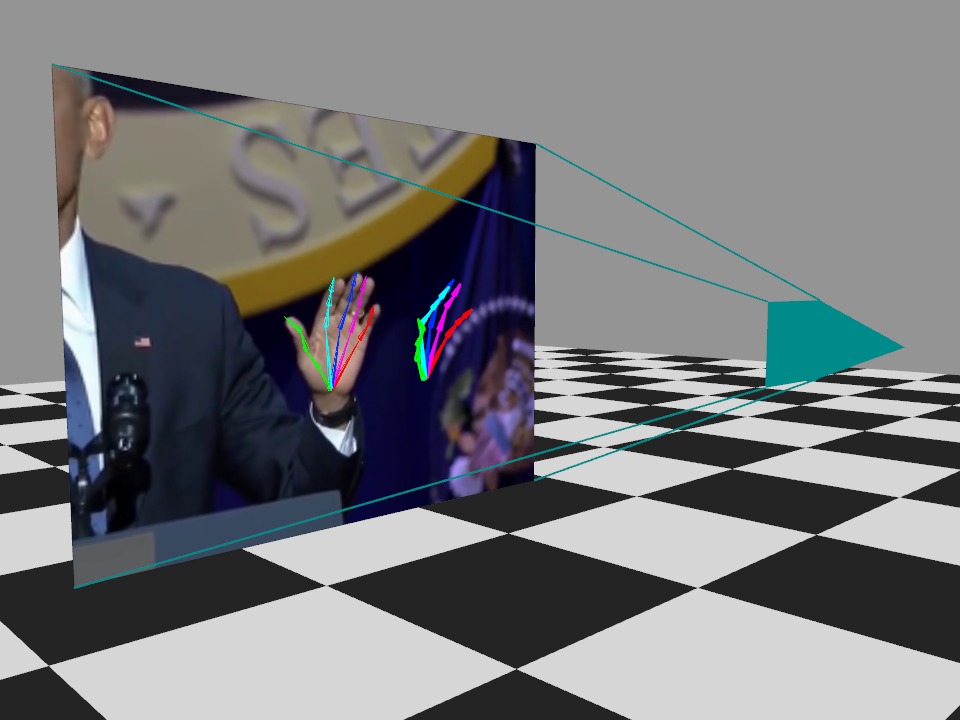}};
		\node[rotate=90, left of=textD, xshift=-1.5cm,yshift=0.3cm] (textE_top)  {Tracked 3D};
		\node[rotate=90, above of=textE_top, yshift=-1.5cm] (textE)  {(zoomed)};
		\node[right of=textE,xshift=0.9cm] (zoom_skel_3d_1) 
		{\includegraphics[width=\figScale,trim=10cm 6.2cm 9.5cm 9cm,
			clip]{images_supp/skel3D/Nudelsalat/000075_edit.jpg}};
		\node[right of=zoom_skel_3d_1, xshift=2.3cm] (zoom_skel_3d_2) 
		{\includegraphics[width=\figScale,trim=10.5cm 6.5cm 9.5cm 9cm,
			clip]{images_supp/skel3D/Nudelsalat/000079_edit.jpg}};
		\node[right of=zoom_skel_3d_2, xshift=2.3cm] (zoom_skel_3d_3) 
		{\includegraphics[width=\figScale,trim=9cm 6cm 8.5cm 8cm,
			clip]{images_supp/skel3D/Nudelsalat/000123.jpg}};
		\node[right of=zoom_skel_3d_3, xshift=2.3cm] (zoom_skel_3d_4) 
		{\includegraphics[width=\figScale,trim=9.5cm 8.85cm 11cm 7.1cm,
			clip]{images_supp/skel3D/Obama/000040.jpg}};
		\node[right of=zoom_skel_3d_4, xshift=2.3cm] (zoom_skel_3d_5) 
		{\includegraphics[width=\figScale,trim=9.5cm 7.85cm 11cm 8.1cm,
			clip]{images_supp/skel3D/Obama/000142.jpg}};
			\draw[orange,ultra thick,dashed] (1.2,-7.2) rectangle (2.6,-8.2);
			\draw[orange,ultra thick,dashed] (4.7,-7.3) rectangle (6.1,-8.2);
			\draw[orange,ultra thick,dashed] (8.0,-7.0) rectangle (9.6,-8.2);
			\draw[orange,ultra thick,dashed] (11.3,-7.0) rectangle (12.6,-8.0);
			\draw[orange,ultra thick,dashed] (14.7,-7.0) rectangle (16,-8.0);
					\draw[orange,ultra thick] (1.2,-8.2) -- (0.55,-8.9);
					\draw[orange,ultra thick] (2.6,-8.2) -- (3.65,-8.9);
					\draw[orange,ultra thick] (4.7,-8.2) -- (3.85,-8.9);
					\draw[orange,ultra thick] (5.9,-8.2) -- (6.97,-8.9);
					\draw[orange,ultra thick] (8.0,-8.2) -- (7.10,-8.9);
					\draw[orange,ultra thick] (9.62,-8.2) -- (10.25,-8.9);
					\draw[orange,ultra thick] (11.3,-8.0) -- (10.45,-8.9);
					\draw[orange,ultra thick] (12.65,-8.0) -- (13.55,-8.9);
					\draw[orange,ultra thick] (14.7,-8.0) -- (13.75,-8.9);
					\draw[orange,ultra thick] (16,-8.0) -- (16.85,-8.9);
			\draw[orange,ultra thick] (0.55,-8.9) rectangle (3.65,-11.1);
			\draw[orange,ultra thick] (3.85,-8.9) rectangle (6.97,-11.1);
			\draw[orange,ultra thick] (7.10,-8.9) rectangle (10.25,-11.1);	
			\draw[orange,ultra thick] (10.45,-8.9) rectangle (13.55,-11.1);	
			\draw[orange,ultra thick] (13.75,-8.9) rectangle (16.85,-11.1);	
		\end{tikzpicture}
	}
		\caption{Qualitative results on community videos from YouTube. \reg{} output (rows 1,2) and final tracking.}
		\label{fig:qualitative-community}
	\end{figure*}
}

\parahead{Training Details} 
We train \reg~in the Caffe~\cite{jia_caffe_ICM} framework for 300,000 iterations with a batch size of 32.
We use the AdaDelta~\cite{zeiler2012adadelta} solver with an initial learning rate of $0.1$ which is lowered to $0.01$ after 150,000 iterations.
All layers which are shared between our network and \emph{ResNet50} are initialized with the weights obtained from ImageNet pretraining~\cite{ILSVRC15}.
Both, the 2D heatmap loss and the local 3D joint position loss, are formulated using the Euclidean loss with loss weights of 1 and 100, respectively.

\parahead{Computational Time}
A forward pass of \reg~in our real-time tracking system takes 13~ms on a GTX 1080 Ti.

\section{Comparison with RGB-D methods}
\label{sec:quantitative}
The 3D tracking of hands in purely RGB images is an extremely
challenging problem due to inherent depth ambiguities of monocular RGB images.
While our method advances the state-of-the-art of RGB-only hand tracking methods, there is still a gap between RGB-only and RGB-D methods~\cite{mueller_iccv2017,sharp2015accurate,sridhar_cvpr2015}.
A quantitative analysis of this accuracy gap is shown in Fig.~\ref{fig:quantitative_3d}, where we compare our results (dark blue) with the RGB-D method from Sridhar \etal \cite{sridhar_eccv2016} (red). 

In order to better understand the source of errors, we perform an additional experiment where we translated the global z-position of our RGB-only results to best match the depth of the ground truth.
In Fig.~\ref{fig:quantitative_3d} we compare these depth-normalized results (light blue) with our original results (blue). It can be seen that a significant portion of the gap between methods based on RGB and RGB-D is due to inaccuracies in the estimation of the hand root position. Reasons for an inaccurate hand root position include a skeleton that does not perfectly fit the user's hand (in terms of bone lengths), as well as inaccuracies in the 2D predictions. 

\section{Detailed Qualitative Evaluation}
\label{sec:qualitative}
In Figs. \ref{fig:qualitative-desk} and \ref{fig:qualitative-community} we qualitatively evaluate each of the intermediate stages along our tracking solution as well as the final result.
In particular, Fig.~\ref{fig:qualitative-desk} shows results on the EgoDexter dataset~\cite{mueller_iccv2017} where a subject grabs different objects in an office environment, and Fig.~\ref{fig:qualitative-community} shows results on community videos downloaded from YouTube.
In both figures, we provide visualizations of: heatmap maxima of the 2D joint detections (first row); root-relative 3D joint detections (second row); global 3D tracked hand projected into camera plane (third row); and global 3D tracked hand visualized in a virtual scenario with the original camera frustum (fourth and fifth rows).
Please see the supplementary video for complete sequences.

%% file: main_paper.bbl
\begin{thebibliography}{10}\itemsep=-1pt

\bibitem{abadi2016tensorflow}
M.~Abadi, A.~Agarwal, P.~Barham, E.~Brevdo, Z.~Chen, C.~Citro, G.~S. Corrado,
  A.~Davis, J.~Dean, M.~Devin, et~al.
\newblock Tensorflow: Large-scale machine learning on heterogeneous distributed
  systems.
\newblock {\em arXiv preprint arXiv:1603.04467}, 2016.

\bibitem{ballan_eccv2012}
L.~Ballan, A.~Taneja, J.~Gall, L.~V. Gool, and M.~Pollefeys.
\newblock {Motion Capture of Hands in Action using Discriminative Salient
  Points}.
\newblock In {\em European Conference on Computer Vision (ECCV)}, 2012.

\bibitem{brau20163d}
E.~Brau and H.~Jiang.
\newblock 3d human pose estimation via deep learning from 2d annotations.
\newblock In {\em 3D Vision (3DV), 2016 Fourth International Conference on},
  pages 582--591. IEEE, 2016.

\bibitem{casiez20121}
G.~Casiez, N.~Roussel, and D.~Vogel.
\newblock 1\euro{} filter: a simple speed-based low-pass filter for noisy input
  in interactive systems.
\newblock In {\em Proceedings of the SIGCHI Conference on Human Factors in
  Computing Systems}, pages 2527--2530. ACM, 2012.

\bibitem{chen_iccv2017}
Q.~Chen and V.~Koltun.
\newblock Photographic image synthesis with cascaded refinement networks.
\newblock In {\em International Conference on Computer Vision (ICCV)}, 2017.

\bibitem{choi2017robust}
C.~Choi, S.~Ho~Yoon, C.-N. Chen, and K.~Ramani.
\newblock Robust hand pose estimation during the interaction with an unknown
  object.
\newblock In {\em Proceedings of the IEEE Conference on Computer Vision and
  Pattern Recognition}, pages 3123--3132, 2017.

\bibitem{choi2017learning}
C.~Choi, S.~Kim, and K.~Ramani.
\newblock Learning hand articulations by hallucinating heat distribution.
\newblock In {\em Proceedings of the IEEE Conference on Computer Vision and
  Pattern Recognition}, pages 3104--3113, 2017.

\bibitem{ganin2014unsupervised}
Y.~Ganin and V.~Lempitsky.
\newblock Unsupervised domain adaptation by backpropagation.
\newblock {\em arXiv preprint arXiv:1409.7495}, 2014.

\bibitem{ge_cvpr2016}
L.~Ge, H.~Liang, J.~Yuan, and D.~Thalmann.
\newblock {Robust 3D Hand Pose Estimation in Single Depth Images: from
  Single-View CNN to Multi-View CNNs}.
\newblock In {\em IEEE Conference on Computer Vision and Pattern Recognition
  (CVPR)}, 2016.

\bibitem{gomez_arxiv2017}
F.~Gomez{-}Donoso, S.~Orts{-}Escolano, and M.~Cazorla.
\newblock Large-scale multiview 3d hand pose dataset.
\newblock {\em CoRR}, abs/1707.03742, 2017.

\bibitem{goodfellow2014generative}
I.~Goodfellow, J.~Pouget-Abadie, M.~Mirza, B.~Xu, D.~Warde-Farley, S.~Ozair,
  A.~Courville, and Y.~Bengio.
\newblock Generative adversarial nets.
\newblock In {\em Advances in neural information processing systems}, pages
  2672--2680, 2014.

\bibitem{hamer2009tracking}
H.~Hamer, K.~Schindler, E.~Koller-Meier, and L.~Van~Gool.
\newblock Tracking a hand manipulating an object.
\newblock In {\em IEEE International Conference On Computer Vision (ICCV)},
  2009.

\bibitem{he_cvpr2016}
K.~He, X.~Zhang, S.~Ren, and J.~Sun.
\newblock Deep residual learning for image recognition.
\newblock In {\em IEEE Conference on Computer Vision and Pattern Recognition
  (CVPR)}, June 2016.

\bibitem{heap1996towards}
T.~Heap and D.~Hogg.
\newblock Towards 3d hand tracking using a deformable model.
\newblock In {\em Automatic Face and Gesture Recognition, 1996., Proceedings of
  the Second International Conference on}, pages 140--145. IEEE, 1996.

\bibitem{isola2017pix}
P.~Isola, J.-Y. Zhu, T.~Zhou, and A.~A. Efros.
\newblock Image-to-image translation with conditional adversarial networks.
\newblock {\em {IEEE Conference on Computer Vision and Pattern Recognition
  (CVPR)}}, 2017.

\bibitem{jia_caffe_ICM}
Y.~Jia, E.~Shelhamer, J.~Donahue, S.~Karayev, J.~Long, R.~Girshick,
  S.~Guadarrama, and T.~Darrell.
\newblock Caffe: Convolutional architecture for fast feature embedding.
\newblock In {\em Proceedings of the ACM International Conference on
  Multimedia}, 2014.

\bibitem{johnson2010lsp}
S.~Johnson and M.~Everingham.
\newblock Clustered pose and nonlinear appearance models for human pose
  estimation.
\newblock In {\em Proceedings of the British Machine Vision Conference (BMVC)},
  2010.

\bibitem{kesin_iccvw2011}
C.~Keskin, F.~Kıra\c, Y.~E. Kara, and L.~Akarun.
\newblock Real time hand pose estimation using depth sensors.
\newblock In {\em IEEE International Conference on Computer Vision Workshops
  (ICCVW)}, pages 1228--1234, 2011.

\bibitem{kingma2014adam}
D.~Kingma and J.~Ba.
\newblock Adam: A method for stochastic optimization.
\newblock {\em arXiv preprint arXiv:1412.6980}, 2014.

\bibitem{krejov2013multi}
P.~Krejov and R.~Bowden.
\newblock Multi-touchless: Real-time fingertip detection and tracking using
  geodesic maxima.
\newblock In {\em IEEE International Conference and Workshops on Automatic Face
  and Gesture Recognition (FG)}, pages 1--7, 2013.

\bibitem{lee2009multithreaded}
T.~Lee and T.~Hollerer.
\newblock Multithreaded hybrid feature tracking for markerless augmented
  reality.
\newblock {\em IEEE Transactions on Visualization and Computer Graphics},
  15(3):355--368, 2009.

\bibitem{liu2017learning}
J.~Liu and A.~Mian.
\newblock Learning human pose models from synthesized data for robust rgb-d
  action recognition.
\newblock {\em arXiv preprint arXiv:1707.00823}, 2017.

\bibitem{markussen2014vulture}
A.~Markussen, M.~R. Jakobsen, and K.~Hornb{\ae}k.
\newblock Vulture: A mid-air word-gesture keyboard.
\newblock In {\em Proceedings of ACM Conference on Human Factors in Computing
  Systems}, CHI '14, pages 1073--1082. ACM, 2014.

\bibitem{mehta_SIGGRAPH2017}
D.~Mehta, S.~Sridhar, O.~Sotnychenko, H.~Rhodin, M.~Shafiei, H.-P. Seidel,
  W.~Xu, D.~Casas, and C.~Theobalt.
\newblock Vnect: Real-time 3d human pose estimation with a single rgb camera.
\newblock volume~36, 2017.

\bibitem{mueller_iccv2017}
F.~Mueller, D.~Mehta, O.~Sotnychenko, S.~Sridhar, D.~Casas, and C.~Theobalt.
\newblock Real-time hand tracking under occlusion from an egocentric rgb-d
  sensor.
\newblock In {\em International Conference on Computer Vision ({ICCV})}, 2017.

\bibitem{oberweger_cvpr2016}
M.~Oberweger, G.~Riegler, P.~Wohlhart, and V.~Lepetit.
\newblock {Efficiently Creating 3D Training Data for Fine Hand Pose
  Estimation}.
\newblock In {\em {IEEE Conference on Computer Vision and Pattern Recognition
  (CVPR)}}, 2016.

\bibitem{oberweger_iccv2015}
M.~Oberweger, P.~Wohlhart, and V.~Lepetit.
\newblock Training a feedback loop for hand pose estimation.
\newblock In {\em IEEE International Conference on Computer Vision (ICCV)},
  pages 3316--3324, 2015.

\bibitem{oikonomidis2011full}
I.~Oikonomidis, N.~Kyriazis, and A.~A. Argyros.
\newblock Full dof tracking of a hand interacting with an object by modeling
  occlusions and physical constraints.
\newblock In {\em Computer Vision (ICCV), 2011 IEEE International Conference
  on}, pages 2088--2095. IEEE, 2011.

\bibitem{panteleris2017back}
P.~Panteleris and A.~Argyros.
\newblock Back to rgb: 3d tracking of hands and hand-object interactions based
  on short-baseline stereo.
\newblock {\em arXiv preprint arXiv:1705.05301}, 2017.

\bibitem{peng2017synthetic}
X.~Peng and K.~Saenko.
\newblock Synthetic to real adaptation with deep generative correlation
  alignment networks.
\newblock {\em arXiv preprint arXiv:1701.05524}, 2017.

\bibitem{piumsomboon2013user}
T.~Piumsomboon, A.~Clark, M.~Billinghurst, and A.~Cockburn.
\newblock User-defined gestures for augmented reality.
\newblock In {\em CHI'13 Extended Abstracts on Human Factors in Computing
  Systems}. ACM, 2013.

\bibitem{qian_cvpr2014}
C.~Qian, X.~Sun, Y.~Wei, X.~Tang, and J.~Sun.
\newblock {Realtime and Robust Hand Tracking from Depth}.
\newblock In {\em IEEE Conference on Computer Vision and Pattern Recognition
  (CVPR)}, pages 1106--1113, 2014.

\bibitem{rhodin2016egocap}
H.~Rhodin, C.~Richardt, D.~Casas, E.~Insafutdinov, M.~Shafiei, H.-P. Seidel,
  B.~Schiele, and C.~Theobalt.
\newblock Egocap: egocentric marker-less motion capture with two fisheye
  cameras.
\newblock {\em ACM Transactions on Graphics (TOG)}, 35(6):162, 2016.

\bibitem{romero_icra2010}
J.~Romero, H.~Kjellstr{\"o}m, and D.~Kragic.
\newblock Hands in action: real-time 3{D} reconstruction of hands in
  interaction with objects.
\newblock In {\em IEEE International Conference on Robotics and Automation
  (ICRA)}, 2010.

\bibitem{unet2015}
O.~Ronneberger, P.~Fischer, and T.~Brox.
\newblock U-net: Convolutional networks for biomedical image segmentation.
\newblock {\em CoRR}, abs/1505.04597, 2015.

\bibitem{ILSVRC15}
O.~Russakovsky, J.~Deng, H.~Su, J.~Krause, S.~Satheesh, S.~Ma, Z.~Huang,
  A.~Karpathy, A.~Khosla, M.~Bernstein, A.~C. Berg, and L.~Fei-Fei.
\newblock {ImageNet Large Scale Visual Recognition Challenge}.
\newblock {\em International Journal of Computer Vision (IJCV)},
  115(3):211--252, 2015.

\bibitem{sangkloy2016scribbler}
P.~Sangkloy, J.~Lu, C.~Fang, F.~Yu, and J.~Hays.
\newblock Scribbler: Controlling deep image synthesis with sketch and color.
\newblock {\em {IEEE Conference on Computer Vision and Pattern Recognition
  (CVPR)}}, 2017.

\bibitem{Schonemann:1966ch}
P.~H. Sch{\"o}nemann.
\newblock {A generalized solution of the orthogonal procrustes problem}.
\newblock {\em Psychometrika}, 31(1):1--10, Mar. 1966.

\bibitem{sharp2015accurate}
T.~Sharp, C.~Keskin, D.~Robertson, J.~Taylor, J.~Shotton, D.~Kim, C.~Rhemann,
  I.~Leichter, A.~Vinnikov, Y.~Wei, et~al.
\newblock Accurate, robust, and flexible real-time hand tracking.
\newblock In {\em Proceedings of ACM Conference on Human Factors in Computing
  Systems (CHI)}, pages 3633--3642. ACM, 2015.

\bibitem{shrivastava2017learning}
A.~Shrivastava, T.~Pfister, O.~Tuzel, J.~Susskind, W.~Wang, and R.~Webb.
\newblock Learning from simulated and unsupervised images through adversarial
  training.
\newblock In {\em IEEE Conference on Computer Vision and Pattern Recognition
  (CVPR)}, 2017.

\bibitem{simon2017hand}
T.~Simon, H.~Joo, I.~Matthews, and Y.~Sheikh.
\newblock Hand keypoint detection in single images using multiview
  bootstrapping.
\newblock In {\em IEEE Conference on Computer Vision and Pattern Recognition
  (CVPR)}, 2017.

\bibitem{sinha_cvpr2016}
A.~Sinha, C.~Choi, and K.~Ramani.
\newblock Deephand: robust hand pose estimation by completing a matrix imputed
  with deep features.
\newblock In {\em IEEE Conference on Computer Vision and Pattern Recognition
  (CVPR)}, pages 4150--4158, 2016.

\bibitem{sridhar_chi2015}
S.~Sridhar, A.~M. Feit, C.~Theobalt, and A.~Oulasvirta.
\newblock Investigating the dexterity of multi-finger input for mid-air text
  entry.
\newblock In {\em ACM Conference on Human Factors in Computing Systems}, pages
  3643--3652, 2015.

\bibitem{sridhar_cvpr2015}
S.~Sridhar, F.~Mueller, A.~Oulasvirta, and C.~Theobalt.
\newblock {Fast and Robust Hand Tracking Using Detection-Guided Optimization}.
\newblock In {\em IEEE Conference on Computer Vision and Pattern Recognition
  (CVPR)}, 2015.

\bibitem{sridhar_eccv2016}
S.~Sridhar, F.~Mueller, M.~Zollh{\"o}efer, D.~Casas, A.~Oulasvirta, and
  C.~Theobalt.
\newblock {Real-time Joint Tracking of a Hand Manipulating an Object from RGB-D
  Input}.
\newblock In {\em European Conference on Computer Vision ({ECCV})}, 2016.

\bibitem{sridhar_iccv2013}
S.~Sridhar, A.~Oulasvirta, and C.~Theobalt.
\newblock {Interactive markerless articulated hand motion tracking using RGB
  and depth data}.
\newblock In {\em IEEE Conference on Computer Vision and Pattern Recognition
  (CVPR)}, pages 2456--2463, 2013.

\bibitem{sridhar2014real}
S.~Sridhar, H.~Rhodin, H.-P. Seidel, A.~Oulasvirta, and C.~Theobalt.
\newblock Real-time hand tracking using a sum of anisotropic gaussians model.
\newblock In {\em 3D Vision (3DV), 2014 2nd International Conference on},
  volume~1, pages 319--326. IEEE, 2014.

\bibitem{stenger2006model}
B.~Stenger, A.~Thayananthan, P.~H. Torr, and R.~Cipolla.
\newblock Model-based hand tracking using a hierarchical bayesian filter.
\newblock {\em IEEE transactions on pattern analysis and machine intelligence},
  28(9):1372--1384, 2006.

\bibitem{tagliasacchi_sgp2015}
A.~Tagliasacchi, M.~Schroeder, A.~Tkach, S.~Bouaziz, M.~Botsch, and M.~Pauly.
\newblock {Robust Articulated-ICP for Real-Time Hand Tracking}.
\newblock {\em Computer Graphics Forum (Symposium on Geometry Processing)},
  34(5), 2015.

\bibitem{tang2014latent}
D.~Tang, H.~Jin~Chang, A.~Tejani, and T.-K. Kim.
\newblock Latent regression forest: Structured estimation of 3d articulated
  hand posture.
\newblock In {\em Proceedings of the IEEE Conference on Computer Vision and
  Pattern Recognition}, pages 3786--3793, 2014.

\bibitem{tang_iccv2015}
D.~Tang, J.~Taylor, P.~Kohli, C.~Keskin, T.-K. Kim, and J.~Shotton.
\newblock Opening the black box: Hierarchical sampling optimization for
  estimating human hand pose.
\newblock In {\em Proc. ICCV}, 2015.

\bibitem{taylor_siggraph2016}
J.~Taylor, L.~Bordeaux, T.~Cashman, B.~Corish, C.~Keskin, T.~Sharp, E.~Soto,
  D.~Sweeney, J.~Valentin, B.~Luff, et~al.
\newblock Efficient and precise interactive hand tracking through joint,
  continuous optimization of pose and correspondences.
\newblock {\em ACM Transactions on Graphics (TOG)}, 35(4):143, 2016.

\bibitem{TenBerge:2006ih}
J.~M.~F. Ten~Berge.
\newblock {The rigid orthogonal Procrustes rotation problem}.
\newblock {\em Psychometrika}, 71(1):201--205, 2006.

\bibitem{tompson_tog2014}
J.~Tompson, M.~Stein, Y.~Lecun, and K.~Perlin.
\newblock Real-time continuous pose recovery of human hands using convolutional
  networks.
\newblock {\em ACM Transactions on Graphics}, 33, August 2014.

\bibitem{tzeng2017adversarial}
E.~Tzeng, J.~Hoffman, K.~Saenko, and T.~Darrell.
\newblock Adversarial discriminative domain adaptation.
\newblock {\em arXiv preprint arXiv:1702.05464}, 2017.

\bibitem{tzionas_ijcv2016}
D.~Tzionas, L.~Ballan, A.~Srikantha, P.~Aponte, M.~Pollefeys, and J.~Gall.
\newblock Capturing hands in action using discriminative salient points and
  physics simulation.
\newblock {\em International Journal of Computer Vision (IJCV)}, 2016.

\bibitem{tzionas20153d}
D.~Tzionas and J.~Gall.
\newblock 3d object reconstruction from hand-object interactions.
\newblock In {\em Proceedings of the IEEE International Conference on Computer
  Vision}, pages 729--737, 2015.

\bibitem{varol_cvpr2017}
G.~Varol, J.~Romero, X.~Martin, N.~Mahmood, M.~J. Black, I.~Laptev, and
  C.~Schmid.
\newblock Learning from synthetic humans.
\newblock In {\em IEEE Conf. on Computer Vision and Pattern Recognition
  (CVPR)}, July 2017.

\bibitem{wan2017crossing}
C.~Wan, T.~Probst, L.~Van~Gool, and A.~Yao.
\newblock Crossing nets: Combining gans and vaes with a shared latent space for
  hand pose estimation.
\newblock In {\em Proceedings of the IEEE Conference on Computer Vision and
  Pattern Recognition}, pages 680--689, 2017.

\bibitem{wan_eccv2016}
C.~Wan, A.~Yao, and L.~Van~Gool.
\newblock Hand pose estimation from local surface normals.
\newblock In {\em European Conference on Computer Vision}, pages 554--569.
  Springer, 2016.

\bibitem{wang20116d}
R.~Wang, S.~Paris, and J.~Popovi{\'c}.
\newblock 6d hands: markerless hand-tracking for computer aided design.
\newblock In {\em Proc.\ of UIST}, pages 549--558. ACM, 2011.

\bibitem{wang_TOG2009}
R.~Y. Wang and J.~Popovi{\'c}.
\newblock {Real-time hand-tracking with a color glove}.
\newblock {\em ACM Transactions on Graphics}, 28(3), 2009.

\bibitem{xu2013efficient}
C.~Xu and L.~Cheng.
\newblock Efficient hand pose estimation from a single depth image.
\newblock In {\em Proceedings of the IEEE International Conference on Computer
  Vision}, pages 3456--3462, 2013.

\bibitem{ye_eccv2016}
Q.~Ye, S.~Yuan, and T.-K. Kim.
\newblock Spatial attention deep net with partial pso for hierarchical hybrid
  hand pose estimation.
\newblock In {\em European Conference on Computer Vision (ECCV)}, pages
  346--361. Springer, 2016.

\bibitem{zeiler2012adadelta}
M.~D. Zeiler.
\newblock Adadelta: an adaptive learning rate method.
\newblock {\em arXiv preprint arXiv:1212.5701}, 2012.

\bibitem{zhang2016stereo}
J.~Zhang, J.~Jiao, M.~Chen, L.~Qu, X.~Xu, and Q.~Yang.
\newblock 3d hand pose tracking and estimation using stereo matching.
\newblock {\em arXiv preprint arXiv:1610.07214}, 2016.

\bibitem{zhou_ijcai2016}
X.~Zhou, Q.~Wan, W.~Zhang, X.~Xue, and Y.~Wei.
\newblock Model-based deep hand pose estimation.
\newblock In {\em IJCAI}, 2016.

\bibitem{CycleGAN2017}
J.-Y. Zhu, T.~Park, P.~Isola, and A.~A. Efros.
\newblock Unpaired image-to-image translation using cycle-consistent
  adversarial networks.
\newblock In {\em International Conference on Computer Vision (ICCV)}, 2017.

\bibitem{Zimmermann:2017um}
C.~Zimmermann and T.~Brox.
\newblock {Learning to Estimate 3D Hand Pose from Single RGB Images.}
\newblock In {\em International Conference on Computer Vision (ICCV)}, 2017.

\end{thebibliography}
